\documentclass[twocolumn,natbib]{svjour3}

\usepackage[T1]{fontenc}
\usepackage[latin1]{inputenc}
\usepackage{color}
\usepackage{array}
\usepackage{rotating}
\usepackage{bm}
\usepackage{amsmath}
\usepackage{amssymb}
\usepackage{amsfonts}
\usepackage{graphicx}
\usepackage{esint}
\usepackage[unicode=true,pdfusetitle,
 bookmarks=false,
 breaklinks=true,pdfborder={0 0 1},backref=section,colorlinks=false]
 {hyperref}
\hypersetup{
 pagebackref=true,letterpaper=true,colorlinks}
\usepackage{xspace}
\usepackage{multirow}
\usepackage{subfigure}
\usepackage{mathtools}
\usepackage{booktabs}
\usepackage{enumitem}

\usepackage{float}
\usepackage{algorithm}
\usepackage{soul}

\makeatletter
\DeclareRobustCommand\onedot{\futurelet\@let@token\@onedot}
\def\@onedot{\ifx\@let@token.\else.\null\fi\xspace}
\def\eg{\emph{e.g}\onedot} 
\def\ie{\emph{i.e}\onedot} 
 
\def\etc{\emph{etc}\onedot}

\def\wrt{\emph{w.r.t}\onedot}

\def\eq{Eq.~}
\makeatother

\DeclareMathOperator*{\argmax}{arg\,max}

\makeatletter
\newcommand\footnoteref[1]{\protected@xdef\@thefnmark{\ref{#1}}\@footnotemark}
\makeatother

\newcommand{\refp}[1]{(\ref{#1})}
\newcommand{\red}[1]{{\color{black}{#1}}}
\newcommand{\remm}[1]{} 

\makeatother

\begin{document}

\authorrunning{J. Revaud, P. Weinzaepfel, Z. Harchaoui and C. Schmid}
\titlerunning{DeepMatching: Hierarchical Deformable Dense Matching}
\title{DeepMatching: Hierarchical Deformable Dense Matching}

\author{Jerome Revaud \qquad{} Philippe Weinzaepfel \qquad{} Zaid Harchaoui \qquad {}Cordelia Schmid \\
 INRIA \\
 \texttt{\small{firstname.lastname@inria.fr}}}

\institute{LEAR team, Inria Grenoble Rhone-Alpes, Laboratoire Jean Kuntzmann, CNRS, Univ. Grenoble Alpes, France.}

\maketitle

\begin{abstract}

\red{We introduce a novel matching algorithm, called DeepMatching, to compute 
dense correspondences between images. DeepMatching relies on a hierarchical,
multi-layer, \red{correlational} architecture designed for matching images and 
was inspired by deep convolutional approaches.}
The proposed matching algorithm can handle non-rigid deformations and 
repetitive textures and efficiently determines dense 
correspondences in the presence of significant changes between images.

We evaluate the performance of DeepMatching, in comparison with state-of-the-art matching algorithms, 
on the Mikolajczyk~\citep{Mikolajczyk2005}, the MPI-Sintel~\citep{sintel} 
 and the Kitti~\citep{kitti} datasets.
DeepMatching outperforms the state-of-the-art algorithms 
and shows excellent results in particular for repetitive textures. 

We also propose a method for estimating optical flow, called DeepFlow,
by integrating DeepMatching in \red{the large displacement optical flow (LDOF)
approach of \cite{Bro11a}.
Compared to existing matching algorithms, 
additional robustness to large displacements and complex motion is obtained thanks
to our matching approach. DeepFlow obtains competitive performance on
public benchmarks for optical flow estimation.}


\end{abstract}

\section{Introduction}
\label{sec:intro}

Computing correspondences between related images is a central issue in many
computer vision problems, ranging from scene recognition to optical
flow estimation \citep{forsyth2011computer,Szeliski2010}. 
The goal of a matching algorithm is to discover shared visual content 
between two images, and 
to establish as many as possible precise point-wise correspondences, called \emph{matches}. 
An essential
aspect of matching approaches is the amount of rigidity they assume
when computing the correspondences.  In fact, matching
approaches range between two extreme cases:  stereo matching, where
matching hinges upon strong geometric constraints, 
and matching ``in the wild'', where the set of possible
transformations from the source image to the target one  is large and the
problem is basically almost unconstrained. 
Effective approaches have
been designed for matching rigid objects across images in the presence
of  large viewpoint changes~\citep{Lowe2004,Barnes2010,HaCohen2011}.
However, the performance of current state-of-the-art matching
algorithms for images ``in the wild'', such as consecutive images in
real-world videos featuring fast non-rigid motion, still calls for
improvement~\citep{mdpof,Chen2013}.   
In this paper, we aim at tackling matching in such a general setting.

Matching algorithms for images ``in the wild'' need to accommodate several requirements, that turn out to be often in contradiction. On one hand,
matching objects necessarily requires rigidity assumptions to some extent. 
It is also mandatory that these objects have sufficiently discriminative textures to make the problem well-defined. 
On the other hand, 
many objects or regions are not rigid objects, like humans or animals. Furthermore, large portions of an image
are usually occupied by weakly-to-no textured regions, often with repetitive textures, like sky or bucolic background.

Descriptor matching approaches, such as SIFT~\citep{Lowe2004}
or HOG~\citep{Dalal2005,Bro11a} matching, compute discriminative feature representations from rectangular patches. 
However, while these approaches succeed in case of rigid motion, they fail to match regions with weak or repetitive textures, as local patches are poorly discriminative.
Furthermore, matches are usually poor and imprecise in case of non-rigid deformations, as these approaches rely on rigid patches.
Discriminative power can be traded against increased robustness to non-rigid deformations. Indeed, propagation-based approaches, 
such as Generalized PatchMatch~\citep{Barnes2010} or Non-rigid Dense Correspondences~\citep{HaCohen2011}, 
compute simple feature representations from small patches and propagate matches to neighboring patches. They yield good  performance in case of non-rigid deformations. However, matching repetitive textures remains beyond the reach of these approaches. 

In this paper we propose a novel approach, called DeepMatching, that gracefully combines the strengths of these two families of approaches. 
DeepMatching is computed using a multi-layer architecture, which 
breaks down patches into a hierarchy of sub-patches. This architecture allows to
 work at several scales and handle repetitive textures. Furthermore, within each layer, local 
matches are computed assuming a restricted set of feasible rigid deformations. Local matches
are then propagated up the hierarchy, which progressively discard spurious incorrect matches. 
We called our approach DeepMatching, as it is inspired by deep convolutional approaches. 

\noindent In summary, we make three contributions:

\textbf{$\bullet$ Dense matching}: we propose a
matching algorithm, DeepMatching, that allows to robustly
determine dense correspondences between two images. It
explicitly handles non-rigid deformations, with bounds on the
deformation tolerance, and  incorporates a multi-scale scoring of the
matches, making it robust to repetitive or weak textures. 
\red{Furthermore, our approach is based on gradient histograms, and thus robust to appearance changes caused by illumination and color variations.}

\textbf{ $\bullet$ Fast, scale/rotation-invariant matching}: we propose 
a computationally efficient version of DeepMatching, which performs almost as well
as exact DeepMatching, but at a much lower memory cost. Furthermore, this fast version of 
DeepMatching can be extended to a scale and rotation-invariant version, 
making it an excellent competitor to state-of-the-art descriptor matching approaches.

\red{\textbf{ $\bullet$ Large-displacement optical flow}: we propose an 
optical flow approach which uses DeepMatching in the matching
term of the large displacement variational energy minimization of \cite{Bro11a}. 
We show that DeepMatching is a better choice compared to the HOG
descriptor used by \cite{Bro11a} and other state-of-the-art 
matching algorithms. The approach, named DeepFlow, obtains competitive
results on public optical flow benchmarks. } 

This paper is organized as follows. After a review of previous works
(Section~\ref{sec:related}), we start by presenting the proposed matching
algorithm, DeepMatching, in Section~\ref{sec:dm}. Then, Section~\ref{sec:extensions}
describes several extensions of DeepMatching. In particular, we propose an 
optical flow approach, DeepFlow, in Section~\ref{sec:flow}. Finally,
we present experimental results in Section~\ref{sec:xp}. 

A preliminary version of this article has appeared in~\cite{DeepFlow}.
This version adds (1)~an in-depth presentation of DeepMatching;
(2)~an enhanced version of DeepMatching, which improves the 
match scoring and the selection of entry points for backtracking;
(3)~proofs on time and memory complexity of DeepMatching as well as its
deformation tolerance; (4)~a discussion on the connection between
Deep Convolutional Neural Networks and  DeepMatching; (5)~a fast 
approximate version of DeepMatching; (6)~a scale and rotation
invariant version of DeepMatching; and (7)~an extensive 
experimental evaluation of DeepMatching on several state-of-the-art
 benchmarks. 
The code for DeepMatching as well as DeepFlow are available at 
\url{http://lear.inrialpes.fr/src/deepmatching/} and 
\url{http://lear.inrialpes.fr/src/deepflow/}. 
Note that we provide a GPU implementation in addition to the CPU one.

\section{Related work}
\label{sec:related}

In this section we review related work on  ``general'' image matching,
that is matching without prior knowledge and constraints, 
and on matching in the context of optical flow estimation, 
that is matching consecutive images in videos. 

\subsection{General image matching}

Image matching based on local features has been extensively
studied in the past decade. It has been applied successfully to various
domains, such as wide baseline stereo matching \citep{Furukawa2010}
and image retrieval \citep{Philbin2010}. It consists of
two steps, \ie, extracting local descriptors and matching them. 
Image descriptors are extracted in rigid (generally square) local frames 
at sparse invariant image locations~\citep{Mikolajczyk2005,Szeliski2010}. 
Matching then equals nearest neighbor search between
descriptors, followed by an optional geometric verification. 
\red{Note that a confidence value can be obtained by computing the uniqueness of a match, \ie, by looking at the distance of its nearest neighbors~\citep{Lowe2004,Bro11a}.}
While this class of techniques is well suited for well-textured rigid
objects, it fails to match non-rigid objects and 
weakly textured regions. 

In contrast, the proposed matching algorithm, called \emph{DeepMatching},
is inspired by non-rigid 2D warping and deep convolutional
networks \citep{LeCun98,Uchida1998,Keysers2007}. This family of approaches
explicitly models non-rigid deformations.
We employ a novel family of feasible warpings that does not enforce
monotonicity nor continuity constraints, in contrast to traditional 
2D warping~\citep{Uchida1998,Keysers2007}. This makes the problem
computationally much less expensive.

It is also worthwhile to mention the similarity with non-rigid matching approaches
developed for a broad range of applications. 
\cite{Ecker2009} proposed a similar pipeline to ours (albeit more complex) to measure
the similarity of small images. However, their method lacks a way
of merging correspondences belonging to objects with contradictory
motions, \eg, on different focal planes. 
For the purpose of establishing dense correspondences between images,
\cite{Wills2006} estimated a non-rigid matching 
by robustly fitting smooth parametric models (homography and splines)
to local descriptor matches. In contrast, our approach is non-parametric
and model-free.

Recently, fast algorithms for dense patch matching
have taken advantage of the redundancy between overlapping 
patches~\citep{Barnes2010,Korman2011,kpm,daisyff}.
The insight is to propagate good matches to their neighborhood in
a loose fashion, yielding dense non-rigid matches. In practice,
however, the lack of a  smoothness constraint leads to highly
discontinuous matches. \red{Several works have proposed ways to fix this.}
\cite{HaCohen2011} reinforce
neighboring matches using an iterative multiscale expansion and contraction
strategy, performed in a coarse-to-fine manner. 
\remm{Yet, the algorithm matches poorly discriminative patches
and, as such, cannot overcome the inherent weaknesses of patch
matching approaches.}
\red{\cite{daisyff} include a guided filtering stage
on top of PatchMatch, which obtains smooth correspondence fields by 
locally approximating a MRF. }
Finally, \cite{Kim2013}
propose a hierarchical matching to obtain dense correspondences,
using a coarse-to-fine (top-down) strategy. Loopy belief propagation
is used to perform inference. 

In contrast to these approaches, DeepMatching proceeds bottom-up and, then, top-down. 
\red{Due to its hierarchical nature, DeepMatching is able to consider
patches at several scales, thus overcoming the lack of distinctiveness
that affects small patches.} Yet, the multi-layer construction
allows to efficiently perform matching allowing semi-rigid local deformations. 
In addition, DeepMatching can be computed efficiently, and can be
further accelerated to satisfy low-memory requirements with negligible
loss in accuracy.  

\subsection{Matching for flow estimation}

Variational energy minimization is currently the most popular
framework for optical flow estimation. Since the pioneering work of \cite{Horn1981}, 
research has focused on alleviating the drawbacks of this approach. 
A series of improvements were proposed over 
the years~\citep{Black1996,Werlberger2009,Bruhn2005,
Papenberg2006,middlebury,Sun2014,Vogel2013data}.
The variational approach of~\cite{Bro04a} combines most of these improvements in a unified framework.
The energy decomposes into several terms, resp. the data-fitting and the
smoothness terms. Energy minimization is performed by solving the Euler-Lagrange
equations, reducing the problem to solving a sequence of large
and structured linear systems.

More recently, the addition of a descriptor matching term in the energy to be minimized
was proposed by~\cite{Bro11a}. Following this idea, several papers~\citep{Tola2008,Bro11a,siftflow,siftscales}
show that dense descriptor matching improves performance. 
\red{Strategies such as reciprocal nearest-neighbor verification~\citep{Bro11a} allow to prune most of the false matches.}
However, a variational energy minimization approach that includes
such a descriptor matching term may fail at locations where matches
are missing or wrong. 

Related approaches tackle the problem of dense scene correspondence.
\red{SIFT-flow \citep{siftflow}, one of the most famous method in this context, also formulates the 
matching problem in a variational framework. 
\cite{siftscales} improve over SIFT-flow by using multi-scale patches.
However, this decreases performance in cases where 
scale invariance is not required. }
\cite{mdpof}
integrate matching of SIFT \citep{Lowe2004} and PatchMatch \citep{Barnes2010} to refine the flow initialization at each
level. Excellent results are obtained for optical flow estimation, yet at the cost of expensive
fusion steps. \cite{Leordeanu2013} extends
sparse matches with locally affine constraints to dense matches and,
then, uses a total variation algorithm to refine the flow estimation.
We present here a computationally efficient and competitive approach
for large displacement optical flow by integrating the proposed DeepMatching
algorithm into the approach of \cite{Bro11a}.

\section{DeepMatching}
\label{sec:dm}

This section introduces our matching algorithm DeepMatching.
DeepMatching is a matching algorithm based on correlations at the patch-level,  
that proceeds in a multi-layer fashion. 
The multi-layer architecture relies on a quadtree-like patch subdivision scheme, 
with an extra degree of freedom to locally re-optimize the positions of each quadrant. 
In order to enhance the contrast of the spatial correlation maps output by the local correlations, 
a nonlinear transformation is applied after each layer.

We first give an overview of DeepMatching in Section~\ref{sub:insight}
and show that it can be decomposed in a bottom-up pass followed by a top-down pass. 
We, then,  present the bottom-up pass in Section~\ref{sub:responsepyramid} and the top-down one in Section~\ref{sub:backtrackcorrespondences}.  
Finally, we analyze DeepMatching in Section~\ref{sub:analysis}.

\subsection{Overview of the approach}
\label{sub:insight} 

\def\R{\mathbf{R}}

\begin{figure}
  \centering
  \includegraphics[width=1\linewidth]{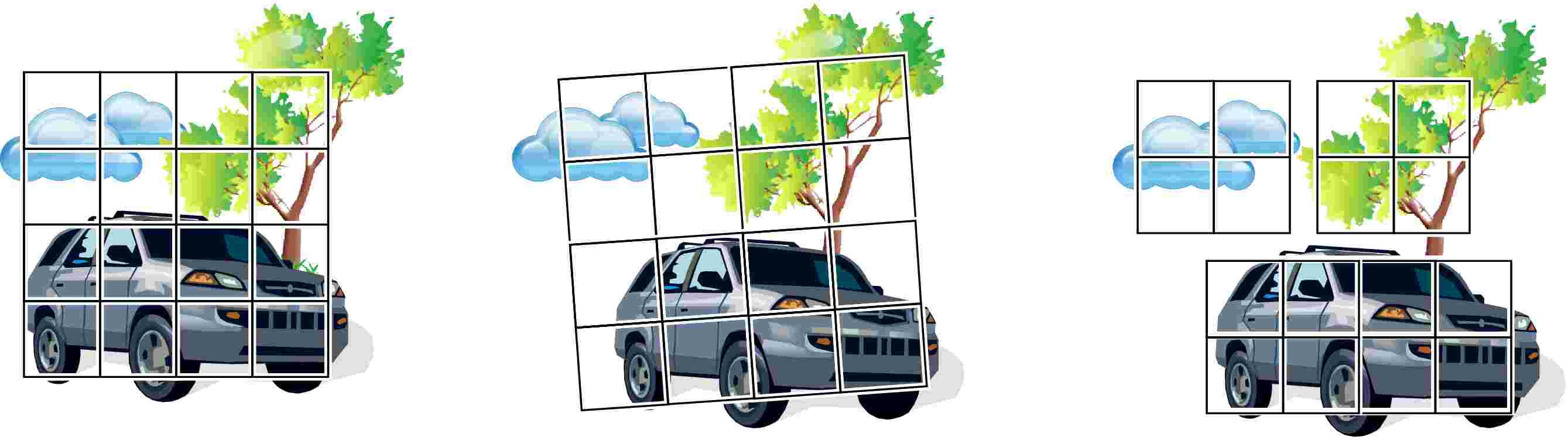}
  \caption{Illustration of moving quadrant similarity: a quadrant is a quarter of a SIFT patch, \ie a group of $2\times2$ cells.
    \emph{Left:} SIFT descriptor in the first image.
    \emph{Middle:} second image with optimal standard SIFT matching (rigid).
    \emph{Right:} second image with optimal \emph{moving quadrant} SIFT matching.
    In this example, the patch covers various objects moving in different directions: for instance
    the car moves to the right while the cloud to the left. Rigid
    matching fails to capture this, whereas the moving quadrant approach is able to follow each object.}
  \label{fig:sift}
\end{figure}

A state-of-the-art approach for matching regions between two images is based on the SIFT descriptor \citep{Lowe2004}. 
SIFT is a histogram of gradients with $4\times4$ spatial
\red{and $8$ orientation bins}, yielding a robust descriptor $\R\in\mathbb{R}^{4\times4\times8}$ that
effectively encodes a square image region.  
Note that its $4\times 4$ cell grid can also be viewed as 4 so-called ``quadrants'' of $2\times 2$ cells, see Figure~\ref{fig:sift}. 
We can, then, rewrite $\R=[\R_0,\, \R_1,\, \R_2,\, \R_3]$ with $\R_{i}\in\mathbb{R}^{2\times2\times8}$. 

\begin{figure}
  \centering
  \includegraphics[bb=50bp 370bp 515bp 550bp,clip,width=1\linewidth]{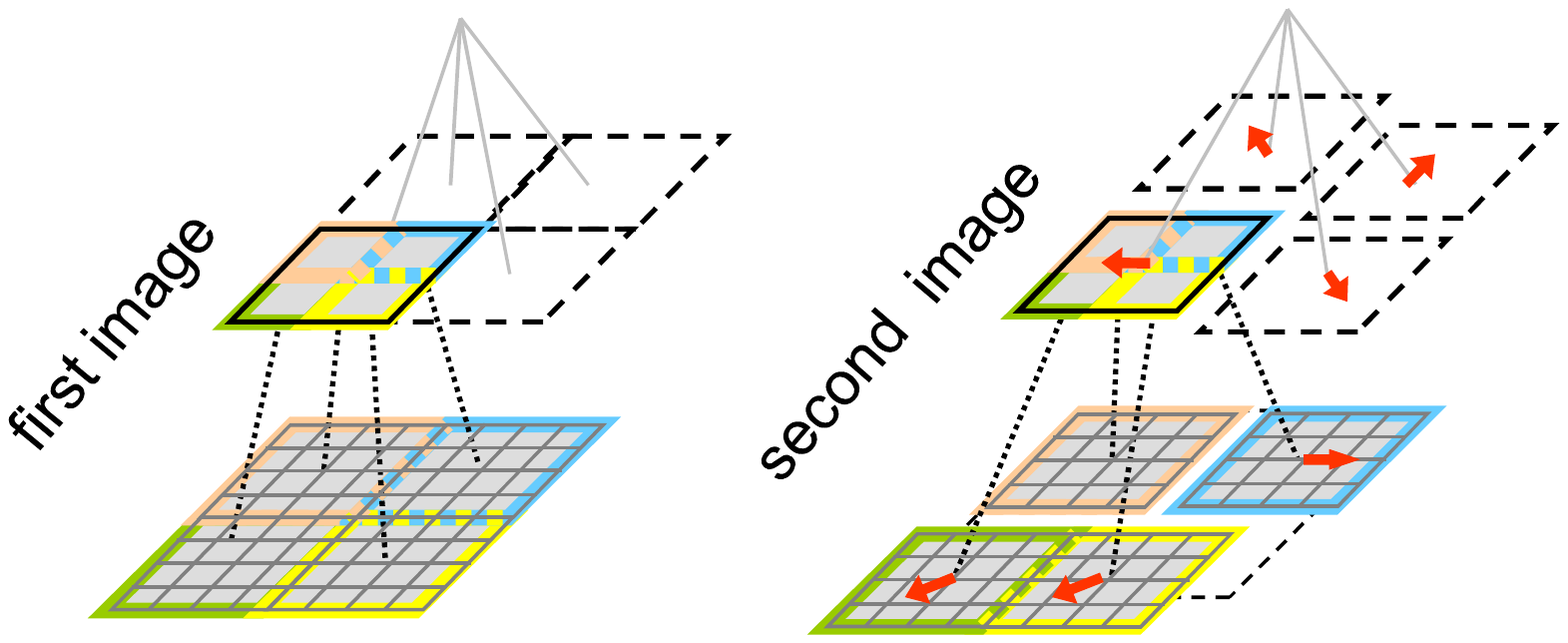}
  \caption{\textit{Left:} Quadtree-like patch hierarchy in the first image.
    \textit{Right:} one possible displacement of corresponding patches in the second image.}
  \label{fig:recur}
\end{figure}

Let $\R$ and $\R'$ be the SIFT descriptors of the corresponding regions in the source and target image.
In order to remove the effect of non-rigid motion,
we propose to optimize the \emph{positions} $p_i \in \mathbb{R}^2$ of the $4$ quadrants of the target descriptor $\R'$ 
(rather than keeping them fixed), in order to maximize 
\begin{equation}
\mathrm{sim}(\R,\R')=\max_{\left\{p_i\right\}}\;\frac{1}{4}\sum_{i=0}^{3}\mathrm{sim}\left(\R_{i}, \R_i'(p_{i})\right),
\label{eqn:quadrant0}
\end{equation}
where $\R_i'(p_i)$ is the descriptor of a single quadrant
extracted at position $p_i$ and $\mathrm{sim}()$ a similarity function. 
Now, $\mathrm{sim}(\R,\R')$ is able to handle situations such as the one presented in Figure~\ref{fig:sift},
where a region contains multiple objects moving in different directions.
Furthermore, if the four quadrants can move independently (of course, within some extent), 
it can be calculated more efficiently as:
\begin{equation}
\mathrm{sim}(\R,\R')=\frac{1}{4}\sum_{i=0}^{3}\max_{p_{i}}\;\mathrm{sim}\left(\R_{i}, \R_i'(p_{i})\right),
\label{eqn:quadrant}
\end{equation}
When applied recursively to each quadrant by subdivided it into 4
sub-quadrants until a minimum patch size is reached (atomic patches),
this strategy allows for accurate non-rigid matching. 
Such a recursive decomposition can be represented as a quad-tree, see Figure~\ref{fig:recur}. 
Given an initial pair of two matching regions, retrieving atomic patch correspondences is then done in a top-down fashion (\ie by recursively applying \eq\refp{eqn:quadrant} to the quadrant's positions $\{p_i\}$).

Nevertheless, in order to first determine the set of matching regions
between the two images, we need to compute beforehand the matching
scores (\ie similarity) of all large-enough patches in the two images
(as in Figure~\ref{fig:sift}), and keep the pairs with maximum
similarity. As indicated by \eq\refp{eqn:quadrant}, the score is
formed by averaging the max-pooled scores of the quadrants.
Hence, the process of computing the matching scores is bottom-up. In
the following, we call \emph{correlation map} the matching scores of a
single patch from the first image at every position in the second
image. Selecting matching patches then corresponds to finding local maxima
in the correlation maps. 

To sum-up, the algorithm can be decomposed in two steps:
(i) first, correlation maps are computed using a bottom-up algorithm,
as shown in Figure~\ref{fig:global}. 
Correlation maps of small patches are first computed and then aggregated to form correlation maps of larger patches;
(ii) next, a top-down method estimates the motion of atomic patches starting from matches of large patches.

In the remainder of this section, we detail the two steps described above 
(Section~\ref{sub:responsepyramid} and Section~\ref{sub:backtrackcorrespondences}), before 
analyzing the properties of DeepMatching in Section~\ref{sub:analysis}.

\subsection{Bottom-up correlation pyramid computation}
\label{sub:responsepyramid}

Let $I$ and $I'$ be two images of resolution $W \times H$ and $W' \times H'$. 

\begin{figure}
 \centering
 \includegraphics[width=\linewidth]{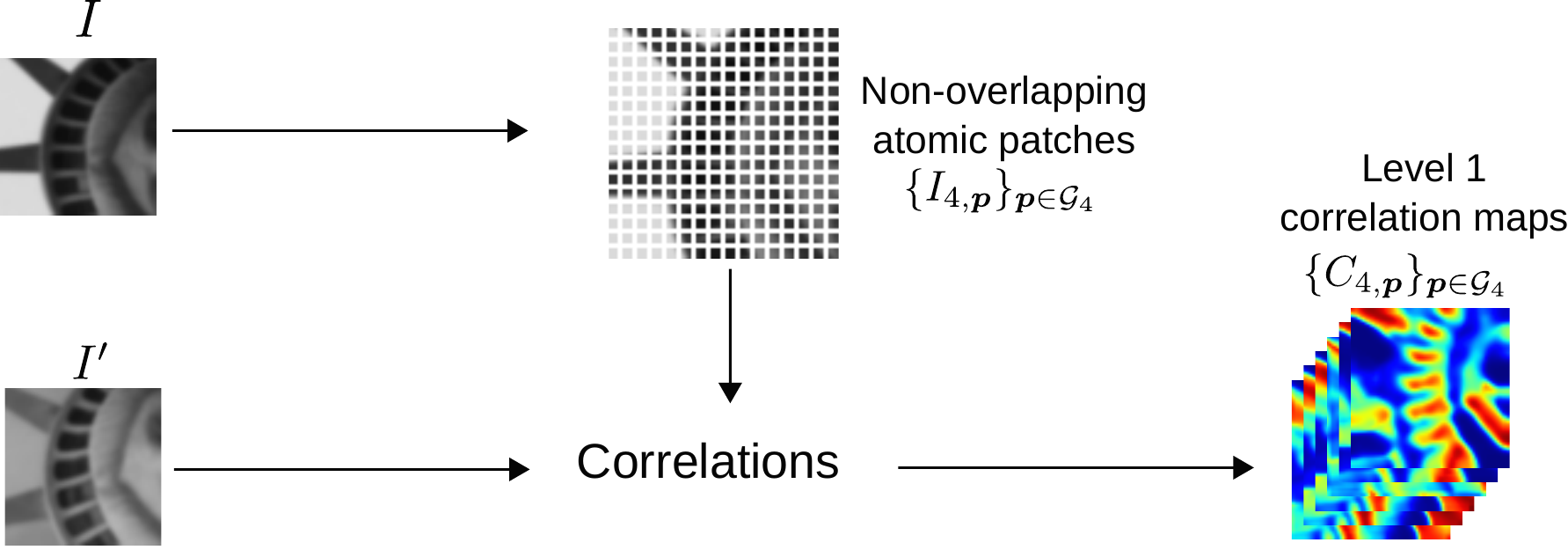}
 \caption{Computing the bottom level correlation maps $\{C_{4,\bm{p}}\}_{\bm{p} \in \mathcal{G}_4}$. Given two images $I$ and $I'$, the first one is split into
 non-overlapping atomic patches of size $4 \times 4$ pixels. For each patch, we compute the \red{correlation at every location of $I'$ to obtain the corresponding correlation map}.}
 \label{fig:convol}
\end{figure}

\paragraph{Bottom level.}
We use patches of size $4 \times 4$ pixels as atomic patches.
We split $I$ into non-overlapping atomic patches, and compute the correlation map with image $I'$
for each of them, see Figure~\ref{fig:convol}. 
The score between two atomic patches $\R$ and $\R'$ is defined as the average  pixel-wise similarity:
\begin{equation}
\mathrm{sim}(\R,\R') = \frac{1}{16} \sum_{i=0}^3\sum_{j=0}^3
\R_{i,j}^\top \R_{i,j}',
\label{eqn:correlation}
\end{equation}
where each pixel $\R_{i,j}$ is represented as a histogram of oriented gradients pooled over a local neighborhood.
We detail below how the pixel descriptor is computed.
\paragraph{\red{Pixel descriptor $\R_{i,j}$: }}
\red{We rely on a robust pixel representation
that is similar in spirit to SIFT and DAISY \citep{Lowe2004,daisy}. 
Given an input image $I$, we first apply a Gaussian smoothing of radius $\nu_{1}$ in order to
denoise $I$ from potential artifacts caused for example by JPEG compression.
We then extract the gradient $(\delta x, \delta y)$ at each pixel and compute its non-negative
projection onto 8 orientations $\left\{ (\cos i\frac{\pi}{4},\sin i\frac{\pi}{4})\right\} _{i=1\ldots8}$. 
At this point, we obtain 8 oriented gradient maps.
We smooth each map with a Gaussian filter of radius $\nu_{2}$.
Next we cap strong gradients using
a sigmoid $x \mapsto 2/(1+\exp(-\varsigma x))-1$, 
to help canceling out effects of varying illumination.
We smooth gradients one more time for each orientation with a Gaussian
filter of radius $\nu_{3}$. Finally, the descriptor for each
pixel is obtained by the $\ell_2$-normalized concatenation of 8 oriented 
gradients and a ninth small constant value $\mu$.
Appending $\mu$ amounts to adding a regularizer that will reduce the importance of
small gradients (\ie noise) and ensures that two pixels lying in 
areas without gradient information will still correlate positively. 
Pixel descriptors $\R_{i,j}$ are compared using dot-product and the similarity function
takes value in the interval $[0,1]$. In Section~\ref{sub:paramstuning}, we evaluate the impact of 
the parameters of this pixel descriptor.}

\begin{figure}
  \centering
  \includegraphics[width=\linewidth]{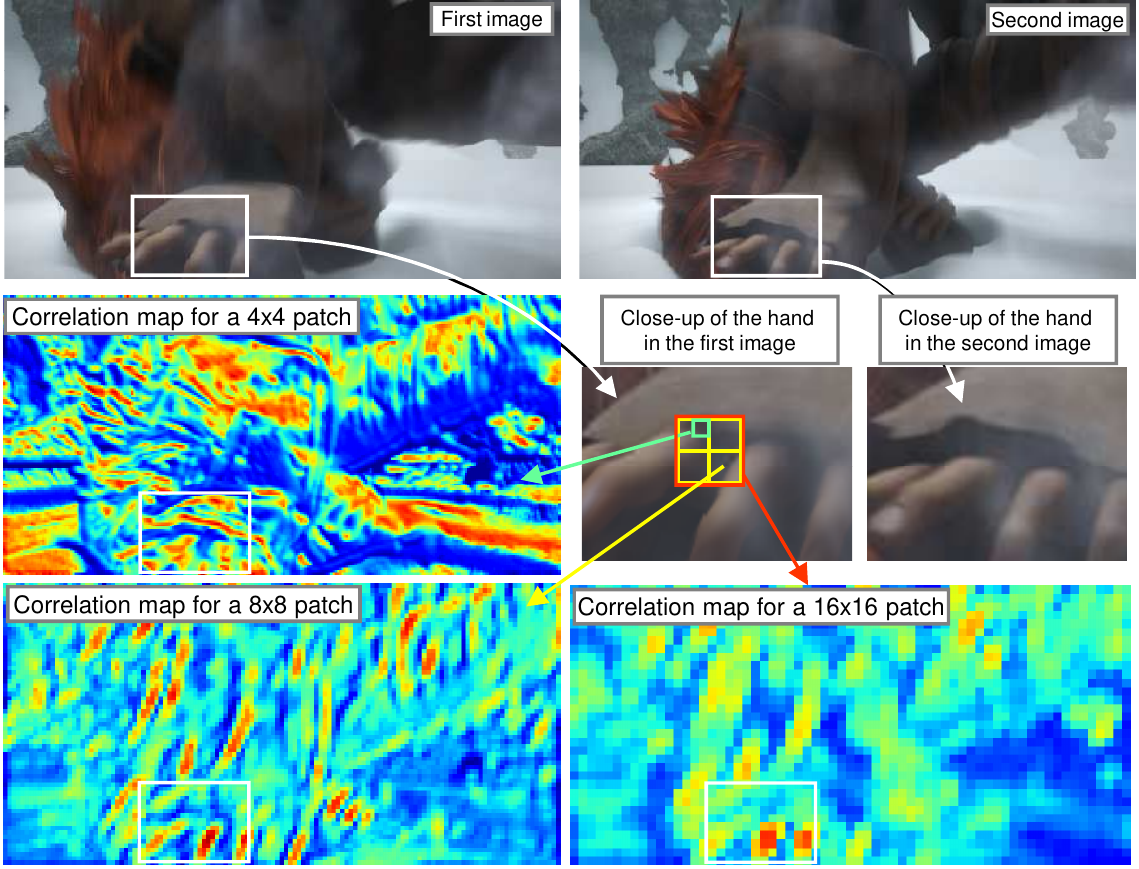}
  \caption{Correlation maps for patches of different size. \textit{Middle-left}: correlation map of a 4x4 patch. \textit{Bottom-right}:
    correlation map of a 16x16 patch obtained by aggregating correlation responses
    of children 8x8 patches (\textit{bottom-left}), themselves obtained
    from 4x4 patches. The map of the 16x16 patch is clearly more discriminative
    than previous ones despite the change in appearance of the region.}
\label{fig:convolution} 
\end{figure}

\paragraph{Bottom-level correlation map:}
We can express the correlation map computation obtained from \eq\refp{eqn:correlation}
 more conveniently in a convolutional framework.
Let $I_{N,\bm{p}}$ be a patch of size $N \times N$ from the first image centered at $\bm{p}$ 
($N\geq4$ is a power of 2).
Let $\mathcal{G}_4 = \{ 2,6,10, ..., W-2 \} \times \{ 2,6,10,..., H-2 \}$ be a grid with step $4$ pixels.
$\mathcal{G}_4$ is the set of the centers of the atomic patches.
For each $\bm{p} \in \mathcal{G}_4$, we convolve the \red{flipped} patch $I_{4,\bm{p}}^\mathrm{F}$ over $I'$
\begin{equation}
 C_{4,\bm{p}} = I_{4,\bm{p}}^\mathrm{F} \star I' ,
 \label{eqn:init}
\end{equation}
to get the correlation map $C_{4,\bm{p}}$, 
\red{where $.^\mathrm{F}$ denotes an horizontal and vertical flip\footnote{This 
amounts to the cross-correlation of the patch and $I'$.}}. 
For any pixel $\bm{p}'$ of $I'$, $C_{4,\bm{p}}(\bm{p}')$ is a measure of similarity between $I_{4,\bm{p}}$ and $I'_{4,\bm{p}'}$.
Examples of such correlation maps are shown in Figure~\ref{fig:convol} and Figure~\ref{fig:convolution}. 
Without surprise we can observe that atomic patches are not discriminative. 
Recursive aggregation of patches in subsequent stages will be the key to create discriminative responses.

\begin{figure}
 \centering
 \includegraphics[width=0.6\linewidth]{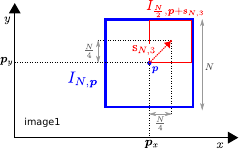}
 \caption{A patch $I_{N,\bm{p}}$ from the first image (blue box) and one of its 4 quadrants $I_{\frac{N}{2},\bm{p}+\frac{N}{4}\bm{o}_3}$ (red box).}
 \label{fig:patchsplit}
\end{figure}

\newcommand{\ch}[2]{\mathrm{s}_{#1,#2}}

\paragraph{Iteration.}
We then compute the correlation maps of larger patches by aggregating those of smaller patches.
As shown in Figure~\ref{fig:patchsplit}, a $N\times N$ patch $I_{N,\bm{p}}$ is the concatenation of $4$ patches 
of size $N/2 \, \times N/2$: 
\begin{equation}
I_{N,\bm{p}} = \left[I_{\frac{N}{2},\bm{p}+\frac{N}{4}\bm{o}_i}\right]_{i=0..3} 
\mathrm{\ \ with\ }
\scalebox{0.7}{$
\begin{cases}
\bm{o}_0 = [-1,-1]^\top, \\
\bm{o}_1 = [-1,+1]^\top, \\
\bm{o}_2 = [+1,-1]^\top, \\
\bm{o}_3 = [+1,+1]^\top.
\end{cases}$
}
\label{eqn:childpos}
\end{equation}
They correspond respectively to the bottom-left, top-left, bottom-right and top-right quadrants.
The correlation map of $I_{N,\bm{p}}$ can thus be computed using its children's correlation maps.
\red{For the sake of clarity, we define the short-hand notation $\ch{N}{i} = \frac{N}{4}\bm{o}_i$
describing the positional shift of a children patch $i\in[0,3]$ 
relatively to its parent patch (see Figure~\ref{fig:patchsplit}).}

\red{Using the above notations, we rewrite \eq\refp{eqn:quadrant}
by replacing $\mathrm{sim}(\R,\R')\stackrel{def}{=}C_{N,\bm{p}}(\bm{p}')$
(\ie assuming here that patch $\mathbf{R}=I_{N,\boldsymbol{p}}$ and that
$\mathbf{R}'$ is centered at $\boldsymbol{p}'\in I'$). Similarly,
we replace the similarity between children patches $\mathrm{sim}\left(\R_{i},\R_{i}'(\bm{p}'_{i})\right)$ by 
$C_{\frac{N}{2},\bm{p}+\ch{N}{i}}(\bm{p}_{i}')$.
For each child, we retain the maximum similarity over a small neighborhood
$\Theta_{i}$ of width and height $\frac{N}{8}$ 
centered at $\boldsymbol{p}'+\ch{N}{i}$. 
We then obtain: 
\begin{equation}
C_{N,\bm{p}}(\bm{p}') = \frac{1}{4}\sum_{i=0}^{3}\max_{\boldsymbol{m}'\in\Theta_{i}}\;C_{\frac{N}{2},\bm{p}+\ch{N}{i}}(\bm{m}')
\label{eqn:quadrant2}
\end{equation}
}

\red{We now explain how we can break down \eq\refp{eqn:quadrant2} into a succession
of simple operations. First, let us assume that $N=4\times2^\ell$,
where $\ell \geq 1$ is the current iteration. During iteration $\ell$,
we want to compute the correlation maps $C_{N,\bm{p}}$ of every patch $I_{N,\bm{p}}$ 
from the first image for which correlation maps of its children have been 
computed in the previous iteration. Formally, the position $\mathcal{G}_{N}$ of such patches is defined 
according to the position of children patches $\mathcal{G}_{\frac{N}{2}}$ according to \eq\refp{eqn:childpos}:}
\begin{eqnarray}
 \mathcal{G}_N = \left\{ \bm{p} \right. & \;|\; & \bm{p}+\ch{N}{i} \in [0,W-1]\times[0,H-1] \wedge \nonumber \\ 
                                &       & \left. \bm{p}+\ch{N}{i} \in \mathcal{G}_{\frac{N}{2}}, i=0,\ldots,3\right\}.
 \label{eqn:patchdef}
\end{eqnarray}

\begin{figure*}
  \centering
  \includegraphics[width=1\textwidth]{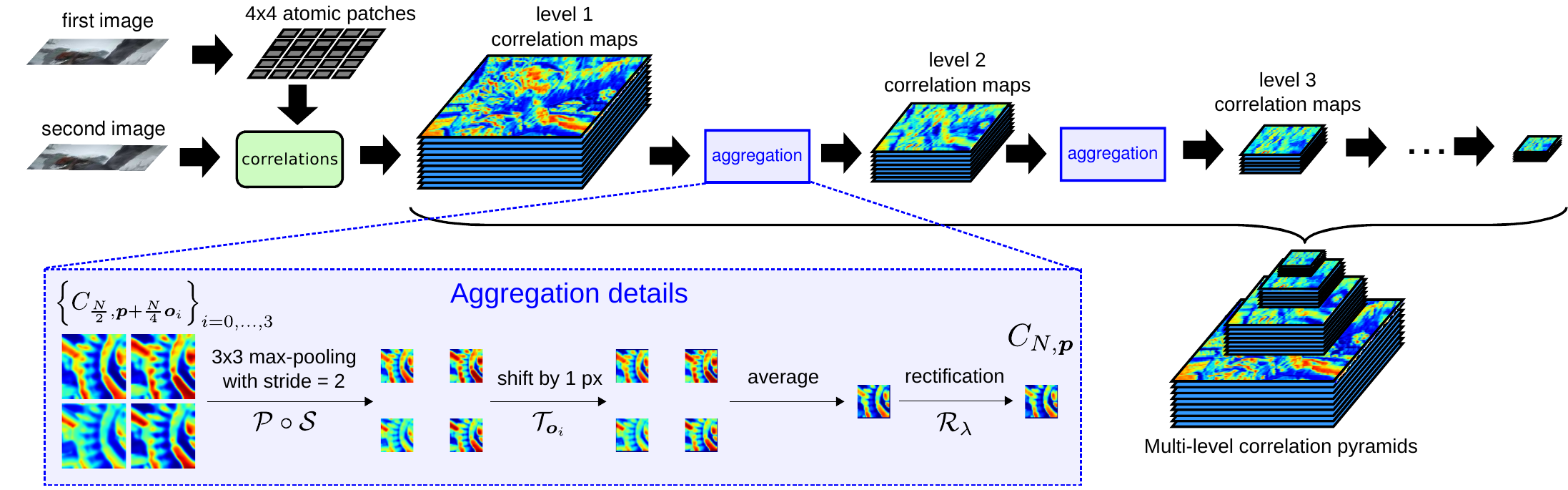}
  \caption{Computing the multi-level correlation pyramid. Starting with the bottom-level
    correlation maps, see Figure~\ref{fig:convol}, they are
    iteratively aggregated to obtain the upper levels. 
    Aggregation consists of max-pooling, subsampling, computing a
    shifted average and non-linear rectification.} 
  \label{fig:global}
\end{figure*}

\noindent We observe that the larger a patch is (\ie after several
iterations), the smaller the spatial variation of its correlation map (see
Figure~\ref{fig:convolution}). This is due to the statistics of
natural images, in which low frequencies significantly dominate over
high frequencies. 
\red{As a consequence, we choose to subsample each map $C_{N,\bm{p}}$ by a factor 2.}
We express this with an operator $\mathcal{S}$:
\begin{equation}
 \mathcal{S}: C(\bm{p}') \rightarrow C(2\bm{p}').
 \label{eqn:subsample}
\end{equation}
The subsampling reduces by $4$ the \red{area} of the correlation maps and, 
as a direct consequence, the computational requirements.
\red{Instead of computing the subsampling on top of \eq\refp{eqn:quadrant2}, it is actually more efficient
to propagate it towards the children maps and perform it jointly with max-pooling.}
It also makes the max-pooling domain $\Theta_i$ become independent from $N$ in the subsampled maps, as
it exactly cancels out the effect of doubling $N=4\times2^\ell$ at each iteration.
We call $\mathcal{P}$ the max-pooling operator with the iteration-independent domain $\Theta=\{-1,0,1\}\times\{-1,0,1\}$:
\begin{equation}
 \mathcal{P}: C(\bm{p}') \rightarrow \max_{\bm{m} \in \{-1,0,1\}^2} C(\bm{p}'+\bm{m}).
 \label{eqn:maxpool}
\end{equation}
For the same reason, the shift $\ch{N}{i}=\frac{N}{4}\bm{o}_i=2^\ell \bm{o}_i$ applied to the correlation maps 
in $\Theta_i$'s definition becomes simply $\bm{o}_i$ after subsampling.
Let $\mathcal{T}_{\bm{t}}$ be the shift (or translation) operator on
the correlation map:
\begin{equation}
 \mathcal{T}_{\bm{t}}: C(\bm{p}') \rightarrow C(\bm{p}'-\bm{t}).
 \label{eqn:shift}
\end{equation}
Finally, we incorporate an additional non-linear mapping at each iteration on top of 
\eq\refp{eqn:quadrant2} by applying a power transform $\mathcal{R}_{\lambda}$ \citep{MalikPerona90,LeCun98}:
\begin{equation}
\mathcal{R}_{\lambda}: C(\bm{.}) \rightarrow C(\bm{.})^\lambda
\label{eqn:nlpow}
\end{equation}
This step, commonly referred to as rectification, is added in order to better propagate high correlations after each level, 
or, in other words, to counterbalance the fact that max-pooling tends to retain only high scores.
\red{Indeed, its effect is to decrease the correlation values (which are in $[0,1]$)
as we use $\lambda>1$.}
Such post-processing is commonly used in deep convolutional networks \citep{lecun-98b,Bengio09}.
In practice, good performance is obtained with $\lambda\simeq1.4$, see Section~\ref{sec:xp}. 
The final expression of \eq\refp{eqn:quadrant2} is:
\begin{equation}
 C_{N,\bm{p}} = \mathcal{R}_{\lambda}\left(\frac{1}{4} \sum_{i=0}^3 \left(\mathcal{T}_{\bm{o}_i} \circ \mathcal{S} \circ \mathcal{P}\right)
 \left( C_{\frac{N}{2},\bm{p}+\ch{N}{i}}\right)\right) 
 \label{eqn:recurr}
\end{equation}

Figure~\ref{fig:global} illustrates the computation of correlation
maps for different patch sizes and Algorithm~\ref{alg:resmap}
summarizes our approach. The resulting set of correlation maps across
iterations is referred to as multi-level correlation pyramid.

\paragraph{Boundary effects:\:}
\red{In practice, a patch $I_{N,\bm{p}}$ 
can overlap with the image boundary, as long as its center $\bm{p}$ remains inside the image
(from \eq\refp{eqn:patchdef}). 
For instance, a patch $I_{N,\bm{p}_0}$ with center at $\bm{p}_0=(0,0) \in \mathcal{G}_N$ 
has only a single valid child (the one for which $i=3$ as $\bm{p}_0 + \ch{N}{3}\in I$).
In such degenerate cases, the average sum in \eq\refp{eqn:recurr} is carried out on valid
children only. For $I_{N,\bm{p}_0}$, it thus only comprises one term weighted by 1 instead of $\frac{1}{4}$.
}

Note that \eq\refp{eqn:recurr} implicitly defines the set of
possible displacements of the approach, see Figures~\ref{fig:recur} and \ref{fig:limit}.
Given the position of a parent patch, each child patch can move only within a small extent, equal to the quarter of its own size.
Figure~\ref{fig:convolution} shows the correlation maps for patches of size $4$, $8$ and $16$. 
Clearly, correlation maps for larger patch \red{are} more and more discriminative, 
while still allowing non-rigid matching.

\begin{algorithm}
\textbf{Input}: Images $I$, $I'$ 

\textbf{For} $\bm{p} \in\mathcal{G}_4$ \textbf{ do}

$\quad$$C_{4,\bm{p}} = I_{4,\bm{p}}^\mathrm{F} \star I'$ (convolution, \eq\refp{eqn:init})

$\quad$$C_{4,\bm{p}} \leftarrow \mathcal{R}_{\lambda} (C_{4,\bm{p}})$ (rectification, \eq\refp{eqn:nlpow})

$N\leftarrow4$

\textbf{While} \red{$N < \max(W,H)$} \textbf{do}

$\quad$\textbf{For} $\bm{p} \in\mathcal{G}_N$ \textbf{ do}

$\quad\quad$$C'_{N,\bm{p}} \leftarrow (\mathcal{S}\circ\mathcal{P})(C_{N,\bm{p}})$ (\red{max-pooling and subsampling})

$\quad N\leftarrow2N$

$\quad$\textbf{For} $\bm{p} \in\mathcal{G}_N$ \textbf{do}

$\quad$$\quad$$C_{N,\bm{p}} = \frac{1}{4} \sum_{i=0}^3 \mathcal{T}_{\bm{o}_i} \left(C'_{\frac{N}{2},\bm{p}+\ch{N}{i}} \right)$ (shift and average)

$\quad$$\quad$$C_{N,\bm{p}} \leftarrow \mathcal{R}_{\lambda} (C_{N,\bm{p}})$ (rectification, \eq\refp{eqn:nlpow})

\textbf{Return} the multi-level correlation pyramid $\left\{ C_{N,\bm{p}} \right\}_{N,\bm{p}}$

\caption{Computing the multi-level correlation pyramid.}
\label{alg:resmap} 
\end{algorithm}

\subsection{Top-down correspondence extraction} 
\label{sub:backtrackcorrespondences}

A score $S = C_{N,\bm{p}}(\bm{p}')$ in the multi-level correlation pyramid 
represents the deformation-tolerant similarity of two patches $I_{N,\bm{p}}$ and $I_{N,\bm{p'}}'$.
Since this score is built from the similarity of 4 matching sub-patches at the lower pyramid level, 
we can thus recursively backtrack a set of correspondences to the bottom level (corresponding to matches of atomic patches).
In this section, we first describe this backtracking.
We, then, present the procedure for merging atomic correspondences
backtracked from different entry points in the multi-level pyramid,
which constitute the final output of DeepMatching. 

Compared to our initial version of DeepMatching \citep{DeepFlow},
we have updated match scoring and entry point selection to optimize the execution time
and the matching accuracy. A quantitative comparison is provided in Section~\ref{sub:oldvsnew}.

\paragraph{Backtracking atomic correspondences.}

\begin{figure*}
 \centering
 \includegraphics[width=0.8\textwidth]{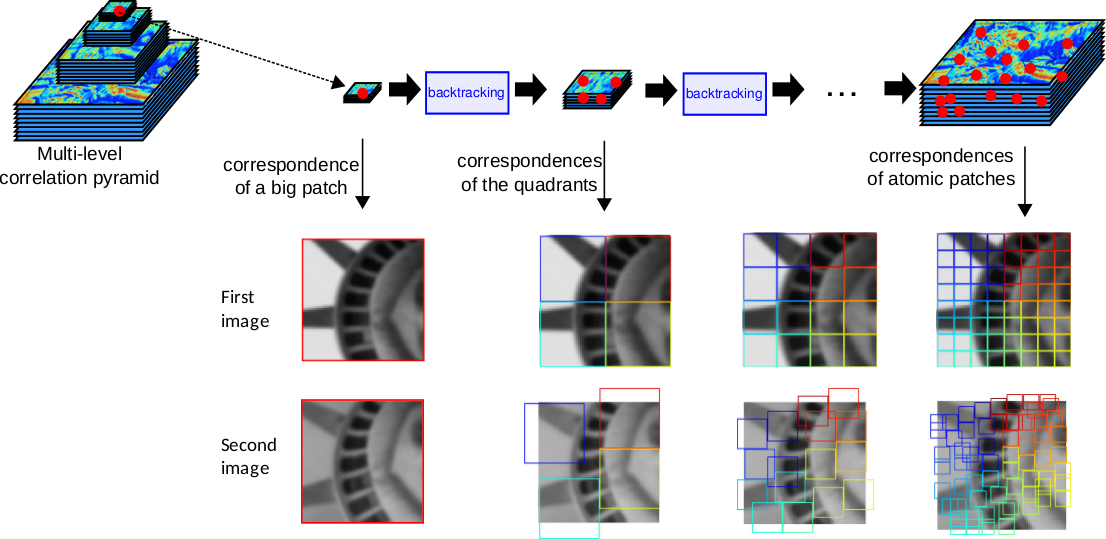}
 \caption{\red{
  Backtracking atomic correspondences from an entry point (red dot) in the top pyramid level \emph{(left)}. 
  At each level, the backtracking consists in undoing the aggregation performed previously in order 
  to recover the position of the four children patches in the lower level.
  When the bottom level is reached, we obtain a set of correspondences for atomic patches \emph{(right)}.
 }}
 \label{fig:undo}
\end{figure*}

Given an entry point $C_{N, \bm{p}}(\bm{p}')$ in the pyramid 
(\ie a match between two patches $I_{N,\bm{p}}$ and $I'_{N,\bm{p}'}$\footnote{
Note that $I'_{N,\bm{p}'}$ only roughly corresponds to a $N \times N$ 
square patch centered at $2^\ell \bm{p}'$ in $I'$, due to subsampling and possible deformations.}),
we retrieve atomic correspondences by successively undoing the steps 
used to aggregate correlation maps during the pyramid construction, see Figure~\ref{fig:undo}.
The entry patch $I_{N,\bm{p}}$ is itself composed of four moving quadrants $I^i_{N,\bm{p}}$, $i=0\ldots3$.
Due to the subsampling, the quadrant $I^i_{N,\bm{p}} = I_{\frac{N}{2},\bm{p}+\ch{N}{i}}$
matches with $I_{\frac{N}{2},2(\bm{p}'+\bm{o}_i)+\bm{m}_i}$ where
\begin{equation}
 \bm{m}_i = \argmax_{\bm{m} \in \{ -1, 0, 1 \}^2}
    C_{\frac{N}{2},\bm{p}+\ch{N}{i}} \left( 2(\bm{p}'+\bm{o}_i)+\bm{m} \right).
 \label{eqn:undomax}
\end{equation}

For the sake of clarity, we define the short-hand notations $\bm{p}_i = \bm{p}+\ch{N}{i}$ and $\bm{p}'_i = 2 (\bm{p}' + \bm{o}_i) + \bm{m}_i$.
Let $B$ be the function that assigns to a tuple $(N, \bm{p}, \bm{p}', s)$, representing a correspondence between pixel $\bm{p}$ and $\bm{p}'$ for patch of size $N$
with a score $s \in \mathbb{R}$, the set of the correspondences of children patches:
{\small{}
\begin{equation}
B(N, \bm{p}, \bm{p}', s) = 
 \begin{cases}
  \left\{ \left( \bm{p}, \bm{p}', s \right) \right\} & \text{ if $N=4$,}
  \\
  \left\{ \left( \frac{N}{2}, \bm{p}_i, \bm{p}'_i, 
     s+C_{\frac{N}{2}, \bm{p}_i}(\bm{p}'_i) \right)\right\}_{i=0}^3&  \text{ else.}
 \end{cases}
 \label{eqn:backtrack}
\end{equation}
}
Given a set $\mathcal{M}$ of such tuples, let $\mathcal{B}(\mathcal{M})$ be the union of the sets $B(c)$
for all $c \in \mathcal{M}$.
Note that if all candidate correspondences $c \in \mathcal{M}$ corresponds to atomic patches,
then $\mathcal{B}(\mathcal{M}) = \mathcal{M}$.

Thus, the algorithm for backtracking correspondences is the following.
Consider an entry match $\mathcal{M}=\{\left(N, \bm{p}, \bm{p}', \right.$ $\left.C_{N,\bm{p}}(\bm{p}')\right)\}$. 
We repeatedly apply $\mathcal{B}$ on $\mathcal{M}$. 
After $\mathfrak{N} = \log_2(N/4)$ calls, 
we get one correspondence for each of the $4^{\mathfrak{N}}$ atomic patches.
Furthermore, their score is equal to the sum of all patch similarities along their backtracking path.

\paragraph{Merging correspondences.}
\label{sub:merge}

We have shown how to retrieve atomic correspondences 
from a match between two deformable (potentially large) patches.
Despite this flexibility, a single match is unlikely to explain  
the complex set of motions that can occur, for example, between two adjacent frames in a video, \ie,
two objects moving independently with significantly different motions exceeds the deformation range of DeepMatching. 
We quantitatively specify this range in the next subsection. 

We thus merge atomic correspondences gathered from different entry points (matches) in the pyramid. 
In the initial version of DeepMatching \citep{DeepFlow}, entry points
were local maxima over all correlation maps. This is now replaced by 
a faster procedure, that starts with all possible matches in the top
pyramid level  
(\ie $\mathcal{M}=\{(N, \bm{p}, \bm{p}',$ $C_{N,\bm{p}}(\bm{p}')) | N=N_{\text{max}}\}$).
Using this level only results in significantly less entry points than
starting from all maxima in the entire pyramid.
We did not observe any impact on the matching performance, see Section~\ref{sub:oldvsnew}.
Because $\mathcal{M}$ contains a lot of overlapping patches, most of the computation
during repeated calls to $\mathcal{M}\leftarrow\mathcal{B}(\mathcal{M})$ can be factorized. In other words, as soon as two tuples in $\mathcal{M}$
are equal in terms of $N$, $\bm{p}$ and $\bm{p}'$, 
the one with the lowest score is simply eliminated.
We thus obtain a set of atomic correspondences $\mathcal{M}'$:
\begin{equation}\label{eqn:atomiccorres}
\mathcal{M}' = (\mathcal{B} \circ \ldots \circ \mathcal{B})(\mathcal{M})
\end{equation}
that we filter with reciprocal match verification. 
The final set of correspondences $\mathcal{M}''$ is obtained as: 
\begin{equation}\label{eqn:reciprocal}
 \mathcal{M}'' = \left\{ (\bm{p}, \bm{p}',s) | \mathrm{BestAt}(\bm{p}) = \mathrm{BestAt}'(\bm{p}')\right\}_{(\bm{p},\bm{p}',s)\in \mathcal{M}'}
\end{equation}
where $\mathrm{BestAt}(\bm{p})$ (resp.  $\mathrm{BestAt}'(\bm{p}')$) returns the best match
in a small vicinity of $4\times 4$ pixels around $\bm{p}$ in $I$ (resp. around $\bm{p}'$ in $I'$)  from $\mathcal{M}'$.

\subsection{Discussion and Analysis of DeepMatching}
\label{sub:analysis}

\paragraph{Multi-size patches and repetitive textures.}

\begin{figure}
  \centering
  \includegraphics[bb=100bp 0bp 1820bp 708bp,clip,width=1\linewidth]{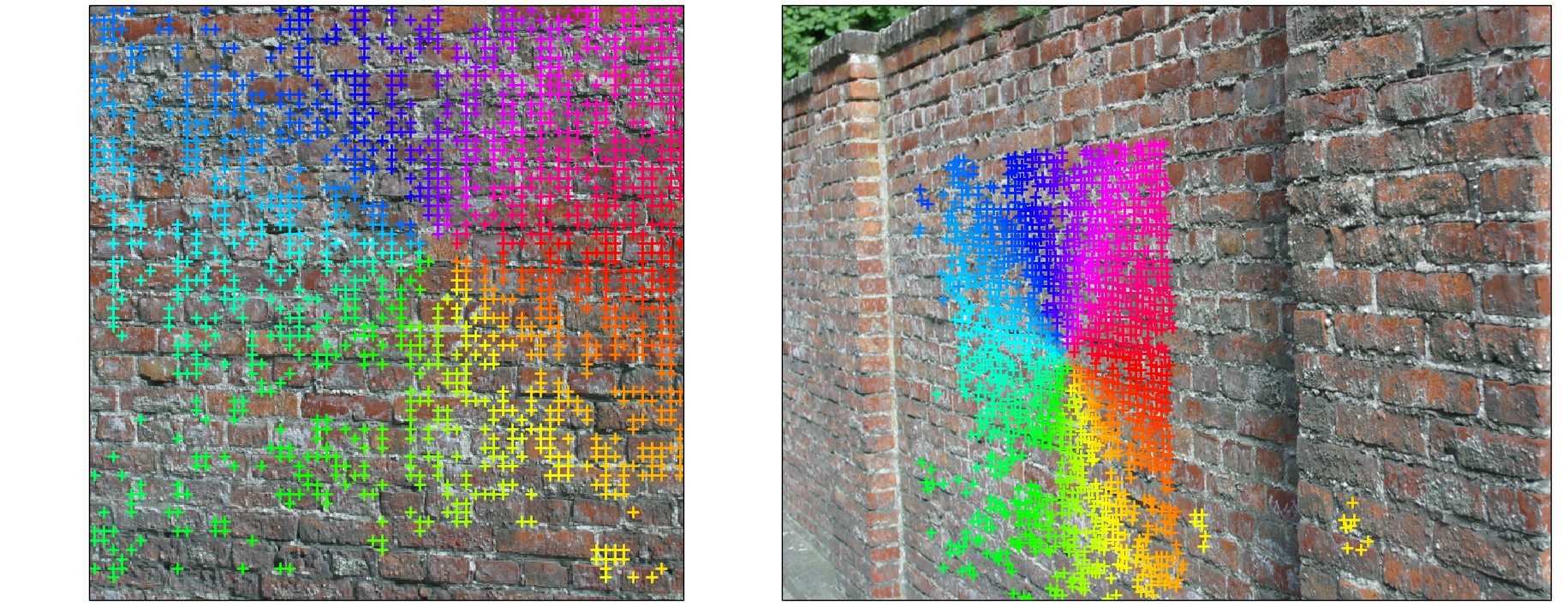}
  \caption{Matching result between two images with repetitive
    textures. \red{Nearly all output correspondences are correct. 
    Wrong matches are due to occluded areas (bottom-right of the first image) or situations where the 
    deformation tolerance of DeepMatching is exceeded (bottom-left of the first image).} }
  \label{fig:repet}
\end{figure}

During the bottom-up pass of the algorithm, we iteratively aggregate
correlation maps of smaller patches to form the correlation maps of
larger patches. Doing so, we effectively consider patches of 
different sizes ($4 \times 2^{\ell}, \ell\geq0$), in contrast to most existing matching methods. 
This is a key feature of our approach when
dealing with repetitive textures. As one moves up to upper levels,
the matching problem gets less ambiguous. Hence, 
our method can correctly match repetitive patterns, see for instance Figure~\ref{fig:repet}.

\paragraph{Quasi-dense correspondences.}
Our method retrieves dense correspondences for every single match
between large regions (\ie entry point for the backtracking in the top-level correlation maps), 
even in weakly textured areas; this is in contrast to 
correspondences obtained when matching descriptors (\eg SIFT).  
A quantitative assessment, which compares the coverage
of matches obtained with several matching schemes, is given in Section~\ref{sec:xp}.

\begin{figure}
  \centering
  \includegraphics[bb=70bp 300bp 700bp 500bp,clip,width=1\linewidth]{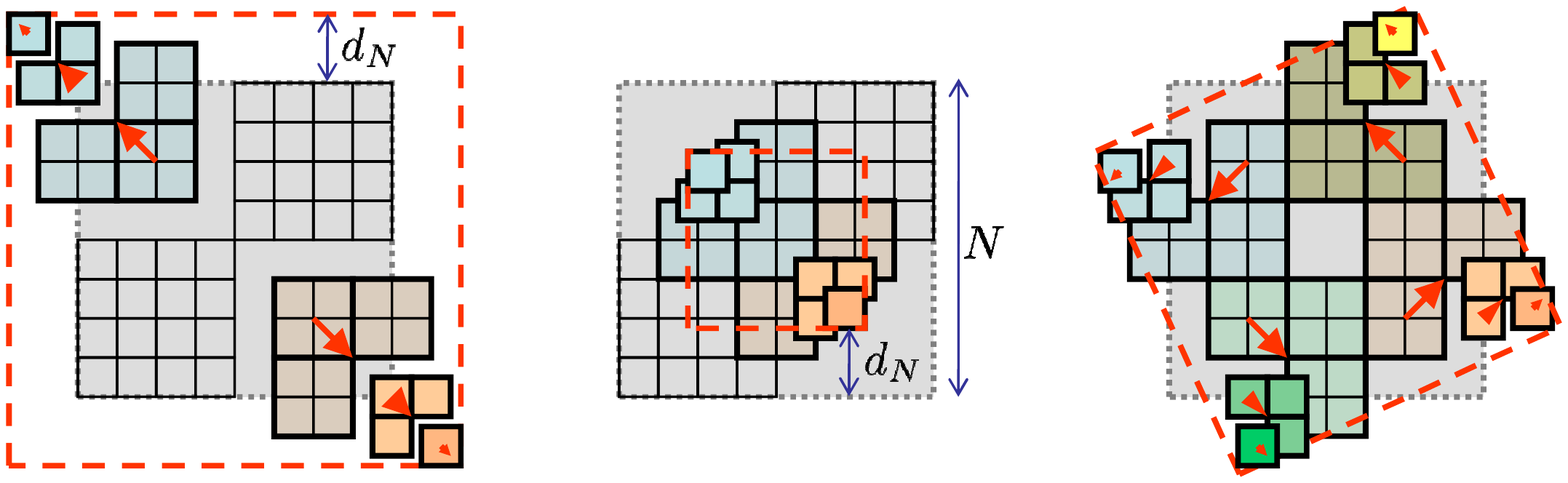}
  \caption{Extent of the tolerance of DeepMatching to deformations. From left
    to right: up-scale of 1.5x, down-scale of 0.5x, rotation of 26\textsuperscript{o}.
    The plain gray (resp. dashed red) square represents the patch in the reference
    (resp. target) image. For clarity, only the corner pixels are maximally deformed.}
  \label{fig:limit}
\end{figure}

\paragraph{Non-rigid deformations.}
Our matching algorithm is able to cope with various sources
of image deformations: object-induced or camera-induced. The set of
feasible deformations, explicitly defined by \eq\refp{eqn:quadrant2}, theoretically
allows to deal with a scaling factor in the range $[\frac{1}{2},\frac{3}{2}]$
and rotations approximately in the range $[-26^{o},26^{o}]$. 
Note also that DeepMatching is translation-invariant by construction,
thanks to the convolutional nature of the processing. 
\begin{proof}
Given a patch of size $N=4\times 2^{\ell}$ located at level $\ell\geqslant1$,
\eq\refp{eqn:quadrant2} allows each of its children patches to
move by at most $N/8$ pixels from their ideal location in $\Theta_i$. By recursively
summing the displacements at each level, the maximal displacements
for an atomic patch is $d_{N}=\sum_{i=1}^{\ell}2^{i-1}=2^{\ell}-1$.
An example is given in Figure~\ref{fig:limit} with $N=32$ and $\ell=3$.
Relatively to $N$, we thus have $\lim_{N\rightarrow\infty} \; (N+2d_{N})/N=\frac{3}{2}$
and $\lim_{N\rightarrow\infty} \; (N-2d_{N})/N=\frac{1}{2}$. For
a rotation, the rationale is similar, see Figure~\ref{fig:limit}.
\qed
\end{proof}
\red{
Note that the displacement tolerance in $\Theta_i$ from \eq\refp{eqn:quadrant2} could 
be extended to $x \times N/8$ pixels with $x \in \{2,3,\ldots\}$ (instead of $x=1$). 
Then the above formula for computing the lower bound on the scale factor of DeepMatching 
generalizes to $\mathrm{LB}(x) = \lim_{N\rightarrow\infty} \; (N-2xd_{N})/N$. 
Hence, for  $x \geq 2$ we obtain $\mathrm{LB}(x)<0$ instead of  $\mathrm{LB}(1)=\frac{1}{2}$.
This implies that the deformation range is extended to a point where any patch can be matched to  a single pixel,  \ie, this results in unrealistic deformations. 
For this reason, we choose to not expand the deformation range of DeepMatching.
}

\begin{figure}
  \centering
  \includegraphics[bb=0bp 5bp 585bp 185bp,clip,width=0.9\linewidth]{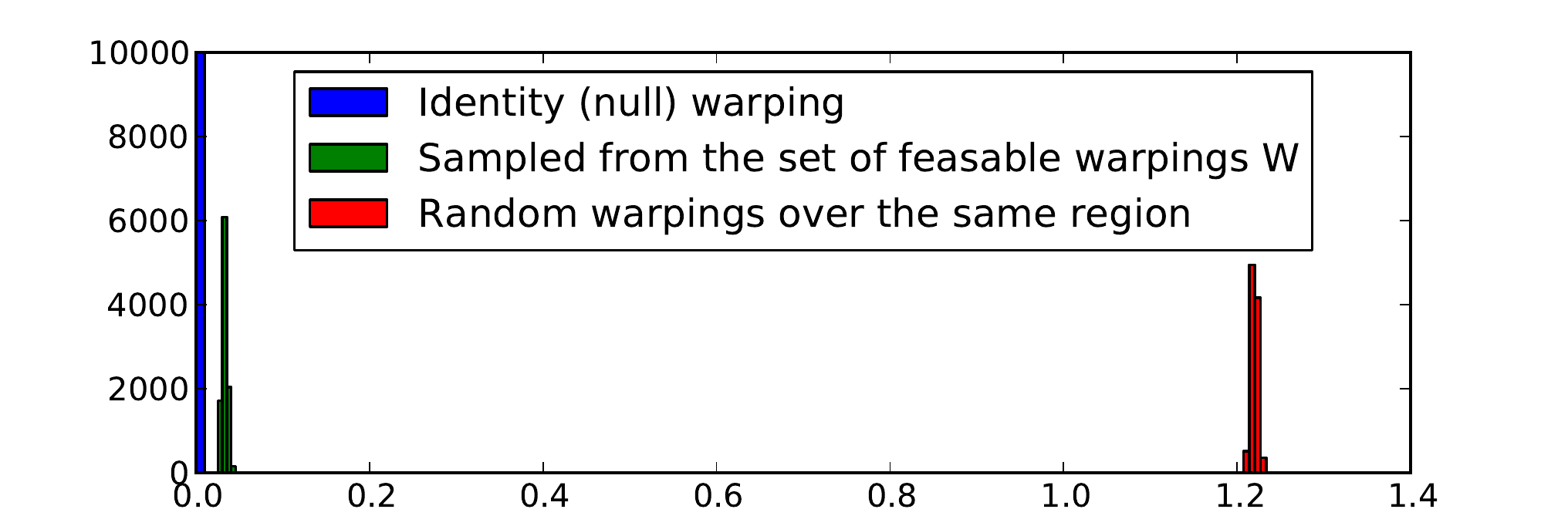}
  \caption{\red{
	Histogram over smoothness for identity warping, warping
        respecting the built-in constraints in DeepMatching and random
        warping. The x-axis indicates the smoothness value.
The smoothness value is low when there are few
        discontinuities, \ie, the warpings are smooth. The
        histogram is obtained with 10,000 different artificial warpings. 
See text for details. 
}}
    \label{fig:smooth}
\end{figure}

\paragraph{Built-in smoothing.}
Furthermore, correspondences generated through backtracking 
of a single entry point in the correlation maps are naturally smooth.
Indeed, feasible deformations cannot be too ``far'' from the identity deformation.
\red{To verify this assumption, we conduct the following experiment. 
We artificially generate two types of correspondences between two
images of size  $128 \times 128$.
The first one is completely random, \ie for each atomic patch in the
first image we assign randomly a match in the second image. 
The second one respects the backtracking constraints. 
Starting from a single entry point in the top level we simulate the
backtracking procedure from Section~\ref{sub:backtrackcorrespondences} 
by replacing in \eq\refp{eqn:undomax} the max operation by a random
sampling over $\{-1,0,1\}^2$. By generating 10,000 sets of possible
atomic correspondences, we simulate a set which respects the 
deformations allowed by DeepMatching.
} 
Figure~\ref{fig:smooth} compares the
smoothness of these two types of artificial correspondences. 
\red{
Smoothness is measured by interpreting the correspondences as 
flow and measuring the gradient flow norm, see
\eq\refp{eqn:smoothterm}. }
Clearly, the two types of warpings are different by orders of
magnitude. Furthermore, the one which respects the built-in constraints of
DeepMatching is close to the identity warping.

\paragraph{Relation to Deep Convolutional Neural Networks (CNNs).}
\red{
DeepMatching relies on a hierarchical, multi-layer, correlational architecture designed for matching images and  
was inspired by deep convolutional approaches~\citep{LeCun98}. 
In the following we describe the major similarities and differences.

Deep networks learn from data the weights of the convolutions. 
In contrast, DeepMatching does not learn any feature representations
and instead directly computes correlations at the patch level.
It uses patches from the first image as convolution
filters for the second one.
However, the bottom-up pipeline of DeepMatching is similar to CNNs. 
It alternates aggregating channels from the previous layer
with channel-wise max-pooling and subsampling.
As in CNNs, max-pooling in DeepMatching allows for 
invariance \wrt small deformations.
Likewise, the algorithm propagates pairwise patch similarity scores
through the 
hierarchy using non-linear rectifying stages in-between layers. 
Finally, DeepMatching includes a top-down pass which is not present in CNNs.
}

\paragraph{Time and space complexity.}
DeepMatching has a complexity $O(LL')$ in memory and time, 
where $L=WH$ and $L'=W'H'$ are the number of pixels per image.
\begin{proof}
Computing the initial correlations is a $O(LL')$ operation. Then,
at each level of the pyramid, the process is repeated while the
complexity is divided by a factor $4$ due to the subsampling step
in the target image (since the cardinality of $|\{\mathcal{G}_N\}|$ 
remains approximately constant). Thus, the total complexity of the correlation maps
computation is, at worst, $O(\sum_{n=0}^{\infty}LL'/4^{n})=O(LL')$.
During the top-down pass, most backtracking paths can be pruned as soon as they cross
a concurrent path with a higher score (see Section~\ref{sub:backtrackcorrespondences}).
Thus, all correlations will be examined
at most once, and there are $\sum_{n=0}^{\infty}LL'/4^{n}$ values
in total. However, this analysis is worst-case. In practice, only correlations lying
on maximal paths are actually examined.
\qed
\end{proof}

\section{Extensions of DeepMatching}
\label{sec:extensions}

\subsection{Approximate DeepMatching}
\label{sub:compr}

As a consequence of its $O(LL')$ space complexity,
DeepMatching requires an amount of RAM that is 
orders of magnitude above other state-of-the-art matching methods.
This could correspond to several gigabytes for images of moderate size (800$\times$600 pixels); see Section~\ref{sub:approxxps}. 
This section introduces an approximation of DeepMatching that allows
to trade matching quality for reduced time and memory usage. 
As shown in Section~\ref{sub:approxxps}, near-optimal
results can be obtained at a fraction of the original cost.

Our approximation proposes to
compress the representation of atomic patches $\{I_{4,\bm{p}}\}$. 
Atomic patches carry little information, and thus
are highly redundant. For instance, in uniform regions, all patches
are nearly identical (\ie, gradient-wise).
To exploit this property, we index atomic
patches with a small set of patch prototypes. We substitute
each patch with its closest neighbor in a fixed dictionary of $D$ prototypes.
Hence, we need to perform and store only $D$ convolutions at the first level, instead of $O(L)$
(with $D\ll O(L)$). This significantly reduces both memory and time complexity. 
Note that higher pyramid levels also
benefit from this optimization. Indeed, two parent
patches at the second level have the exact same correlation map in case their
children are assigned the same prototypes.
The same reasoning also holds for all subsequent
levels, but the gains rapidly diminish due to
statistical unlikeliness of the required condition. This is not really an issue,
since the memory and computational cost mostly rests on the initial 
levels; see Section~\ref{sub:analysis}.

In practice, we build the prototype dictionary using k-means, as it
is designed to minimize the approximation error between
input descriptors and resulting centroids (\ie prototypes). Given
a pair of images to match, we perform on-line clustering of all descriptors of 
atomic patches $\{I_{4,\bm{p}}\}=\{\R\}$  in the first image. Since  the original descriptors
lie on an hypersphere (each pixel descriptor $\R_{i,j}$ has norm 1),
we modify the k-means approach  so as to project the estimated centroids on the hypersphere at each iteration.
We find experimentally that this is important to obtain good results.

\subsection{Scale and rotation invariant DeepMatching}
\label{sub:scalerot}

For a variety of tasks, objects to be matched can appear under image
rotations or at different scales~\citep{Lowe2004,Mikolajczyk2005,Szeliski2010,HaCohen2011}.
As discussed above, DeepMatching (DM) is only robust to moderate
scale changes and rotations. We now present a scale and rotation invariant version.

The approach is straightforward: 
we apply DM to several rotated and scaled versions of the second
image. According to the invariance range of DM, we use steps of
$\pi/4$ for image rotation and power of $\sqrt{2}$ for scale
changes. While iterating over all combinations of scale changes and
rotations, we maintain a list 
$\mathcal{M}'$ of
all atomic correspondences obtained so far, \ie corresponding positions and scores.
As before, the final output correspondences consists of the reciprocal
matches in $\mathcal{M}'$. 
Storing all matches and finally choosing the best ones based on reciprocal verification
permits to capture distinct motions
possibly occurring together in the same scene (\eg one object could have undergone a rotation, while the
rest of the scene did not move). 
The steps of the approach are described in Algorithm \ref{alg:IDM}.

Since we iterate sequentially over a fixed list of rotations and scale changes,
the space and time complexity of the algorithm remains unchanged (\ie $O(LL')$).
In practice, the run-time compared to DM is multiplied by a constant approximately equal to 25, see Section~\ref{sub:dmsota}. 
Note that the algorithm permits a straightforward parallelization. 

\begin{algorithm}
\textbf{Input}: $I$, $I'$ are the images to be matched

\textbf{Initialize} an empty set $\mathcal{M}'=\left\{\right\}$ of correspondences 

\textbf{For} $\sigma\in\left\{ -2,-1.5\ldots,1.5,2\right\}$ \textbf{do}

$\quad$$\sigma_1\leftarrow\max\left(1,2^{+\sigma}\right)$ 
\textcolor{green}{\# either downsize image 1}

$\quad$$\sigma_2\leftarrow\max\left(1,2^{-\sigma}\right)$ 
\textcolor{green}{\# or downsize image 2}

$\quad$\textbf{For }$\theta\in\left\{ 0,\frac{\pi}{4}\ldots,\frac{7\pi}{4}\right\} $
\textbf{do}

$\quad$$\quad$\textcolor{green}{\# get raw atomic correspondences (\eq\refp{eqn:atomiccorres})}

$\quad$$\quad$$\mathcal{M}'_{\sigma,\theta}\leftarrow\mathrm{DeepMatching}\left(I_{\sigma_1},\mathcal{R}_{-\theta}*I'_{\sigma_2}\right)$

$\quad$$\quad$\textcolor{green}{\# Geometric rectification to the input image space:}

$\quad$$\quad$$\mathcal{M}'^R_{\sigma,\theta}\leftarrow\left\{ \left(\sigma_1 \bm{p},\,\sigma_2\mathcal{R}_{\theta}\bm{p}',\,s\right)\,|\,\forall(\bm{p},\bm{p}',s)\in \mathcal{M}'_{\sigma,\theta}\right\} $

$\quad$$\quad$\textcolor{green}{\# Concatenate results:}

$\quad$$\quad$$\mathcal{M}' \leftarrow \mathcal{M}' \bigcup \mathcal{M}'^R_{\sigma,\theta}$

$\mathcal{M}'' \leftarrow \mathrm{reciprocal}(\mathcal{M}')$ \textcolor{green}{\# keep reciprocal
matches (\eq\refp{eqn:reciprocal})}

\textbf{Return} $\mathcal{M}''$\\

\caption{Scale and rotation invariant version of DeepMatching (DM). 
$I_{\sigma}$ denotes the image $I$ downsized by a factor $\sigma$, 
and $\mathcal{R}_{\theta}$ denotes rotation by an angle $\theta$.}

\label{alg:IDM} 
\end{algorithm}

\subsection{DeepFlow}
\label{sec:flow}

We now present our approach for optical flow estimation, DeepFlow.
\red{We adopt the method introduced by \cite{Bro11a}, where a matching term penalizes the differences between optical flow and input matches, and replace their matching approach by DeepMatching.
In addition, we make a few minor modifications introduced recently in the state of the art:
(i) we add a normalization in the data term to
downweight the impact of locations with high spatial image derivatives \citep{Zimmer2011};
(ii) we use a different weight at each level to downweight the matching
term at finer scales \citep{Stoll2012}; and (iii) the smoothness term is locally weighted \citep{mdpof}.}

Let $I_{1},I_{2}:\Omega\rightarrow\mathbb{R}^{c}$ be two consecutive
images defined on $\Omega$ with $c$ channels. The goal is to estimate
the flow $\bm{w}=(u,v)^{\top}:\Omega\rightarrow\mathbb{R}^{2}$. We
assume that the images are already smoothed using a Gaussian filter
of standard deviation $\sigma$. The energy we optimize is a weighted
sum of a data term $E_{D}$, a smoothness term $E_{S}$ and a matching
term $E_{M}$: 
\begin{equation}\label{eqn:variational}
E(\bm{w})=\int_{\Omega}E_{D}+\alpha E_{S}+\beta E_{M}\bm{dx}
\end{equation}
For the three terms, we use a robust penalizer $\Psi(s^{2})=\sqrt{s^{2}+\epsilon^{2}}$
with $\epsilon=0.001$ which has shown excellent results \citep{Sun2014}.

\paragraph{Data term.} The data term is a separate penalization
of the color and gradient constancy assumptions with a normalization
factor as proposed by \cite{Zimmer2011}. We start from
the optical flow constraint assuming brightness constancy: $(\nabla_{3}^{\top}I)\bm{w}=0\textrm{ with }\nabla_{3}=(\partial x,\partial y,\partial t)^{\top}$ the spatio-temporal gradient.
A basic way to build a data term is to penalize it, \ie $E_{D}=\Psi(\bm{w}^{\top}\bm{J}_{0}\bm{w})$
with $\bm{J}_{0}$ the tensor defined by $\bm{J}_{0}=(\nabla_{3}I)(\nabla_{3}^{\top}I)$.
As highlighted by \cite{Zimmer2011},
such a data term adds a higher weight in locations corresponding to high spatial image derivatives.
We normalize it by the norm of the spatial derivatives plus a small
factor to avoid division by zero, and to reduce a bit the influence
in tiny gradient locations \citep{Zimmer2011}.
Let $\bm{\bar{J}}_{0}$ be the normalized tensor $\bm{\bar{J}}_{0}=\theta_{0}\bm{J}_{0}$
with $\theta_{0}=(\Vert\nabla_{2}I\Vert^{2}+\zeta^{2})^{-1}$. We
set $\zeta=0.1$ in the following. To deal with color images, we consider
the tensor defined for a channel $i$ denoted by upper indices $\bm{\bar{J}}_{0}^{i}$
and we penalize the sum over channels: $\Psi(\sum_{i=1}^{c}\bm{w}^{\top}\bm{\bar{J}}_{0}^{i}\bm{w})$.
We consider images in the RGB color space.

We separately penalize the gradient constancy assumption \citep{Bruhn2005}.
Let $I_{x}$ and $I_{y}$ be the derivatives of the images with respect
to the $x$ and $y$ axis respectively. Let $\bm{\bar{J}}_{xy}^{i}$
be the tensor for the channel $i$ including the normalization 
\begin{multline*}
\bm{\bar{J}}_{xy}^{i}=(\nabla_{3}I_{x}^{i})(\nabla_{3}^{\top}I_{x}^{i})/(\Vert\nabla_{2}I_{x}^{i}\Vert^{2}+\zeta^{2})\\+(\nabla_{3}I_{y}^{i})(\nabla_{3}^{\top}I_{y}^{i})/(\Vert\nabla_{2}I_{y}^{i}\Vert^{2}+\zeta^{2}) .
\end{multline*}
The data term is the sum of two terms, balanced by two weights $\delta$
and $\gamma$: 
\begin{equation}
E_{D}=\delta\Psi\left(\sum_{i=1}^{c}\bm{w}^{\top}\bm{\bar{J}}_{0}^{i}\bm{w}\right)+\gamma\Psi\left(\sum_{i=1}^{c}\bm{w}^{\top}\bm{\bar{J}}_{xy}^{i}\bm{w}\right)\label{eqn:dataterm}
\end{equation}

\paragraph{Smoothness term.} The smoothness term is a robust
penalization of the gradient flow norm: 
\begin{equation}
E_{S}=\Psi\left(\Vert\nabla u\Vert^{2}+\Vert\nabla v\Vert^{2}\right)~~.\label{eqn:smoothterm}
\end{equation}
\red{
The smoothness weight $\alpha$ is locally set according to image derivatives~\citep{Wedel2009,mdpof} with $\alpha(\bm{x}) = \exp(-\kappa \nabla_2 I(\bm{x}) )$ where $\kappa$ is experimentally set to $\kappa=5$.
}

\paragraph{Matching term.} The matching term encourages the
flow estimation to be similar to a precomputed vector field $\bm{w}'$.
To this end, we penalize the difference between $\bm{w}$ and $\bm{w}'$
using the robust penalizer $\Psi$. Since the matching is not totally
dense, we add a binary term $c(\bm{x})$ which is equal to $1$ if
and only if a match is available at $\bm{x}$.

We also multiply each matching penalization by a weight $\phi(\bm{x})$,
\red{which is low} in uniform regions where matching is ambiguous and when 
matched patches are dissimilar. 
To that aim, we rely on $\tilde{\lambda}(\bm{x})$, the minimum eigenvalue of
the autocorrelation matrix multiplied by $10$. We also compute the
visual similarity between matches as
$\Delta(\bm{x})=\sum_{i=1}^{c}\vert I_{1}^{i}(\bm{x})-I_{2}^{i}(\bm{x}-\bm{w}'(\bm{x}))\vert+\vert\nabla I_{1}^{i}(\bm{x})-\nabla I_{2}^{i}(\bm{x}-\bm{w}'(\bm{x}))\vert$.
We then compute the score $\phi$ as a Gaussian kernel on $\Delta$
weighted by $\tilde{\lambda}$ with a parameter $\sigma_{M}$, experimentally
set to $\sigma_{M}=50$. More precisely, we define $\phi(\bm{x})$
at each point $\bm{x}$ with a match $\bm{w}'(\bm{x})$ as: 
\begin{equation*}
\phi(\bm{x})=\sqrt{\tilde{\lambda}(\bm{x})}/(\sigma_{M}\sqrt{2\pi})\;\exp(-\Delta(\bm{x})/2\sigma_{M}).
\end{equation*}
The matching term is then 
$E_{M}=c\phi\Psi(\Vert\bm{w}-\bm{w}'\Vert^{2})$.

\paragraph{Minimization.} This energy objective is non-convex
and non-linear. To solve it, we use a numerical optimization algorithm similar as \cite{Bro04a}.
An incremental coarse-to-fine warping strategy is used with a downsampling
factor $\eta=0.95$. The remaining equations are still non-linear
due to the robust penalizers. We apply 5 inner fixed point iterations
where the non-linear weights and the flow increments are iteratively
updated while fixing the other. To approximate the solution of the
linear system, we use 25 iterations of the Successive Over Relaxation
(SOR) method~\citep{Young:sor}.

To downweight the matching term on fine scales, we use a different
weight $\beta^{k}$ at each level as proposed by \cite{Stoll2012}.
We set $\beta^{k}=\beta(k/k_{\text{max}})^{b}$ \red{where $k$ is the current level of computation, $k_{\max}$
the coarsest level and $b$ a parameter which is optimized together with the other parameters, see Section~\ref{subsub:flowparams}}.

\section{Experiments}
\label{sec:xp}

This section presents an experimental evaluation of DeepMatching and
DeepFlow. The datasets and metrics used to evaluate DeepMatching and
DeepFlow are introduced in Section \ref{sub:dbs}. Experimental results
are given in
Sections~\ref{sub:matchingxp} and~\ref{sub:flowxp} respectively.

\subsection{Datasets and metrics}
\label{sub:dbs}

In this section we briefly introduce the matching and flow datasets
used in our experiments. Since consecutive frames of a video are
well-suited to evaluate a matching approach, 
we use several optical flow datasets for evaluating both the quality
of matching and flow, but we rely on different metrics. 

\paragraph{The Mikolajczyk dataset}\ 
was originally proposed by \cite{Mikolajczyk2005}
to evaluate and compare the performance of keypoint detectors and
descriptors. It is one of the standard benchmarks for evaluating matching approaches. 
The dataset consists of 8 sequences of 6~images each viewing a scene under different
conditions, such as illumination changes or viewpoint changes. 
The images of a sequence are related by homographies. 
During the evaluation, we comply to the standard procedure in 
which the first image of each scene is matched to the 5 remaining ones.
Since our goal is to study robustness of DeepMatching to geometric
distortions, we follow \cite{HaCohen2011} and 
restrict our evaluation to the 4 most difficult sequences with
viewpoint changes: \emph{bark}, \emph{boat}, \emph{graf} and \emph{wall}.

\paragraph{The MPI-Sintel dataset \citep{sintel} }\ 
 is a challenging evaluation benchmark for optical flow estimation, 
 constructed from
realistic computer-animated films. The dataset contains sequences with
large motions and specular reflections. In the training set, more
than $17.5\%$ of the pixels have a motion over $20$ pixels, approximately
$10\%$ over $40$ pixels. We use the ``final'' version, featuring
rendering effects such as motion blur, defocus blur and atmospheric
effects. Note that ground-truth optical flows for the test set are
not publicly available.

\paragraph{The Middlebury dataset \citep{middlebury}}\ 
has been extensively used for evaluating optical flow methods.
The dataset contains complex motions, but most of the motions are small. Less
than $3\%$ of the pixels have a motion over $20$ pixels, and
no motion exceeds $25$ pixels (training set). Ground-truth optical flows for the test set are
not publicly available.

\paragraph{The Kitti dataset \cite{kitti} }\ 
contains real-world sequences taken from a driving platform.
The dataset includes non-Lambertian surfaces, different lighting conditions,
a large variety of materials and large displacements. More than 16\%
of the pixels have motion over 20 pixels. Again, ground-truth optical flows for the test set 
are not publicly available.

\paragraph{Performance metric for matching.}\ 
Choosing a performance measure for matching approaches
is delicate. Matching approaches typically do not return dense
correspondences, but output varying numbers of matches.
Furthermore, correspondences might be concentrated
in different areas of the image. 

Most matching approaches, including DeepMatching, 
are based on establishing correspondences between patches.
Given a pair of matching patches, it is possible to obtain a list 
of pixel correspondences for all pixels within the patches. 
We introduce a measure based on the number of 
correctly matched \emph{pixels} compared to the overall number of pixels. 
We  define ``accuracy@$T$'' as the proportion of
``correct'' pixels from the first image with respect to the total
number of pixels. A pixel is considered correct if its pixel match
in the second image is closer than $T$ pixels to ground-truth.
In practice, we use a threshold of $T=10$ pixels, as
this represents a sufficiently precise estimation (about 1\% of image diagonal for all datasets),
while allowing some tolerance in blurred areas that are difficult to match exactly.
If a pixel belongs to several matches, we choose the one with the
highest score to predict its correspondence.  
Pixels which do not belong to any patch have an infinite error.

\paragraph{Performance metric for optical flow.}\ 
To evaluate optical flow, we follow the standard protocol and 
measure the average endpoint error over all pixels, 
denoted as ``EPE''. The ``s10-40'' variant measures the EPE only for pixels
with a ground-truth displacement between 10 and 40 pixels, and
likewise for ``s0-10'' and ``s40+''. In all cases, scores are averaged
over all image pairs to yield the final result for a given dataset.

\subsection{Matching Experiments}
\label{sub:matchingxp}

In this section, we evaluate DeepMatching (DM). We present results for
all datasets presented above but Middlebury, which does not feature
long-range motions, the main difficulty in image matching. 
When evaluating on the Mikolajczyk dataset, we employ the scale and
rotation invariant version of DM presented in Section~\ref{sub:scalerot}. 
For all the matching experiments reported in this section, we use the Mikolajczyk dataset and the training sets of MPI-Sintel and Kitti. 

\subsubsection{Impact of the parameters}
\label{sub:paramstuning}

We optimize the different parameters of DM jointly on all datasets. 
To prevent overfitting, we use the same parameters across all datasets.

\begin{figure}
\includegraphics[width=1\linewidth]{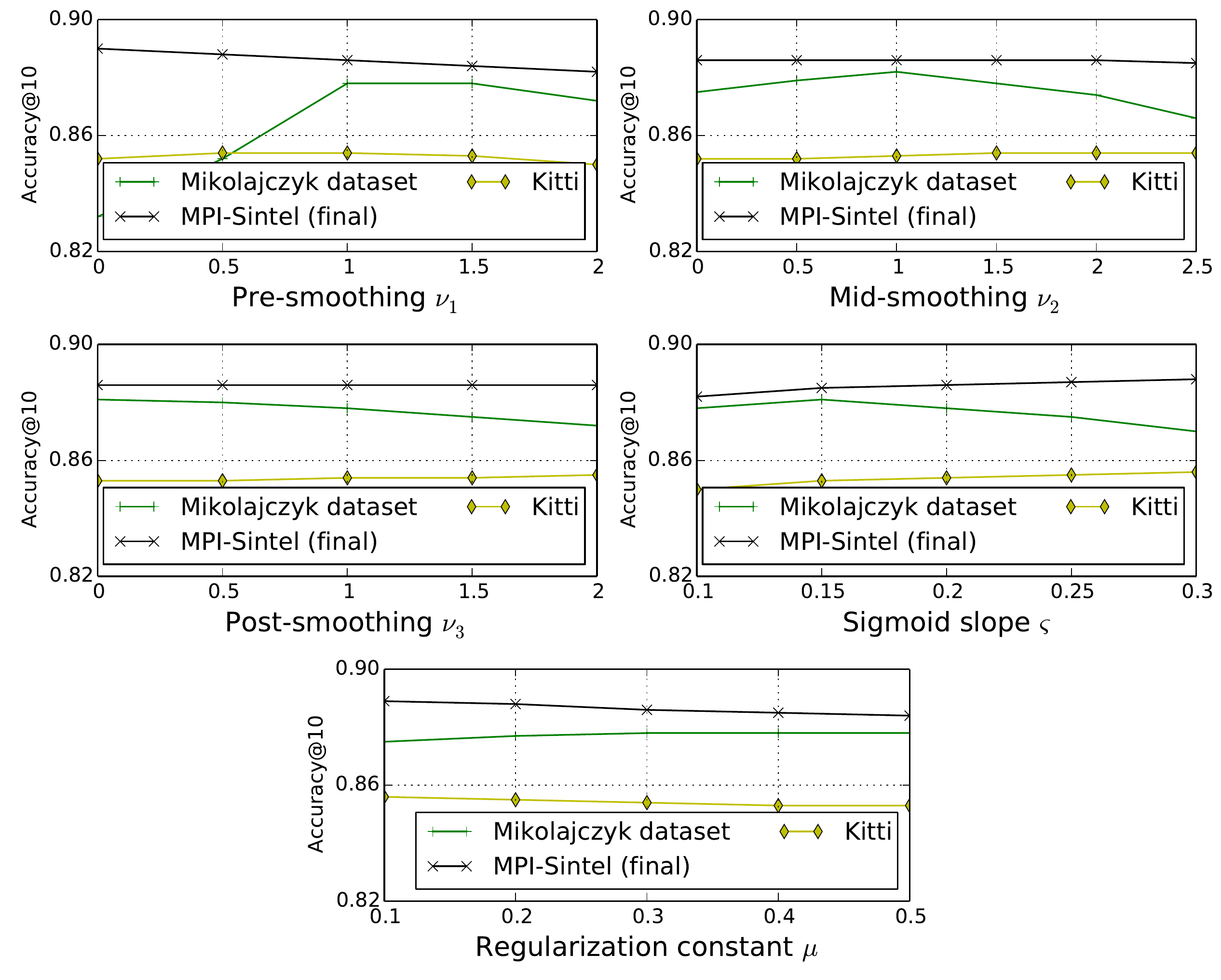}
\vspace{-5mm}
\caption{Impact of the parameters to compute pixel descriptors
on the different datasets.}
\label{fig:pixeldesc}
\end{figure}

\paragraph{Pixel descriptor parameters:\:}
\red{We first optimize the parameters of the pixel representation (Section~\ref{sub:responsepyramid}):
$\nu_1$, $\nu_2$, $\nu_3$ (different smoothing stages), $\varsigma$ (sigmoid slope)
and $\mu$ (regularization constant). After performing a grid search, we find
that good results are obtained at $\nu_1=\nu_2=\nu_3=1$, $\varsigma=0.2$ 
and $\mu=0.3$ across all datasets. Figure~\ref{fig:pixeldesc} shows the accuracy@10 
in the neighborhood of these values for all parameters. Image pre-smoothing seems to be crucial for JPEG images (Mikolajczyk dataset),
as it smooths out compression artifacts, whereas it slightly 
degrades performance for uncompressed PNG images (MPI-Sintel and Kitti). 
As expected, similar findings are observed for the regularization constant $\mu$ 
since it acts as a regularizer that reduces the impact of small gradients (\ie noise). 
In the following, we thus use low values of $\nu_1$ and $\mu$ when dealing with PNG images 
(we set $\nu_1=0$ and $\mu=0.1$, other parameters are unchanged).}

\begin{figure}
\includegraphics[width=1\linewidth]{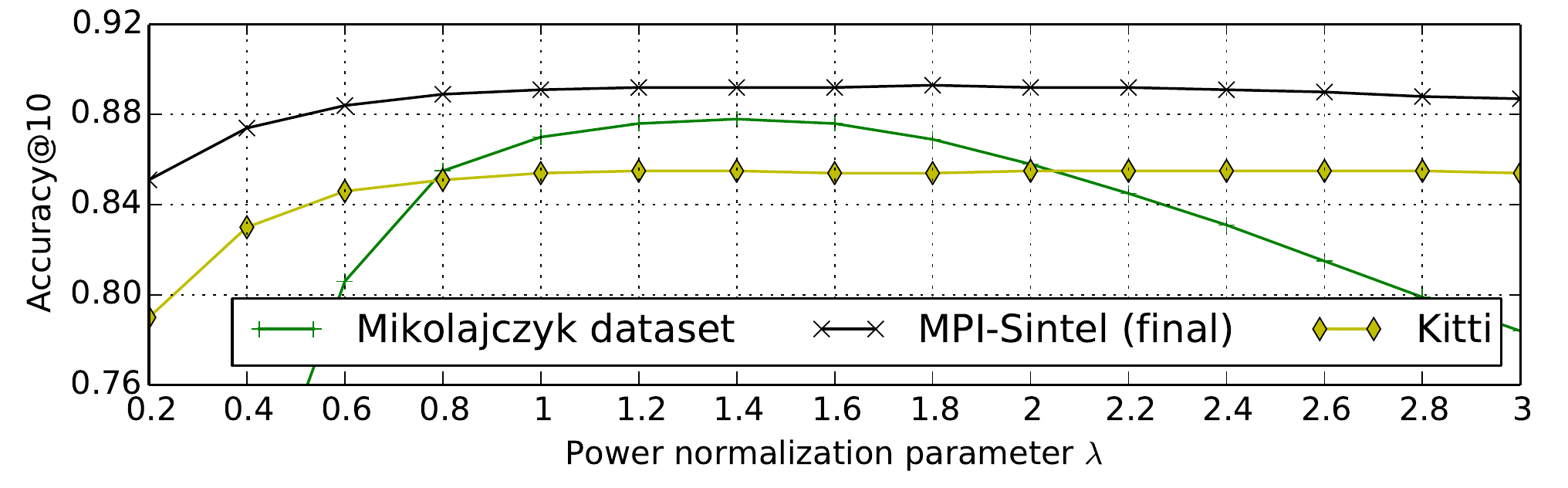}
\vspace{-3mm}
\caption{Impact of the non-linear response rectification (eq. \eqref{eqn:nlpow}).}
\label{fig:rectif}
\end{figure}

\paragraph{Non-linear rectification:\:}
We also evaluate the impact of the parameter $\lambda$ of the non-linear
rectification obtained by applying power normalization, see \eq\refp{eqn:nlpow}.
Figure \ref{fig:rectif} displays the accuracy@10 for various values of $\lambda$.
We can observe that the optimal performance is achieved at $\lambda=1.4$ for all datasets. 
We use this value in the remainder of our experiments.

\subsubsection{\red{Evaluation of the backtracking and scoring schemes}}
\label{sub:oldvsnew}

\newcommand{\DMold}{DM*\xspace}

\red{We now evaluate two improvements of DM with respect
to the previous version published in~\cite{DeepFlow}, referred to as \DMold:
\begin{itemize}
\item Backtracking (BT) entry points: in \DMold we 
select as entry points local maxima in the correlation maps from
all pyramid levels. The new alternative is to
start from all possible points in the top pyramid level.
\item Scoring scheme: In \DMold we scored atomic 
correspondences based on the correlation values of start and end point
of the backtracking path.
The new scoring scheme is the sum of correlation values along the 
full backtracking path. 
\end{itemize}
}

\newcommand{\cm}{\checkmark}

\begin{table}
\centering
\resizebox{\linewidth}{!}{\red{
\begin{tabular}{|ccc|crr|}
\hline 
     & New BT       & New scoring & accuracy@10 & memory & matching\tabularnewline
$R$  & entry points &     scheme  &             & usage  &  time   \tabularnewline
\hline 
\hline 
\multicolumn{6}{|c|}{\textbf{Mikolajczyk dataset}}\tabularnewline
\hline 
1/4 &     &     & 0.620 & 0.9 GB & 1.0 min\tabularnewline
1/2 &     &     & 0.848 & 5.5 GB & 20 min\tabularnewline
1/2 &     & \cm & 0.864 & 5.5 GB & 7.3 min\tabularnewline
1/2 & \cm & \cm & \textbf{0.878} & 4.4 GB & 6.3 min\tabularnewline
\hline 
\hline 
\multicolumn{6}{|c|}{\textbf{MPI-Sintel dataset (final)}}\tabularnewline
\hline 
1/4 &     &     & 0.822 & 0.4 GB & 2.4 sec\tabularnewline
1/2 &     &     & 0.880 & 6.3 GB & 55 sec\tabularnewline
1/2 &     & \cm & 0.890 & 6.3 GB & 16 sec\tabularnewline
1/2 & \cm & \cm & \textbf{0.892} & 4.6 GB & 16 sec\tabularnewline
\hline 
\hline 
\multicolumn{6}{|c|}{\textbf{Kitti dataset}}\tabularnewline
\hline 
1/4 &     &     & 0.772 & 0.4 GB & 2.0 sec\tabularnewline
1/2 &     &     & 0.841 & 6.3 GB & 39 sec\tabularnewline
1/2 &     & \cm & 0.855 & 6.3 GB & 14 sec\tabularnewline
1/2 & \cm & \cm & \textbf{0.856} & 4.7 GB & 14 sec\tabularnewline
\hline 
\end{tabular}
}}
\caption{\label{tab:oldvsnew}\red{
Detailed comparison between the preliminary and current versions of DeepMatching in terms of
performance, run-time and memory usage. 
$R$ denotes the input image resolution and BT backtracking. 
Run-times are computed on 1 core @ 3.6 GHz.}}
\end{table}

\red{
We report results for the different variants in Table~\ref{tab:oldvsnew} on each dataset.
The first two rows for each dataset correspond to the exact settings used for \DMold 
(\ie with an image resolution of 1/4 and 1/2).
We observe a steady increase in performance on all datasets when we
add the new scoring and backtracking approach. 
We can observe that starting from all possible
entry points in the top pyramid level (\ie considering all possible translations)
yields slightly better results than starting from local maxima. This 
demonstrates that some ground-truth matches are not covered by any local maximum. 
By enumerating all possible patch translations from the top-level, 
we instead ensure to fully explore the space of all possible matches. 

Furthermore, it is interesting to note
that memory usage and run-time significantly decreases when using the new options.
This is because (1) searching and storing local maxima (which are exponentially more
numerous in lower pyramid levels) is not necessary anymore,
and (2) the new scoring scheme allows for further optimization, \ie
early pruning of backtracking paths (Section~\ref{sub:backtrackcorrespondences}).
}

\subsubsection{Approximate DeepMatching}
\label{sub:approxxps}

We now evaluate the performance of approximate DeepMatching (Section~\ref{sub:compr})
and report its run-time and memory usage. We evaluate and compare
two different ways of reducing the computational load. The
first one simply consists in downsizing the input images, and upscaling 
the resulting matches accordingly. The second option is the
compression scheme proposed in Section~\ref{sub:compr}.

\begin{figure}
\includegraphics[width=1\linewidth]{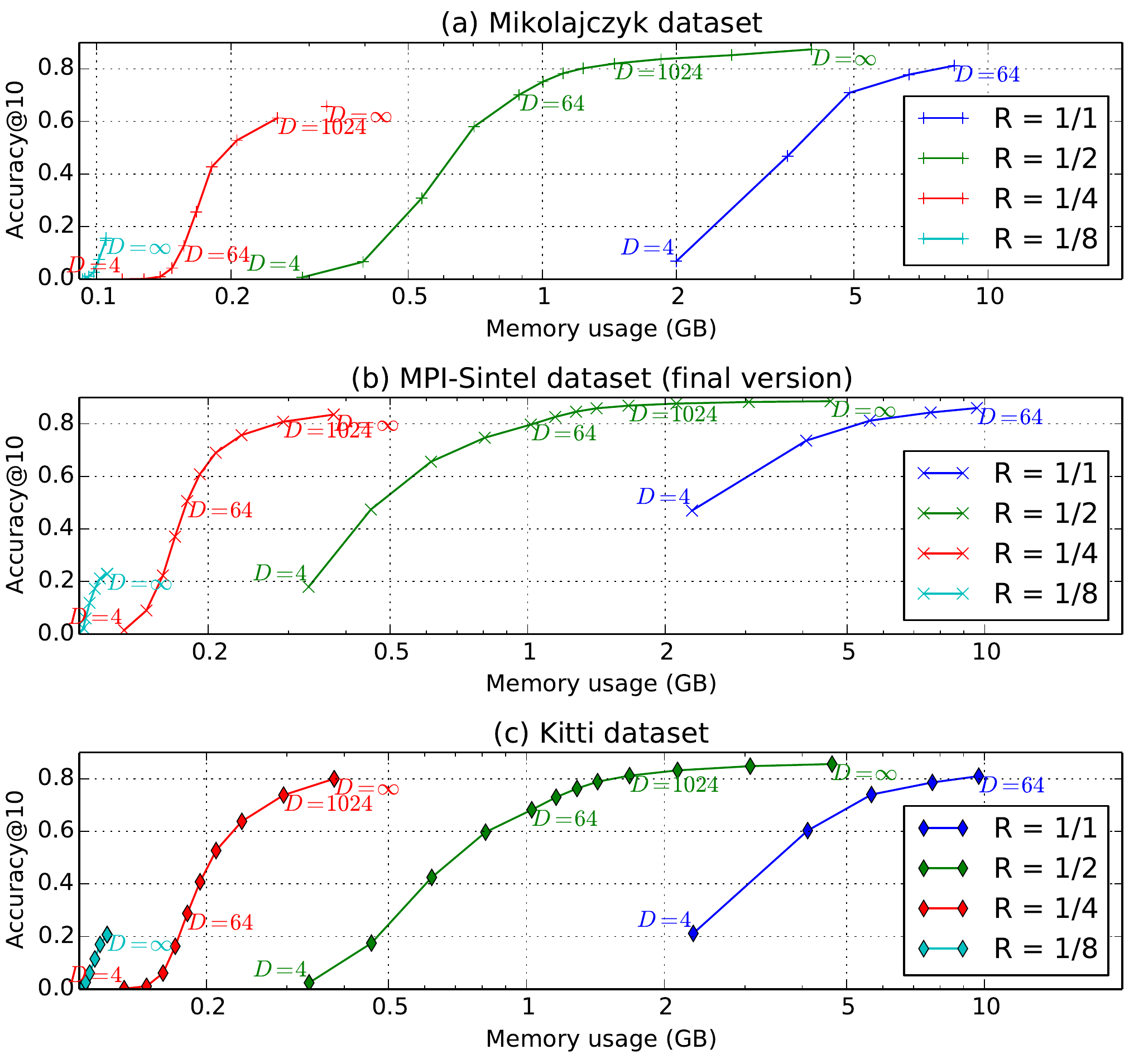}

\caption{\label{fig:mem-perf}Trade-off between memory consumption and matching
performance for the different datasets. Memory usage is controlled
by changing image resolution $R$ (different curves) and dictionary
size $D$ (curve points). 
}
\end{figure}

\begin{figure}
\center
\includegraphics[width=0.8\linewidth]{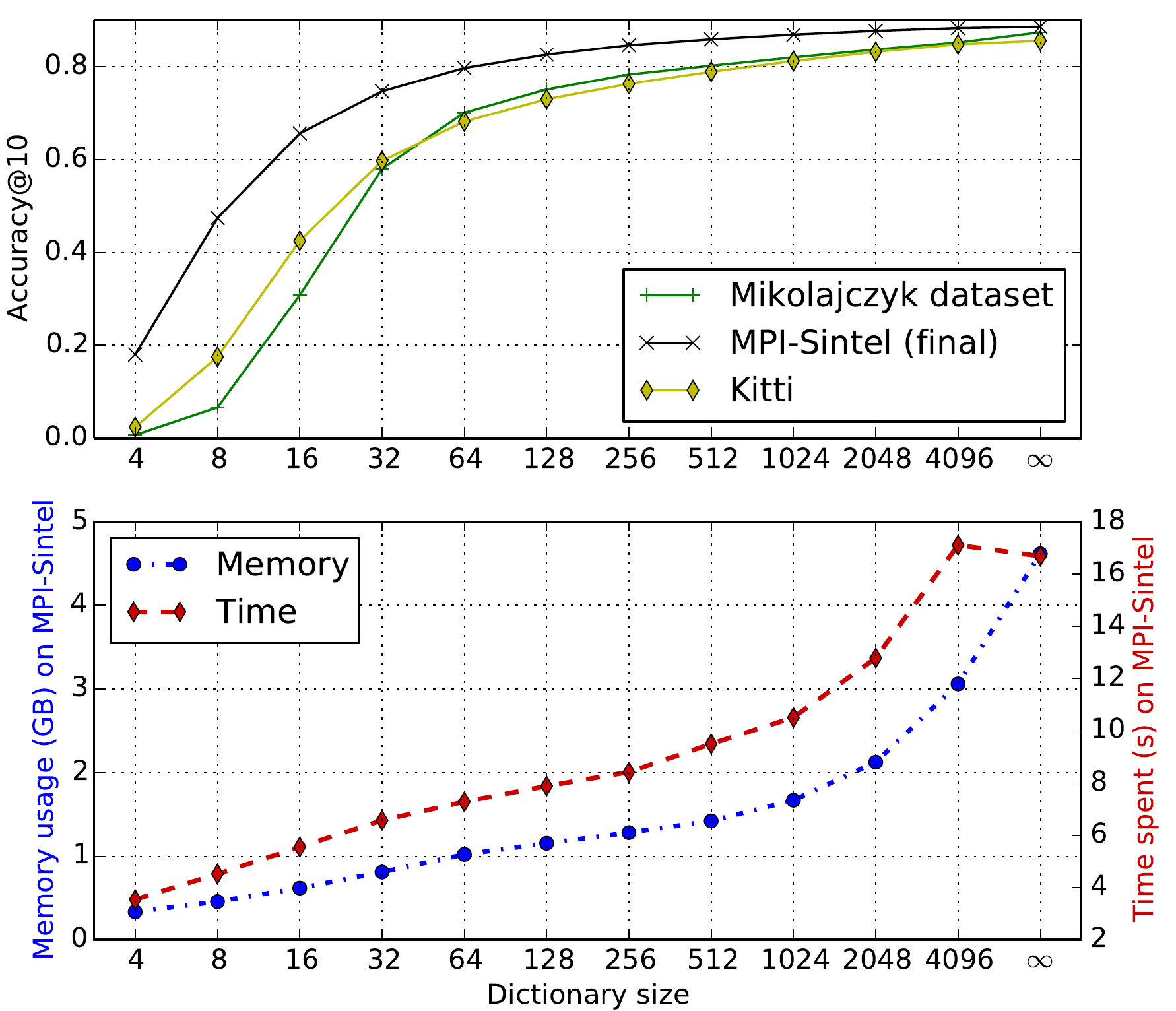}

\caption{\label{fig:tradeoff}Performance, memory usage and run-time for different
levels of compression corresponding to the size $D$ of the prototype
dictionary (we set  the image resolution to $R=1/2$). A dictionary size $D=\infty$ stands for no compression.
\red{Run-times are for a single image pair on 1 core @ 3.6 GHz.}
}
\end{figure}

We evaluate both schemes jointly by varying the input image size (expressed
as a fraction $R$ of the original resolution) and the size
$D$ of the prototype dictionary (\ie parameter of k-means in
Section \ref{sub:compr}). \red{$R=1$ corresponds to the original dataset 
image size (no downsizing). }
We display the results in terms of matching
accuracy (accuracy@10) against memory consumption in Figure \ref{fig:mem-perf}
and as a function of $D$ in Figure \ref{fig:tradeoff}. Figure
\ref{fig:mem-perf} shows that DeepMatching can be computed
in an approximate manner for any given memory budget. Unsurprisingly,
too low settings (\eg $R\leq1/8$, $D\leq64$) result in a strong
loss of performance. It should be noted that that we were unable to compute
DeepMatching at full resolution ($R=1)$ for $D>64$, as the memory
consumption explodes. As a consequence, all subsequent experiments
in the paper are done at $R=1/2$. In Figure \ref{fig:tradeoff},
we observe that 
good trades-off are achieved for dictionary sizes
comprised in $D\in[64,1024]$. For instance, on MPI-Sintel, 
at $D=1024$, 94\%
of the performance of the uncompressed case ($D=\infty$) is reached for half the
computation time and one third of the memory usage. 
\red{Detailed timings of the different stages of DeepMatching are
given in Table~\ref{tab:timings}. As expected, only the bottom-up pass
is affected by the approximation, with a run-time of the different operations involved 
(patch correlations, max-pooling, subsampling, aggregation and non-linear rectification)
roughly proportional to $D$ (or to $\left|\mathcal{G}_4\right|$, the actual number of atomic patches, if $D=\infty$).
}
The overhead of clustering the dictionary prototypes with k-means appears 
negligible, with the exception of the largest dictionary size
($D=4096$) for which it induces a slightly longer run-time than in
the uncompressed case. Overall, the proposed method for approximating
DeepMatching is highly effective.

\newcommand{\caffe}[0]{\texttt{Caffe}\xspace}

\red{
\paragraph{GPU Implementation.}
We have implemented DM on GPU in the \caffe framework~\citep{caffe}. 
Using existing \caffe layers like ConvolutionLayer and PoolingLayer, 
the implementation is straightforward for most layers. 
We had to specifically code a few layers which are not available in \caffe
(\eg the backtracking pass\footnote{Although the 
backtracking is conceptually close to the back-propagation training algorithm, it 
differs in term of how the scores are accumulated for each path.}). For the aggregation layer
which consists in selecting and averaging 4 children channels out of many channels, 
we relied on the sparse matrix multiplication in the cuSPARSE toolbox. 
Detailed timings are given in Table~\ref{tab:timings} on a GeForce Titan X.
Our code runs in about 0.2s for a pair of MPI-Sintel image. As expected, the computation 
bottleneck essentially lies in the computation of bottom-level patch correlations
and the backtracking pass. Note that computing patch descriptors 
takes significantly more time, in proportion, than on CPU: 
it takes about 0.024s = 11\% of total time (not shown in table). 
This is because it involves a succession of many small layers (image smoothing, 
gradient extraction and projection, \etc), which causes overhead and is rather inefficient.
}

\begin{table}
\resizebox{\linewidth}{!}{\red{
\begin{tabular}{|ccc||cccccc|c|}
\hline 
Proc. &   &   &    Patch    & Patch       & Max-pooling  & Aggre- & Non-linear    & Back-    & Total\tabularnewline
Unit       & R & D & clustering  & Correlations& +subsampling & gation & rectification & tracking & time\tabularnewline
\hline 
\bf{CPU} & 1/2 & 64       & 0.3 & 0.2 & 0.4 & 0.9 & 0.8 & 5.1 & 7.7 \tabularnewline
\bf{CPU} & 1/2 & 1024     & 1.3 & 0.7 & 0.6 & 1.0 & 1.3 & 5.8 & 10.7 \tabularnewline
\bf{CPU} & 1/2 & $\infty$ &  -  & 4.3 & 1.6 & 1.0 & 3.2 & 6.2 & 16.4 \tabularnewline
\hline
\bf{GPU} & 1/2 & $\infty$ & - & 0.084 & 0.012 & 0.017 & 0.013 & 0.053 & 0.213 \tabularnewline
\hline 
\end{tabular}
}}
\caption{\label{tab:timings}\red{
Detailed timings of the different stages of DeepMatching, measured for a single
image pair from MPI-Sintel on CPU (1 core @ 3.6GHz) and GPU (GeForce Titan X) in seconds. Stages are: 
patch clustering (only for approximate DM, see Section~\ref{sub:compr}), 
patch correlations (\eq\refp{eqn:init}),
joint max-pooling and subsampling, correlation map aggregation,
non linear rectification (resp. $\mathcal{S} \circ \mathcal{P}$, $\sum \mathcal{T}_{\bm{o}_i}$,
and $\mathcal{R}_\lambda$ in \eq\refp{eqn:recurr}), and correspondence backtracking (Section~\ref{sub:backtrackcorrespondences}).
Other operations (\eg reciprocal verification of \eq\refp{eqn:reciprocal}) have negligible run-time.
For operations applied at several levels like the non-linear rectification, a cumulative timing is given. }}
\end{table}

\subsubsection{Comparison to the state of the art}
\label{sub:dmsota}

We compare DM with several baselines and state-of-the-art matching algorithms,
namely:
\begin{itemize}
\item SIFT keypoints extracted with DoG detector~\citep{Lowe2004}
and matched with FLANN \citep{flann}, referred to as SIFT-NN,\footnote{\label{ourimpl}We implemented this method ourselves.}
\item dense HOG matching, followed by nearest-neighbor matching with reciprocal verification 
as done in LDOF \citep{Bro11a}, referred to as HOG-NN\footnoteref{ourimpl},
\item Generalized PatchMatch (GPM)~\citep{Barnes2010}, with default
parameters, 32x32 patches and 20 iterations (best settings in our experiments)\footnote{\label{onlinecode}We used the online code.},
\item Kd-tree PatchMatch (KPM)~\citep{kpm}, an improved version of
  PatchMatch based on better patch descriptors and
  kd-trees optimized for correspondence propagation\footnoteref{ourimpl},
\item Non-Rigid Dense Correspondences (NRDC)~\citep{HaCohen2011}, an 
improved version of GPM based on a multiscale iterative expansion/contraction
strategy\footnote{We report results from the original paper.},
\item \red{SIFT-flow~\citep{siftflow}, a dense matching algorithm based
on an energy minimization where pixels are represented as SIFT features 
and a smoothness term is incorporated to explicitly preserve spatial discontinuities\footnoteref{onlinecode}, }
\item \red{Scale-less SIFT (SLS)~\citep{siftscales}, an improvement of SIFT-flow to handle 
scale changes (multiple sized SIFTs are extracted and 
combined to form a scale-invariant pixel representation)\footnoteref{onlinecode},}
\item \red{DaisyFilterFlow (DaisyFF)~\citep{daisyff}, a dense matching
approach that combines filter-based efficient flow inference and 
the Patch-Match fast search algorithm to match pixels described using the 
DAISY representation~\citep{daisy}\footnoteref{onlinecode},}
\item \red{Deformable Pyramid Matching (DSP) \citep{Kim2013}, a dense matching
approach based on a coarse-to-fine (top-down) strategy where inference 
is performed with (inexact) loopy belief propagation\footnoteref{onlinecode}. }
\end{itemize}
SIFT-NN, HOG-NN and DM output sparse matches, 
whereas the other methods output fully dense correspondence fields.
SIFT keypoints, GPM, NRDC and DaisyFF are scale and
rotation invariant, whereas HOG-NN, \linebreak KPM, SIFT-flow, SLS and DSP are not. 
We, therefore, do not report results for these latter methods on the
Mikolajczyk dataset which includes image rotations and scale changes.

Statistics about each method  (average number of matches per image and
their coverage) are reported in Table~\ref{tab:mstats}. \red{Coverage
is computed as the proportion of points on a regular grid with 10
pixel spacing for which there exists a correspondence
(in the raw output of the considered method) within a 10 pixel neighborhood.
}
Thus, it measures how well matches ``cover'' the image.
Table~\ref{tab:mstats} shows that DeepMatching outputs 2 to 7 times more matches than SIFT-NN
and a comparable number to HOG-NN. 
Yet, the coverage for DM matches is much higher than for HOG-NN and
SIFT-NN. This shows that DM matches are well distributed 
over the entire image, which is not the case for HOG-NN and SIFT-NN,
as they have difficulties estimating matches in regions with weak or
repetitive textures.

\begin{table}
  \resizebox{\linewidth}{!}{
  \begin{tabular}{|c||c|c||c|c||c|c|}
    \hline 
    \multirow{2}{*}{Method} & \multicolumn{2}{c||}{Mikolajczyk} & \multicolumn{2}{c||}{MPI-Sintel (final)} & \multicolumn{2}{c|}{Kitti}\tabularnewline
    \cline{2-7} 
     & \# & coverage & \# & coverage & \# & coverage\tabularnewline
    \hline 
    \hline 
    SIFT-NN & 2084 & 0.59 & 836 & 0.25 & 1299 & 0.38\tabularnewline
    HOG-NN & - & - & 4576 & 0.39 & 4293 & 0.34\tabularnewline
    \hline 
    KPM & - & - & 446K & 1 & 462K & 1\tabularnewline
    GPM & 545K & 1 & 446K & 1 & 462K & 1\tabularnewline
    NRDC & 545K & 1 & 446K & 1 & 462K & 1\tabularnewline
    \red{SIFT-flow} & - & - & 446K & 1 & 462K & 1\tabularnewline
    \red{SLS} & - & - & 446K & 1 & 462K & 1\tabularnewline
    \red{DaisyFF} & 545K & 1 & 446K & 1 & 462K & 1\tabularnewline
    \red{DSP} & - & - & 446K & 1 & 462K & 1\tabularnewline
    \hline 
    DM (ours) & 3120 & 0.81 & 5920 & 0.96 & 5357 & 0.88\tabularnewline
    \hline 
  \end{tabular}}
  \caption{\label{tab:mstats}Statistics of the different matching methods. The
``\#'' column refers to the average number of matches per image, and
the coverage to the proportion of points on a regular grid with 10 pixel spacing
that have a match within a 10px neighborhood.}
\end{table}

\begin{table}
\centering
\resizebox{\linewidth}{!}{
\begin{tabular}{|l|c|c|crr|}
\hline 
method & $R$ & $D$ & accuracy@10 & memory & matching\tabularnewline
 &  &  &  & usage &  time\tabularnewline
\hline 
\hline 
\multicolumn{6}{|c|}{\textbf{Mikolajczyk dataset}}\tabularnewline
\hline 
\multicolumn{3}{|l|}{SIFT-NN} & 0.674 & 0.2 GB & 1.4 sec\tabularnewline
\multicolumn{3}{|l|}{GPM} & 0.303 & 0.1 GB & 2.4 min\tabularnewline
\multicolumn{3}{|l|}{NRDC} & 0.692 & 0.1 GB & 2.5 min\tabularnewline
\multicolumn{3}{|l|}{\red{DaisyFF}} & 0.410 & 6.1 GB & 16 min\tabularnewline
DM & 1/4 & $\infty$ & 0.657 & 0.9 GB & 38 sec\tabularnewline
DM & 1/2 & 1024 & 0.820 & 1.5 GB & 4.5 min\tabularnewline
DM & 1/2 & $\infty$ & \textbf{0.878} & 4.4 GB & 6.3 min\tabularnewline
\hline 
\hline 
\multicolumn{6}{|c|}{\textbf{MPI-Sintel dataset (final)}}\tabularnewline
\hline 
\multicolumn{3}{|l|}{SIFT-NN} & 0.684 & 0.2 GB & 2.7 sec\tabularnewline
\multicolumn{3}{|l|}{HOG-NN} & 0.712 & 3.4 GB & 32 sec\tabularnewline
\multicolumn{3}{|l|}{KPM} & 0.738 & 0.3 GB & 7.3 sec\tabularnewline
\multicolumn{3}{|l|}{GPM} & 0.812 & 0.1 GB & 1.1 min\tabularnewline
\multicolumn{3}{|l|}{\red{SIFT-flow}} & 0.890 & 1.0 GB & 29 sec\tabularnewline
\multicolumn{3}{|l|}{\red{SLS}} & 0.824 & 4.3 GB & 16 min\tabularnewline
\multicolumn{3}{|l|}{\red{DaisyFF}} & 0.873 & 6.8 GB & 12 min\tabularnewline
\multicolumn{3}{|l|}{\red{DSP}} & 0.853 & 0.8 GB & 39 sec\tabularnewline
DM & 1/4 & $\infty$ & 0.835 & 0.3 GB & 1.6 sec\tabularnewline
DM & 1/2 & 1024 & 0.869 & 1.8 GB & 10 sec\tabularnewline
DM & 1/2 & $\infty$ & \textbf{0.892} & 4.6 GB & 16 sec\tabularnewline
\hline 
\hline 
\multicolumn{6}{|c|}{\textbf{Kitti dataset}}\tabularnewline
\hline 
\multicolumn{3}{|l|}{SIFT-NN} & 0.489 & 0.2 GB & 1.7 sec\tabularnewline
\multicolumn{3}{|l|}{HOG-NN} & 0.537 & 2.9 GB & 24 sec\tabularnewline
\multicolumn{3}{|l|}{KPM} & 0.536 & 0.3 GB & 17 sec\tabularnewline
\multicolumn{3}{|l|}{GPM} & 0.661 & 0.1 GB & 2.7 min\tabularnewline
\multicolumn{3}{|l|}{\red{SIFT-flow}} & 0.673 & 1.0 GB & 25 sec\tabularnewline
\multicolumn{3}{|l|}{\red{SLS}} & 0.748 & 4.4 GB & 17 min\tabularnewline
\multicolumn{3}{|l|}{\red{DaisyFF}} & 0.796 & 7.0 GB & 11 min\tabularnewline
\multicolumn{3}{|l|}{\red{DSP}} & 0.580 & 0.8 GB & 2.9 min\tabularnewline
DM & 1/4 & $\infty$ & 0.800 & 0.3 GB & 1.6 sec\tabularnewline
DM & 1/2 & 1024 & 0.812 & 1.7 GB & 10 sec\tabularnewline
DM & 1/2 & $\infty$ & \textbf{0.856} & 4.7 GB & 14 sec\tabularnewline
\hline 
\end{tabular}
}
\caption{\label{tab:matching}
Matching performance, run-time and memory usage 
for  state-of-the-art methods and DeepMatching (DM). 
For the proposed method, $R$ and $D$ denote the input
image resolution and the dictionary size ($\infty$ stands for no compression).
Run-times are computed on 1 core @ 3.6 GHz.}
\end{table}

\begin{figure*}
\resizebox{\linewidth}{!}{
\begin{tabular}{cc|c|c|c}
 & {\huge bark} & {\huge boat} & {\huge graf} & {\huge wall}\tabularnewline
\begin{sideways}
{\LARGE Ground-truth} 
\end{sideways} & \includegraphics[height=0.2\linewidth]{{fig2_miko_gt_gt_miko_bark_1_6_A}.jpg}
\includegraphics[height=0.2\linewidth]{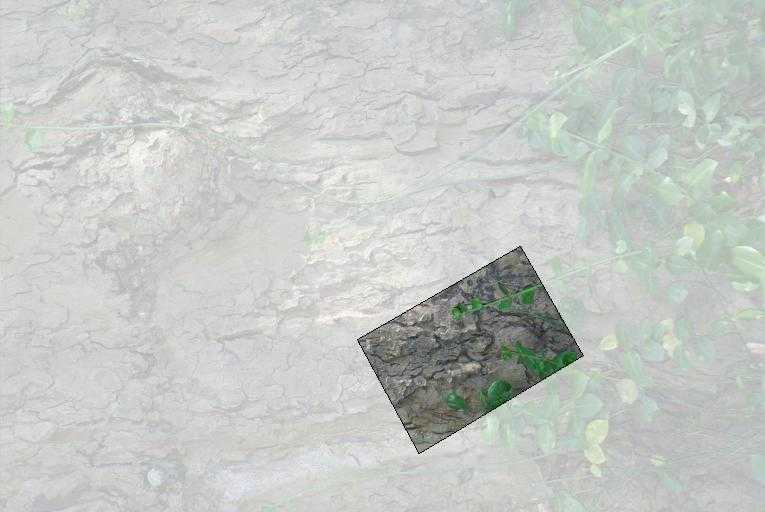} & \includegraphics[height=0.2\linewidth]{{fig2_miko_gt_gt_miko_boat_1_6_A}.jpg}
\includegraphics[height=0.2\linewidth]{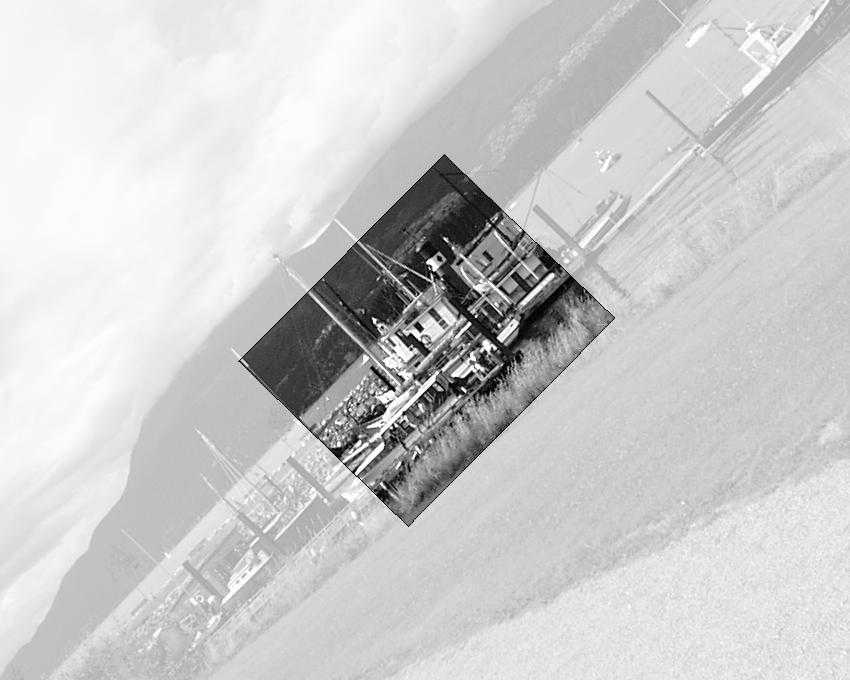} & \includegraphics[height=0.2\linewidth]{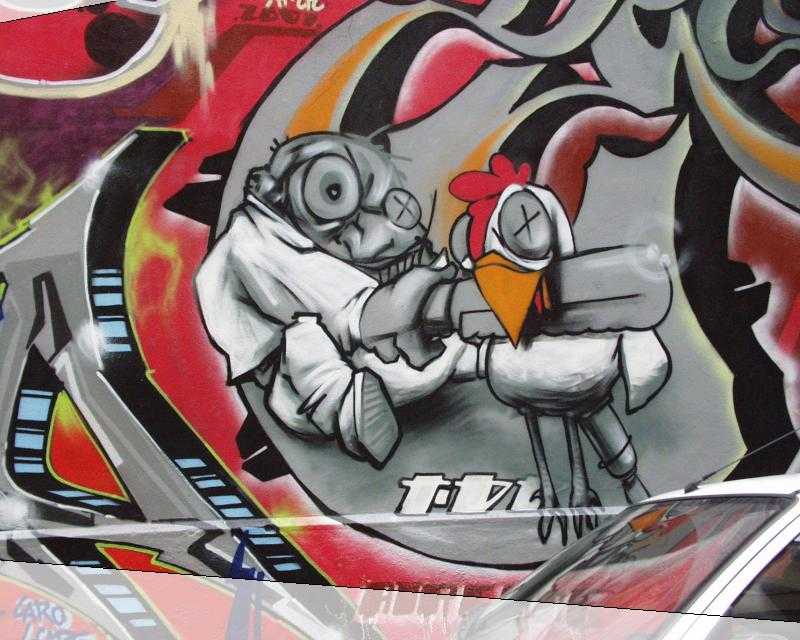}
\includegraphics[height=0.2\linewidth]{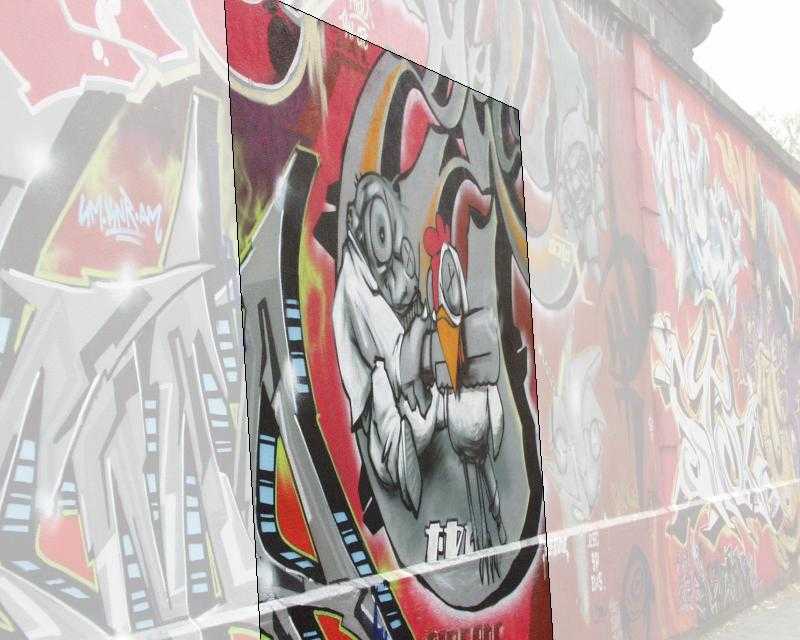} & \includegraphics[height=0.2\linewidth]{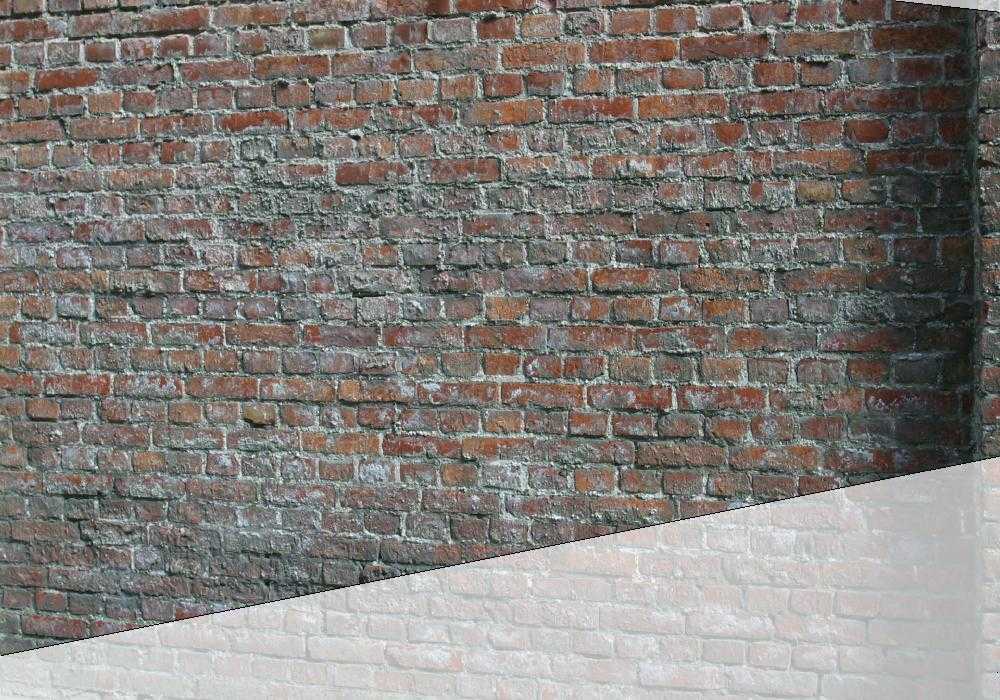}
\includegraphics[height=0.2\linewidth]{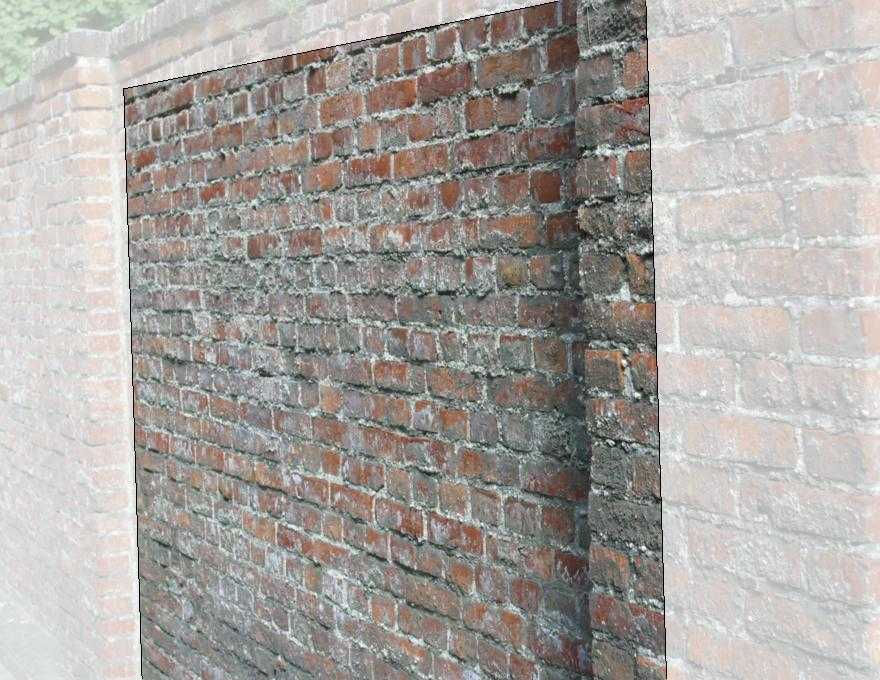}\tabularnewline
\begin{sideways}
~~~~~~{\LARGE SIFT-NN} 
\end{sideways} & \includegraphics[height=0.2\linewidth]{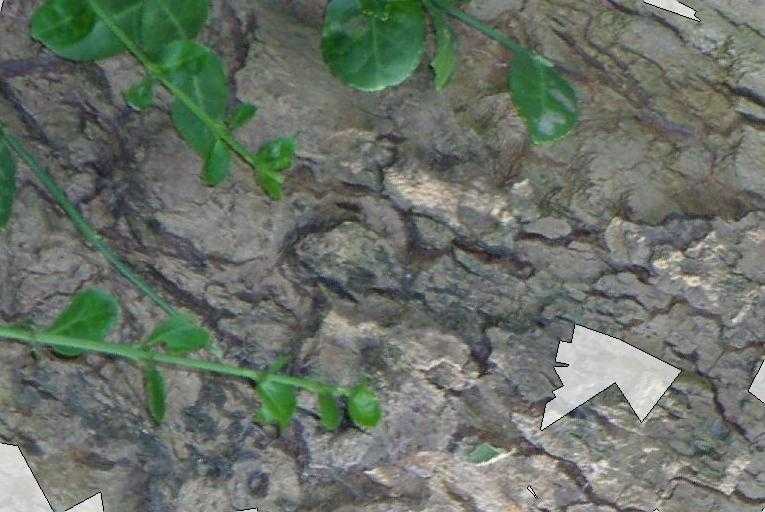}
\includegraphics[height=0.2\linewidth]{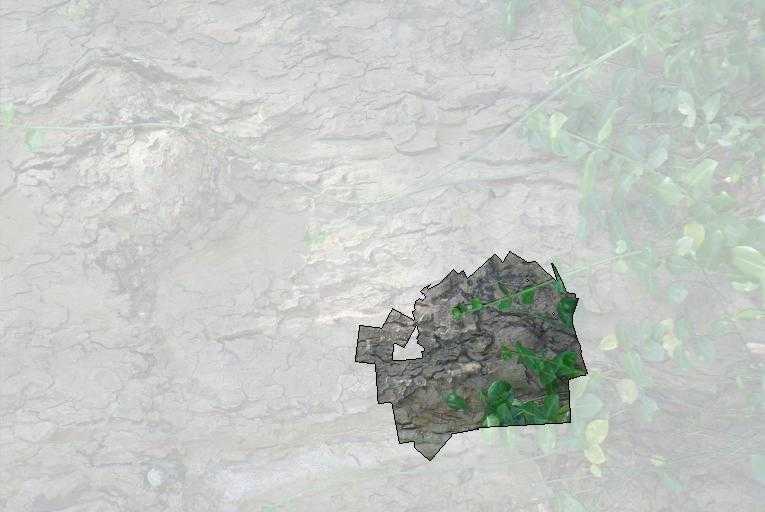} & \includegraphics[height=0.2\linewidth]{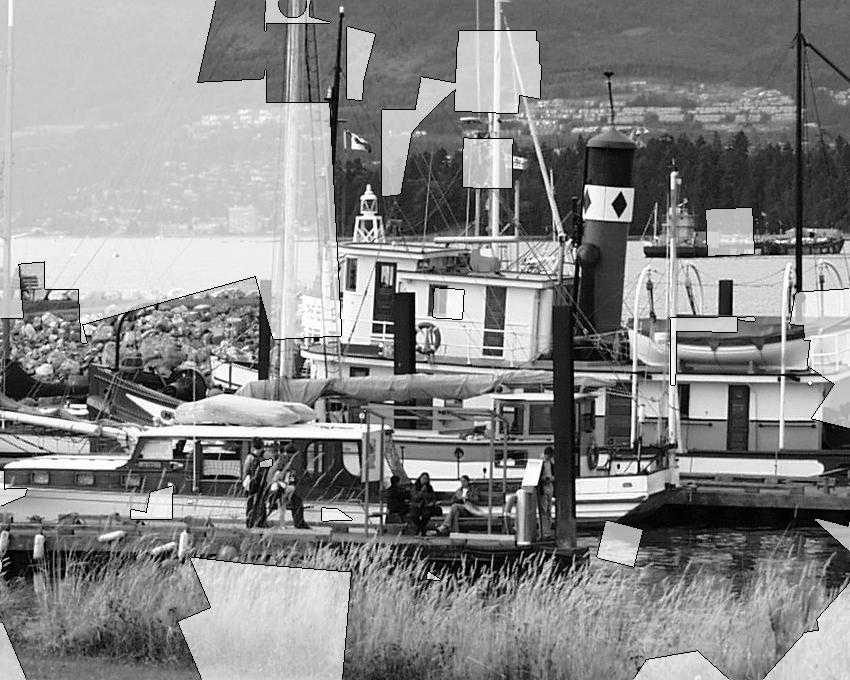}
\includegraphics[height=0.2\linewidth]{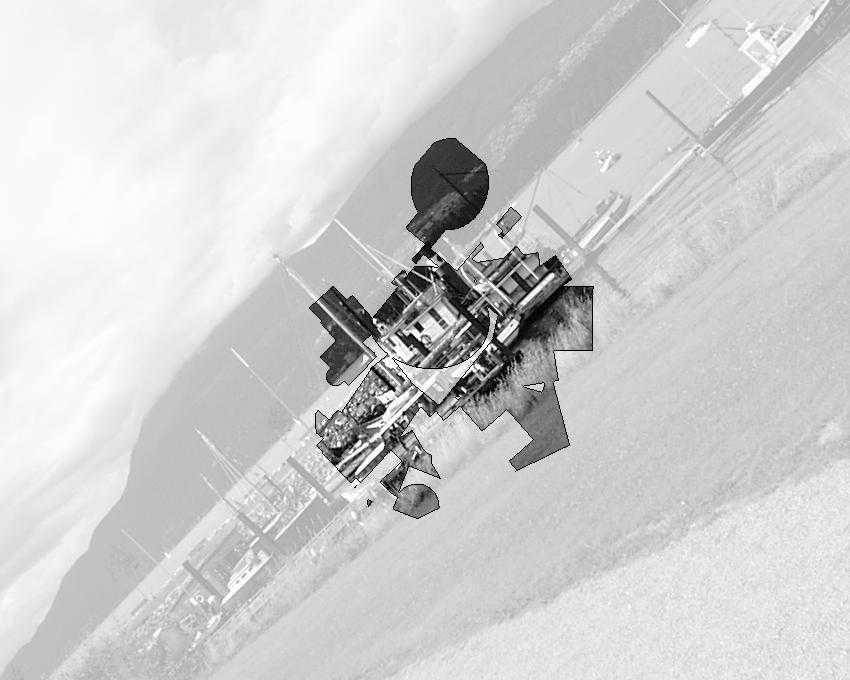} & \includegraphics[height=0.2\linewidth]{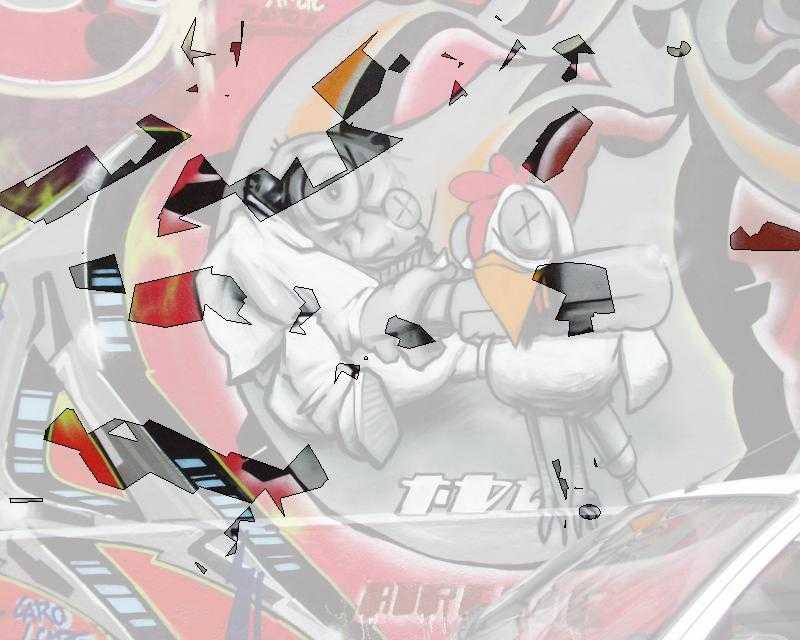}
\includegraphics[height=0.2\linewidth]{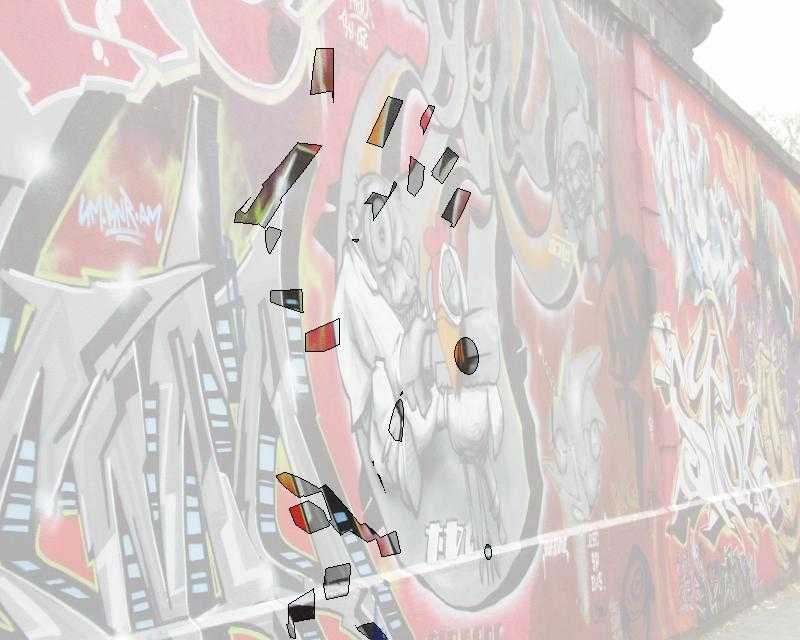} & \includegraphics[height=0.2\linewidth]{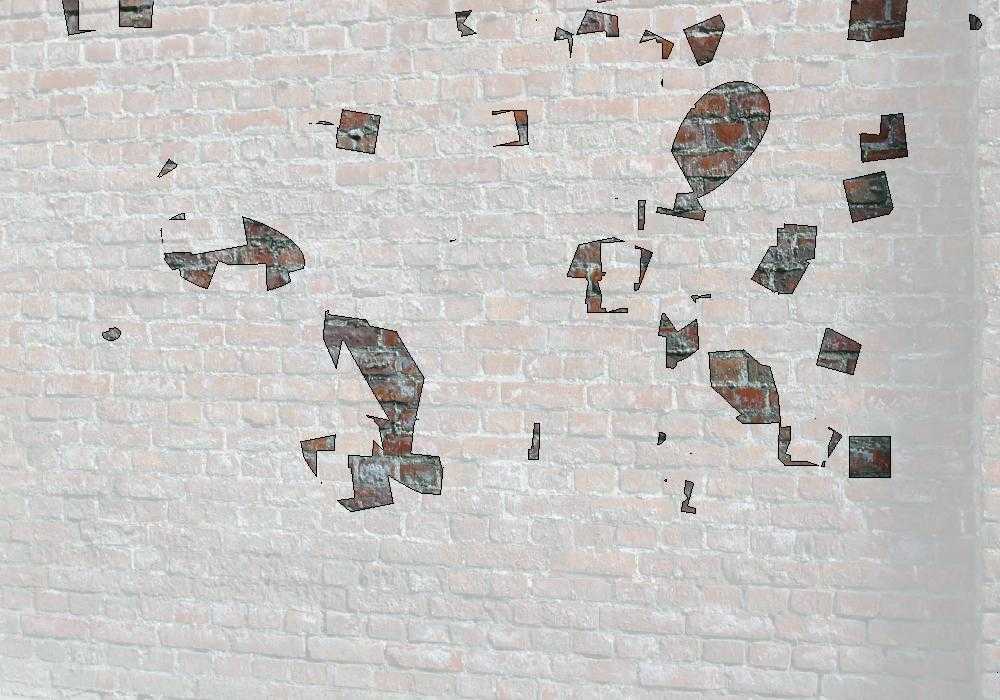}
\includegraphics[height=0.2\linewidth]{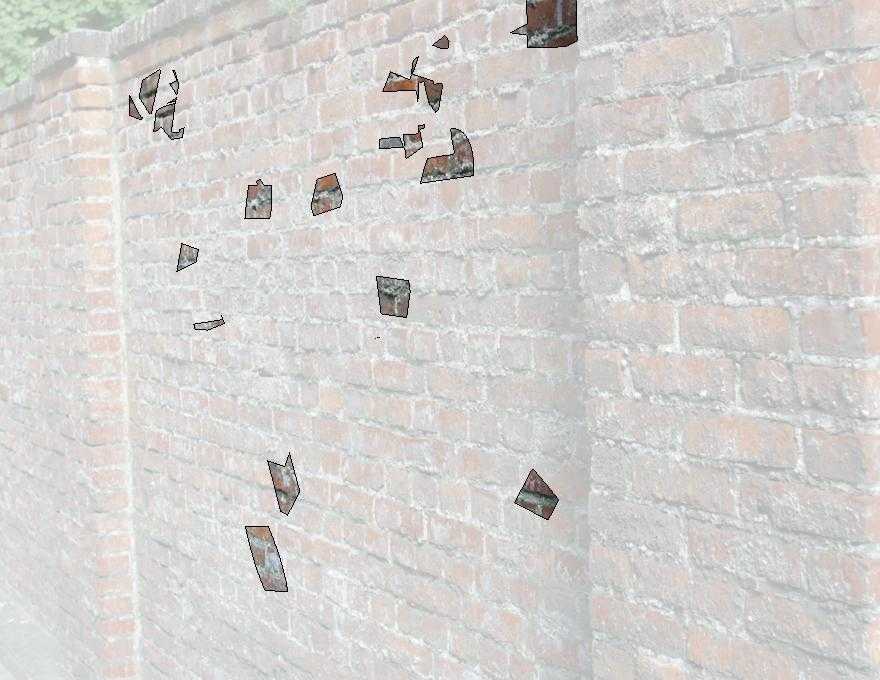}\tabularnewline
\begin{sideways}
~~~~~~~~~{\LARGE GPM} 
\end{sideways} & \includegraphics[height=0.2\linewidth]{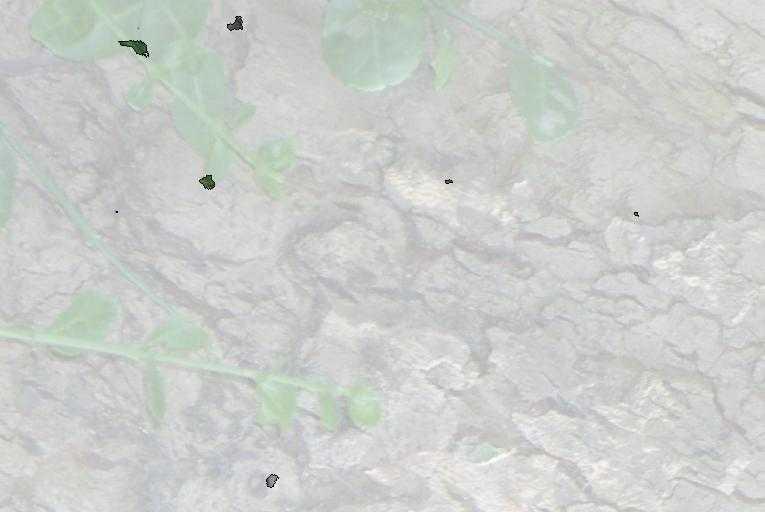}
\includegraphics[height=0.2\linewidth]{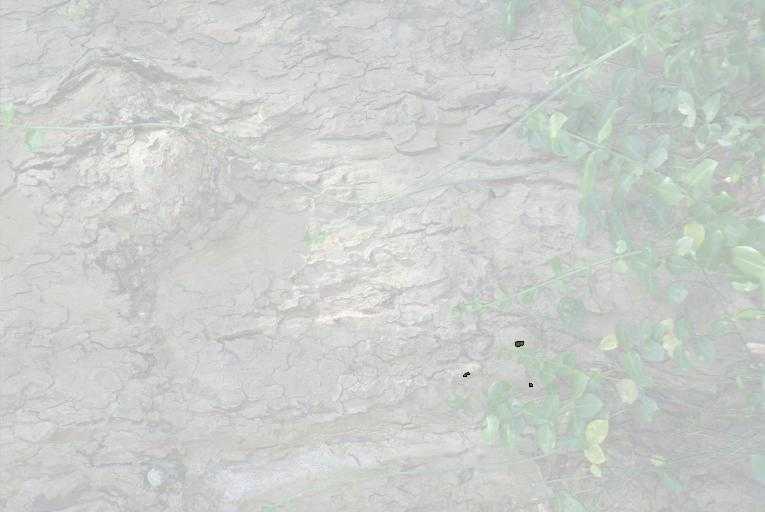} & \includegraphics[height=0.2\linewidth]{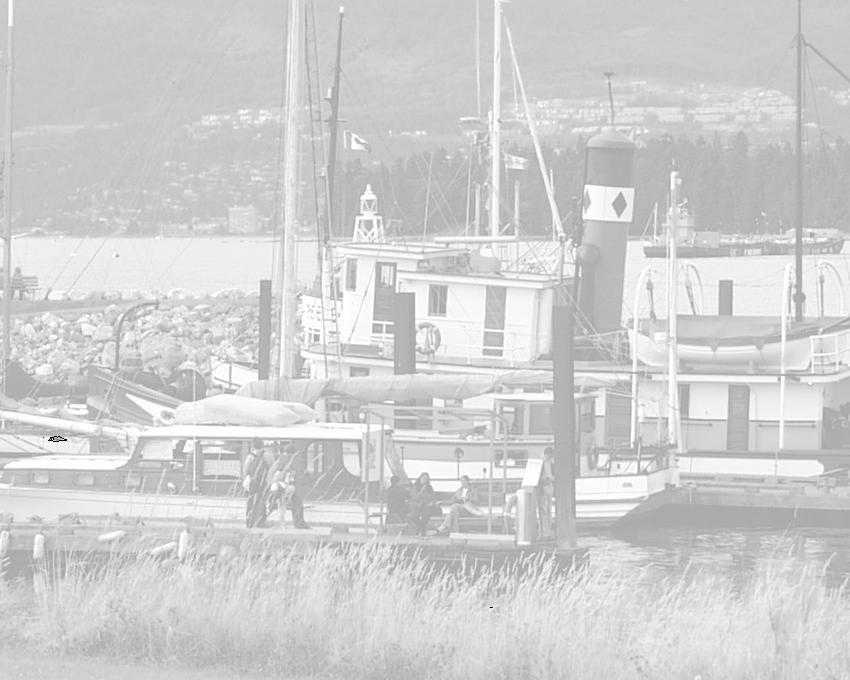}
\includegraphics[height=0.2\linewidth]{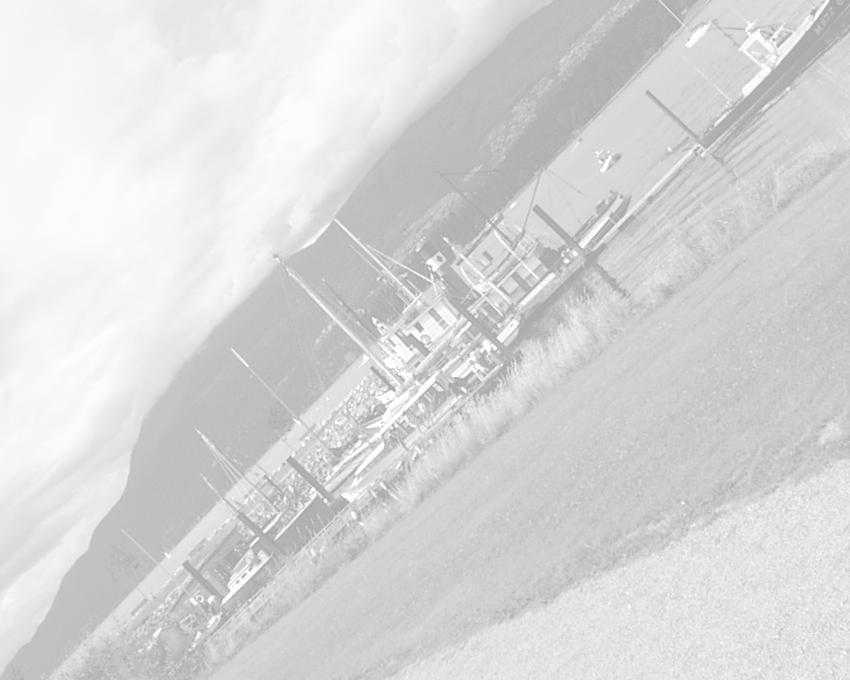} & \includegraphics[height=0.2\linewidth]{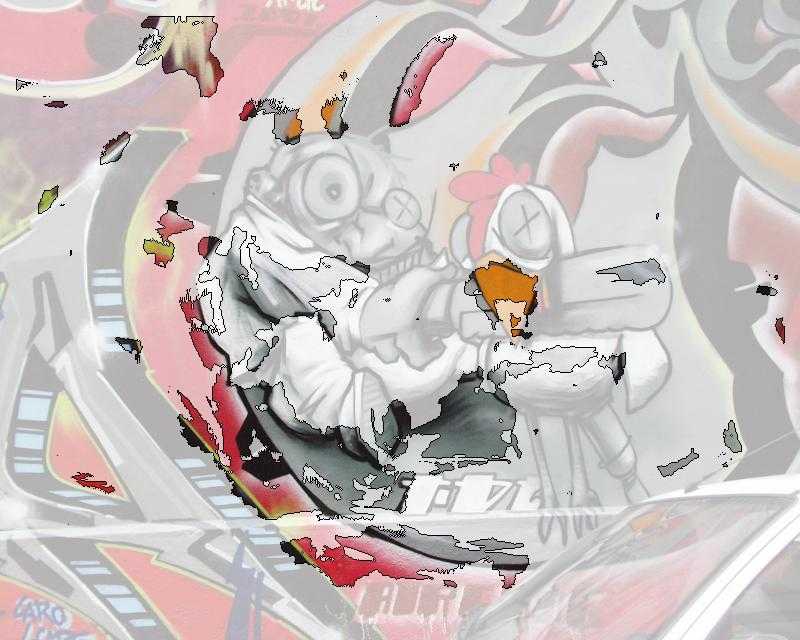}
\includegraphics[height=0.2\linewidth]{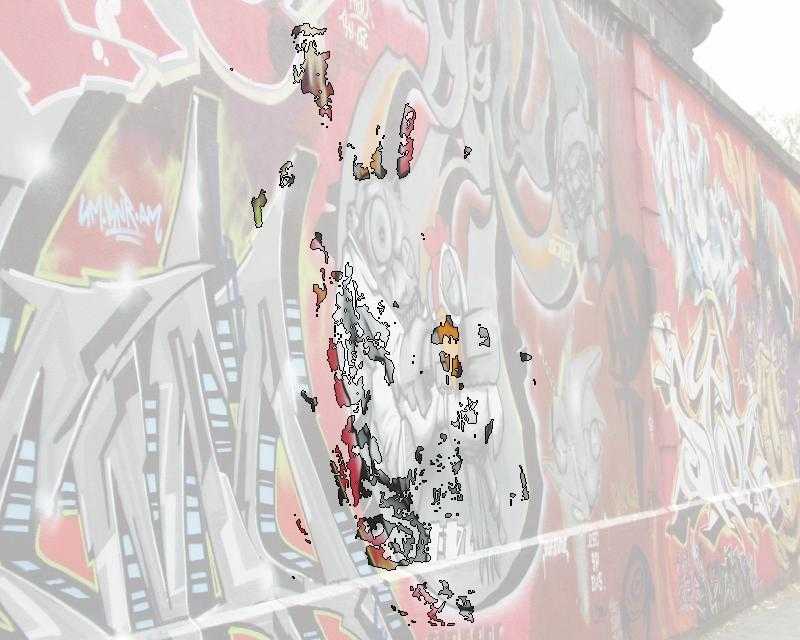} & \includegraphics[height=0.2\linewidth]{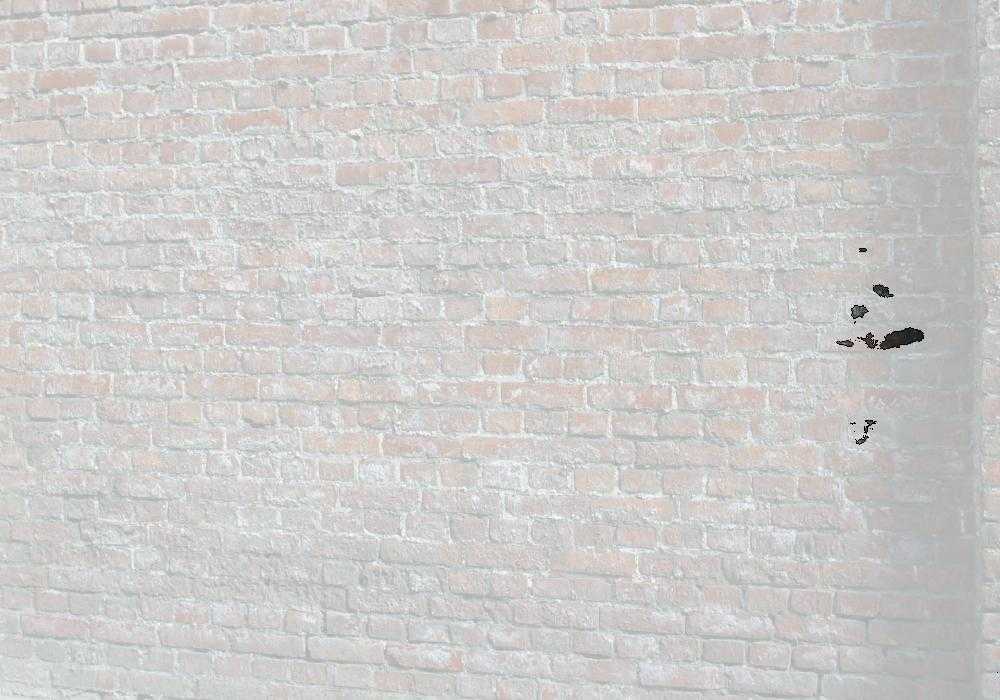}
\includegraphics[height=0.2\linewidth]{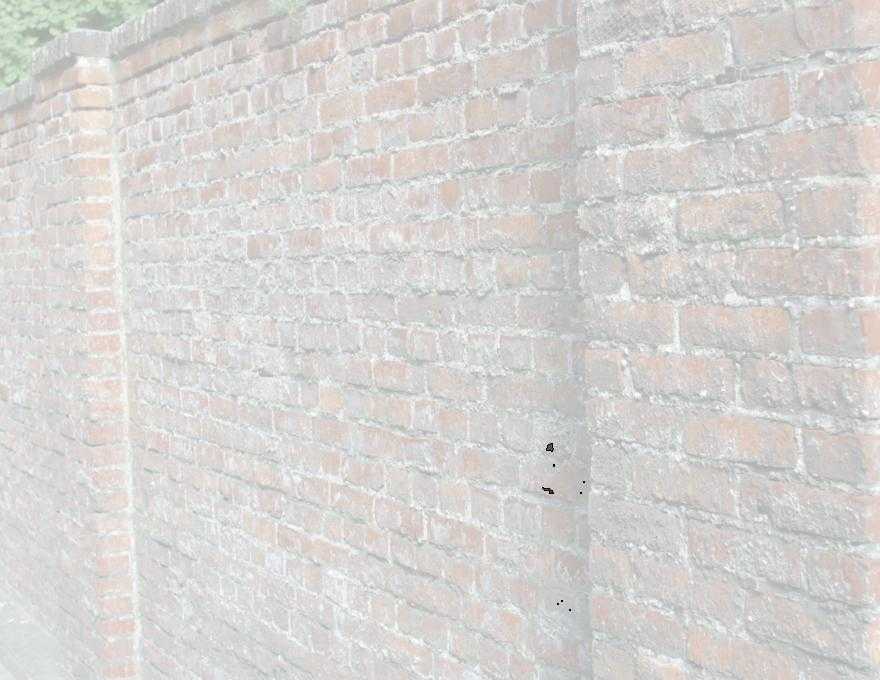}\tabularnewline
\begin{sideways}
~~~~~~~~~{\LARGE NRDC} 
\end{sideways} & \includegraphics[height=0.2\linewidth]{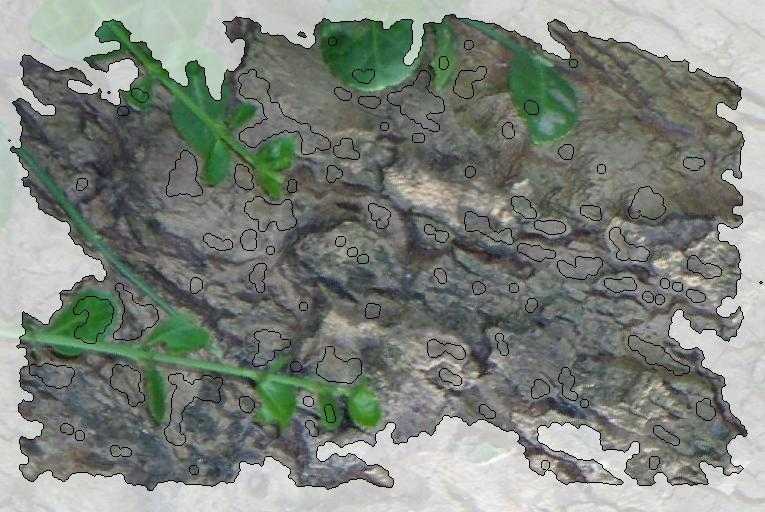}
\includegraphics[height=0.2\linewidth]{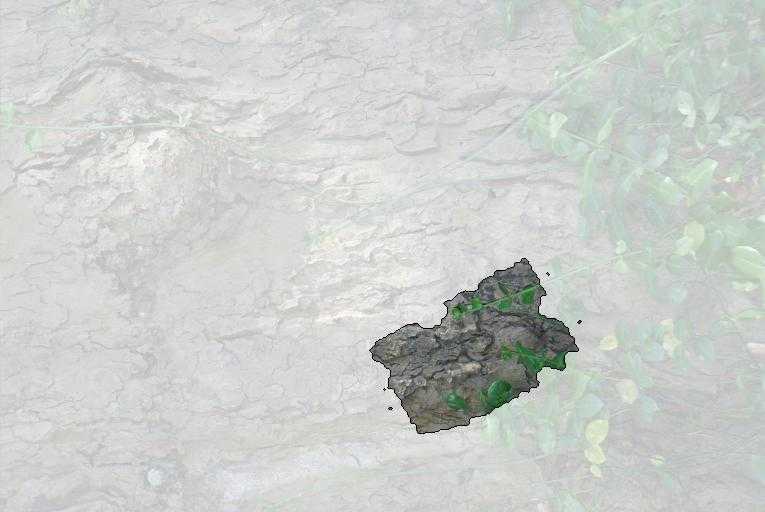} & \includegraphics[height=0.2\linewidth]{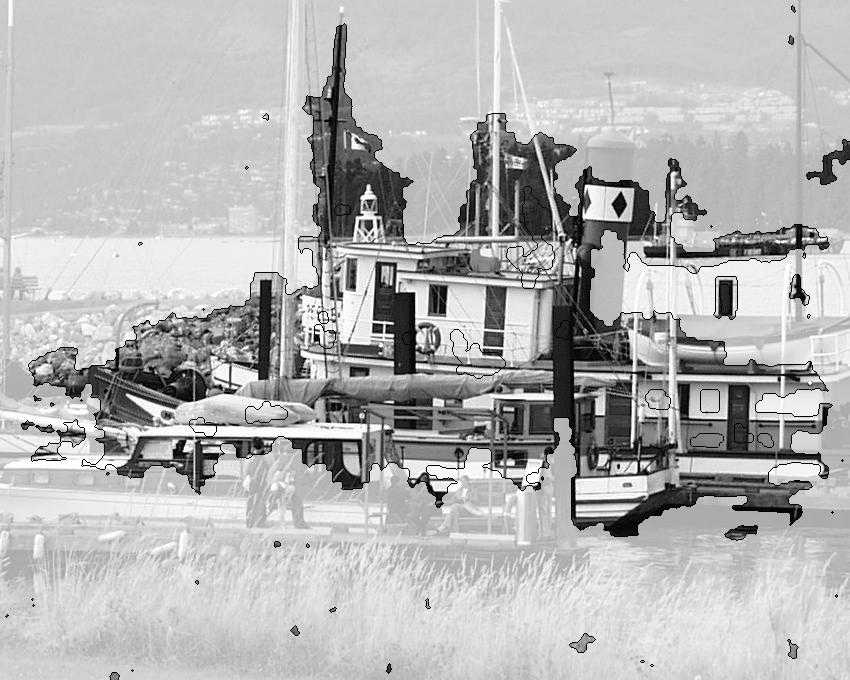}
\includegraphics[height=0.2\linewidth]{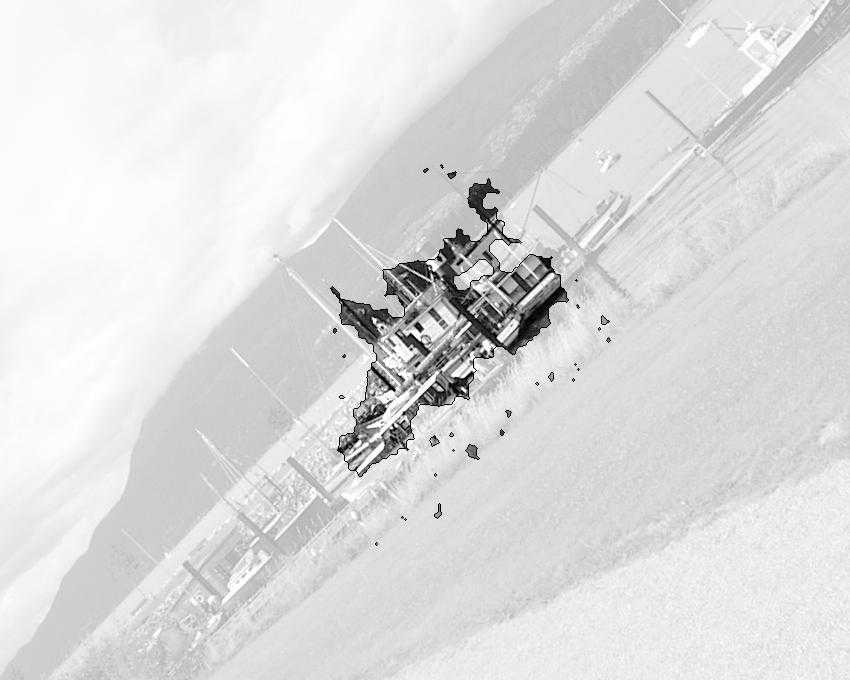} & \includegraphics[height=0.2\linewidth]{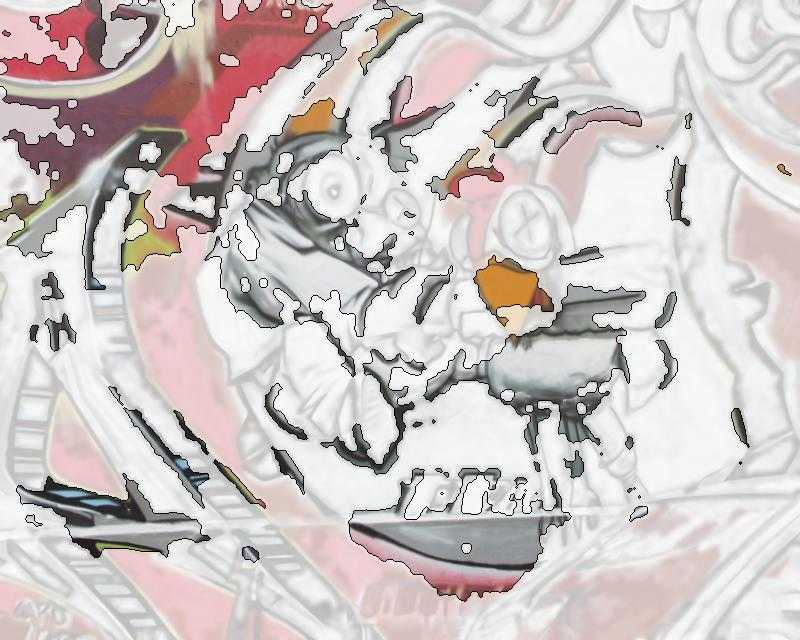}
\includegraphics[height=0.2\linewidth]{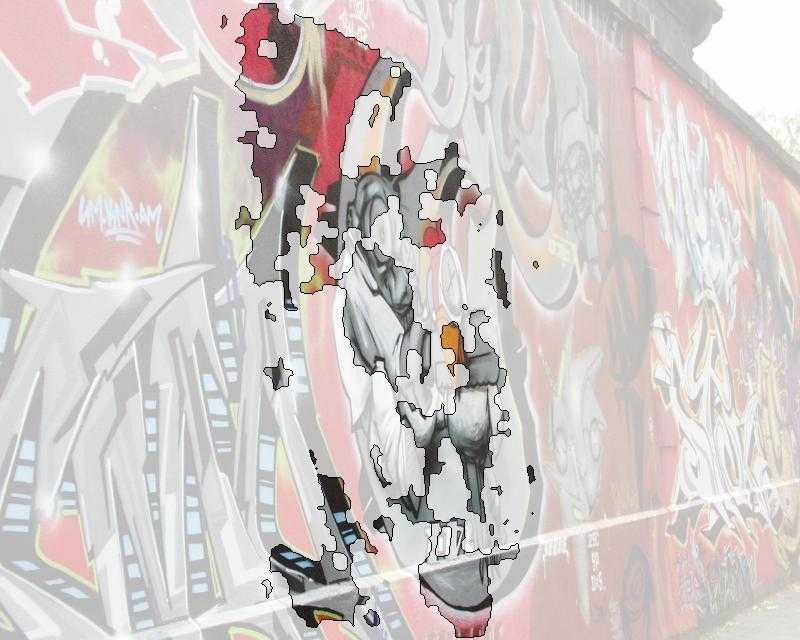} & \includegraphics[height=0.2\linewidth]{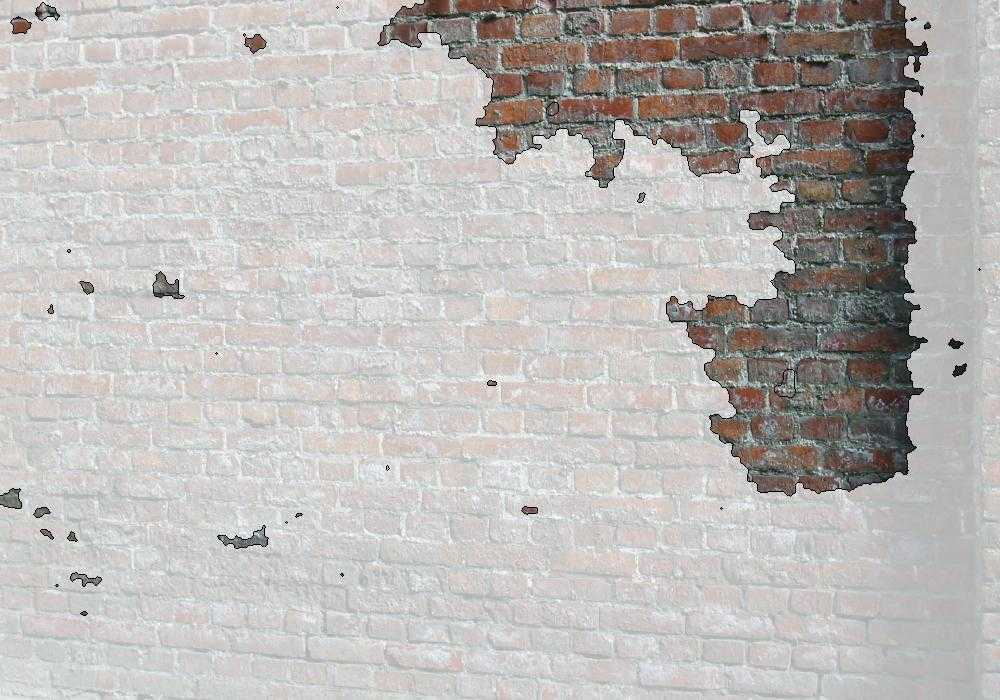}
\includegraphics[height=0.2\linewidth]{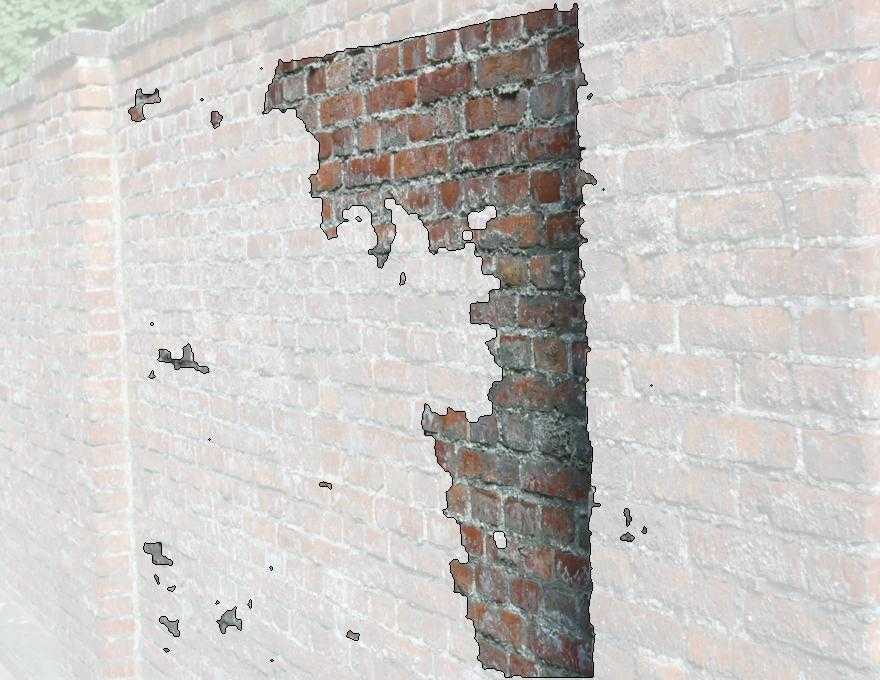}\tabularnewline
\begin{sideways}
~~~~~~~~~{\LARGE DaisyFF} 
\end{sideways} & \includegraphics[height=0.2\linewidth]{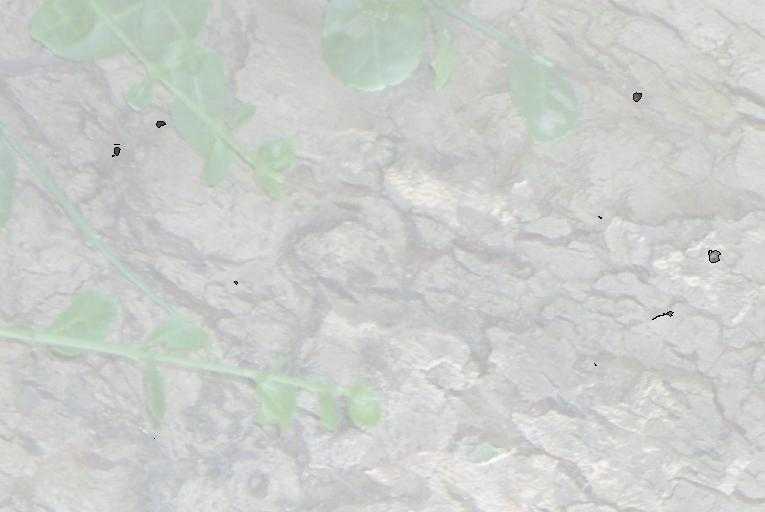}
\includegraphics[height=0.2\linewidth]{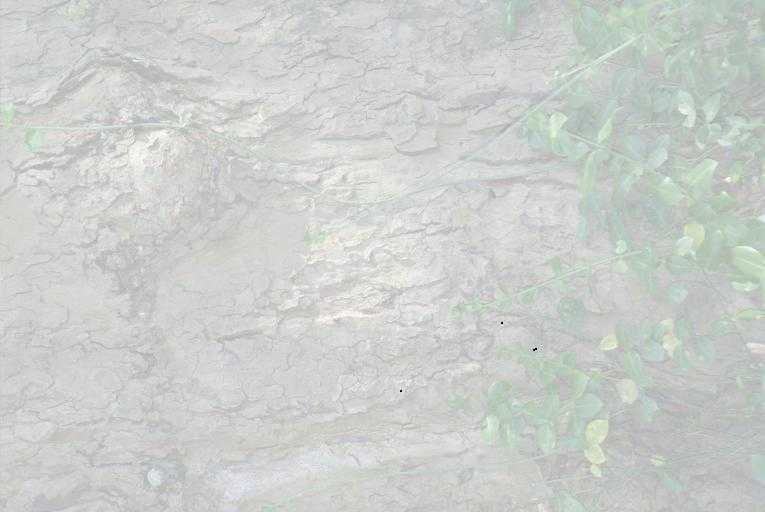} & \includegraphics[height=0.2\linewidth]{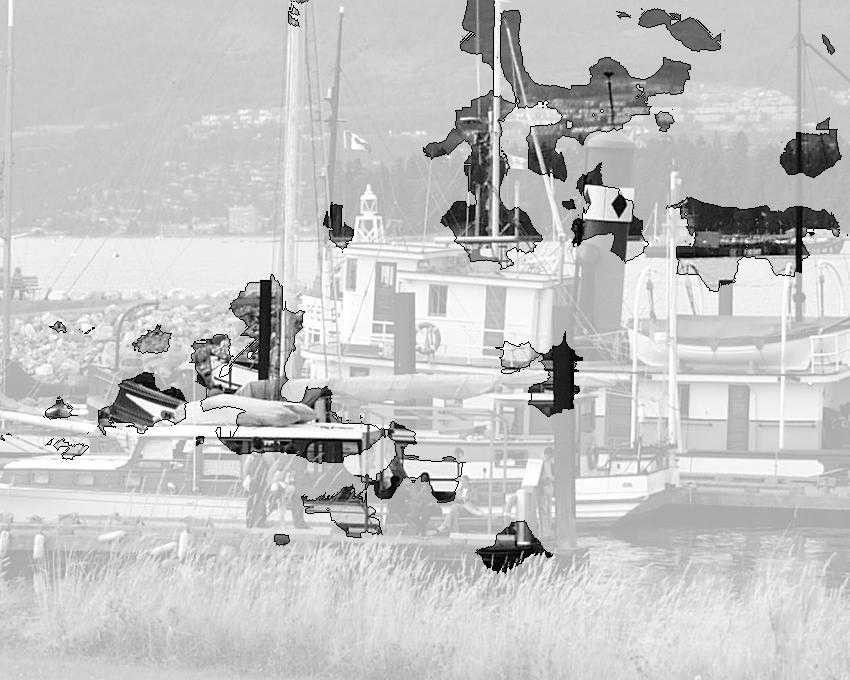}
\includegraphics[height=0.2\linewidth]{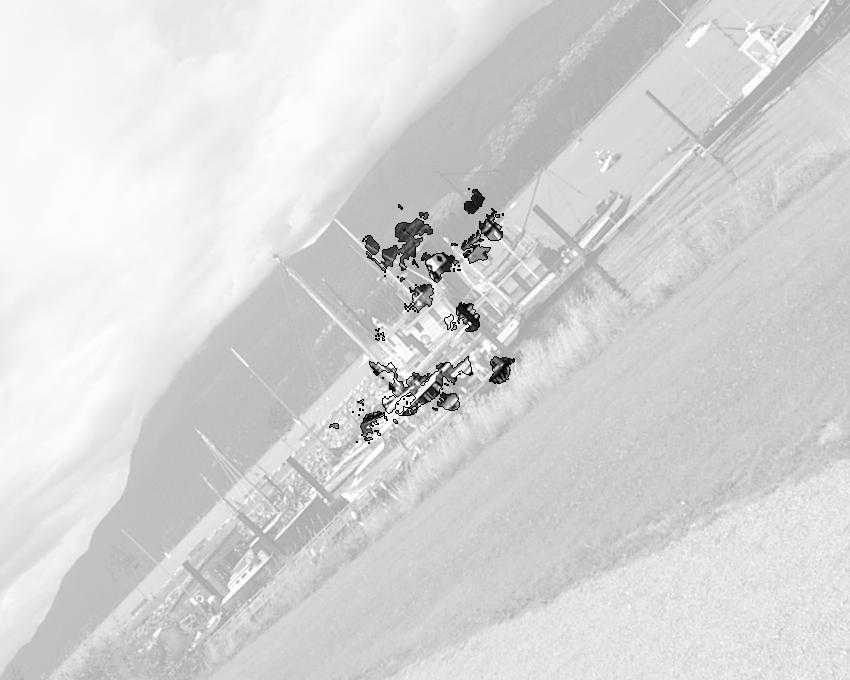} & \includegraphics[height=0.2\linewidth]{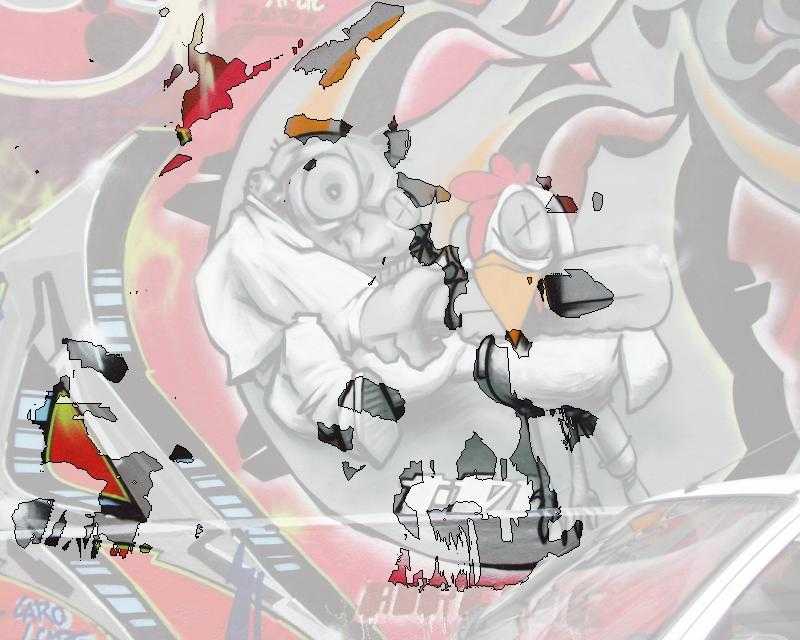}
\includegraphics[height=0.2\linewidth]{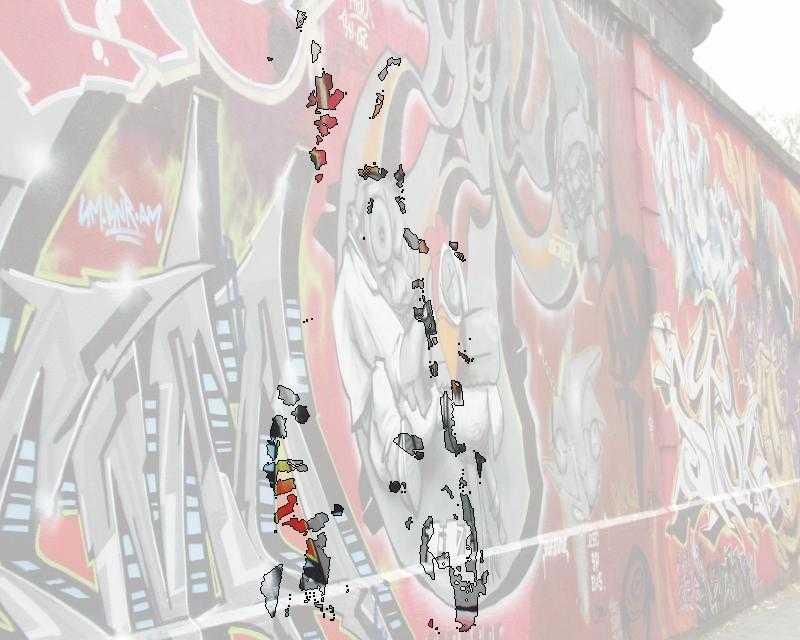} & \includegraphics[height=0.2\linewidth]{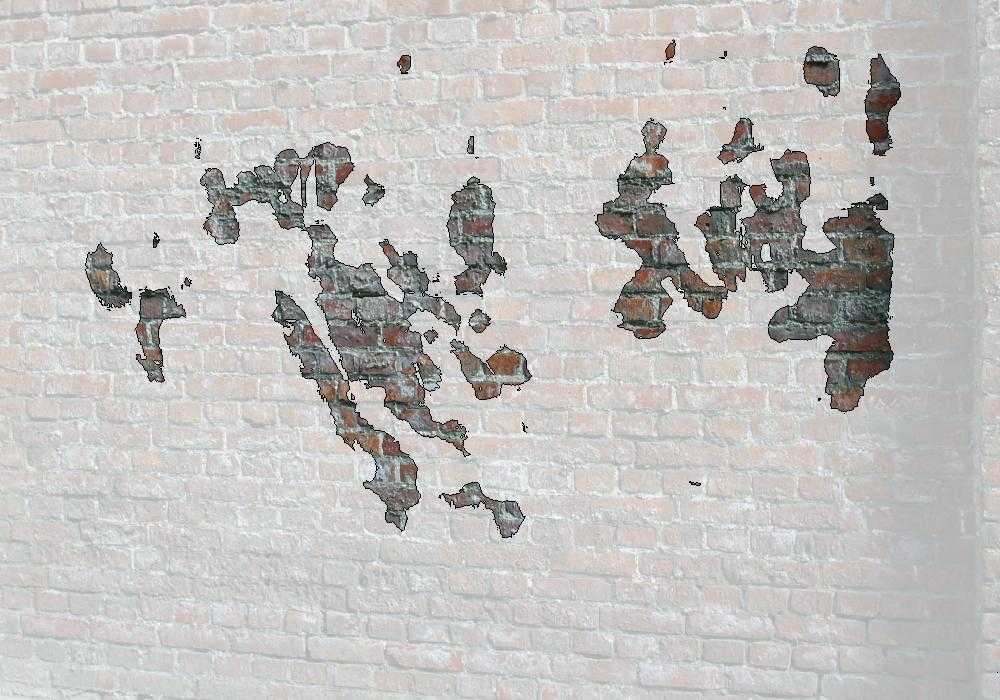}
\includegraphics[height=0.2\linewidth]{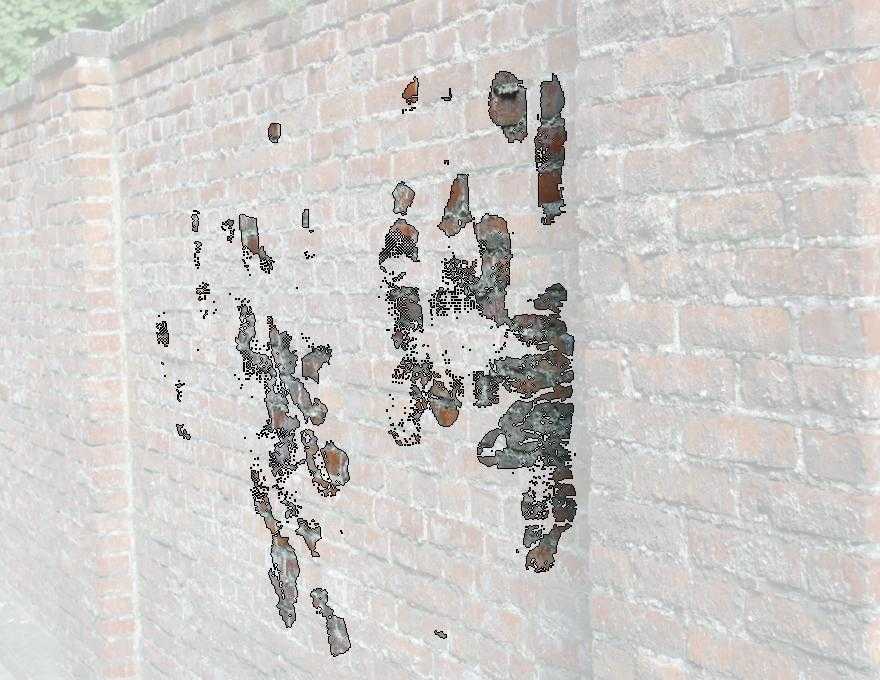}\tabularnewline
\begin{sideways}
{\Large DeepMatching}
\end{sideways} & \includegraphics[height=0.2\linewidth]{{fig2_miko_corres.555772C0505C8BF4_bark_img1-6.555772C0505C8BF4.corres_A}.jpg}
\includegraphics[height=0.2\linewidth]{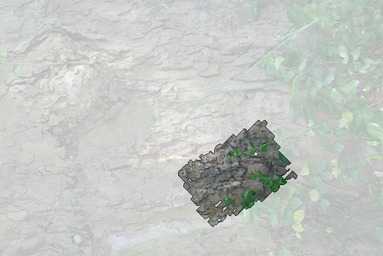} & \includegraphics[height=0.2\linewidth]{{fig2_miko_corres.555772C0505C8BF4_boat_img1-6.555772C0505C8BF4.corres_A}.jpg}
\includegraphics[height=0.2\linewidth]{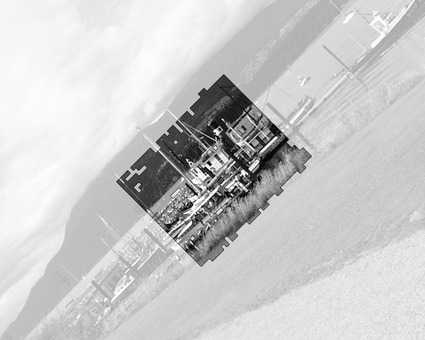}  & \includegraphics[height=0.2\linewidth]{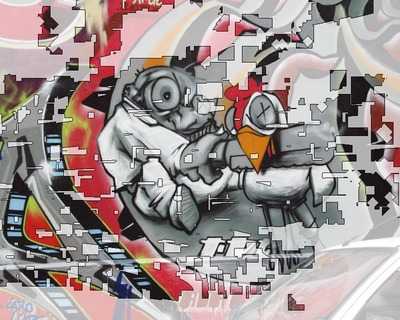}
\includegraphics[height=0.2\linewidth]{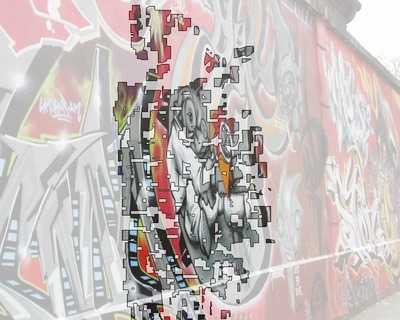} & \includegraphics[height=0.2\linewidth]{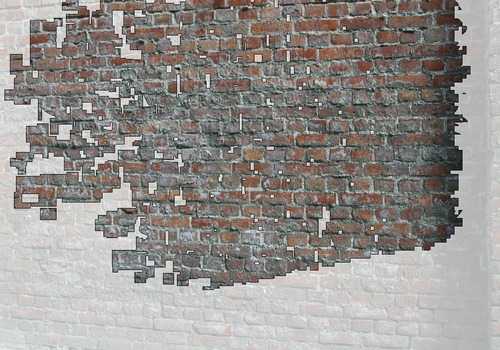}
\includegraphics[height=0.2\linewidth]{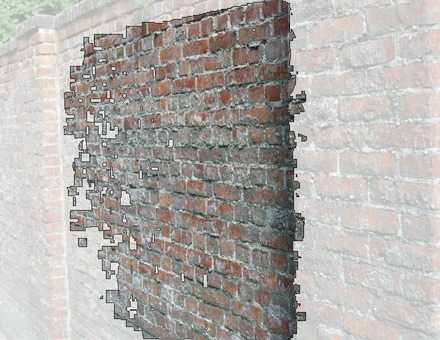}\tabularnewline
\end{tabular}
}
\caption{Comparison of matching results of different methods on the Mikolajczyk dataset. 
Each row shows pixels with correct correspondences for different
methods (from top to bottom: ground-truth, SIFT-NN, GPM, NRDC and DM).  
For each scene, we select two images to match and fade out regions which are 
unmatched, \ie those for which the matching error is above 15px or can not be matched. 
DeepMatching outperforms the other methods, especially on difficult cases like \emph{graf} and \emph{wall}.}
\label{fig:mikoviz}
\end{figure*}

Quantitative results are listed in Table~\ref{tab:matching}, 
and qualitative results in Figures~\ref{fig:mikoviz}, \ref{fig:ex-sintel} and \ref{fig:ex-kitti}. 
Overall, DM significantly outperforms all other methods,
even when reduced settings are used (\eg for image resolution $R=1/2$ and $D=1024$ prototypes). 
As expected, SIFT-NN performs rather well in presence of global image
transformation (Mikolajczyk dataset), but yields the worst result for the case of
more complex motions 
(flow datasets).
Figures~\ref{fig:ex-sintel} and \ref{fig:ex-kitti} illustrate the reason: SIFT's large patches are way too coarse to 
follow motion boundaries precisely. The same issue also holds for HOG-NN.
\red{Methods predicting dense correspondence fields return
a more precise estimate, yet most of them (KPM, GPM, SIFT-flow, DSP) are not robust to 
repetitive textures in the Kitti dataset (Figure~\ref{fig:ex-kitti}) 
as they rely on weakly discriminative small patches. Despite this limitation, SIFT-flow and DSP
are still able to perform well on MPI-Sintel as this dataset contains little scale changes.
Other dense methods, NRDC, SLS and DaisyFF, can handle patches of different sizes
and thus perform better on Kitti. But in turn this is at the cost of 
reduced performance on the MPI-Sintel or Mikolajczyk datasets
(qualitative results are in Figure~\ref{fig:mikoviz}).
In conclusion, DM outperforms all other methods on the 3 datasets,
including DSP which also relies on a hierarchical matching. 
}

In terms of computing resources, DeepMatching with full settings ($R=1/2$, $D=\infty$) 
is one of the most costly method \red{(only SLS and DaisyFF require the same order of
memory and longer run-time)}. 
The scale and rotation invariant version of DM, used for the Mikolajczyk dataset, is slow compared
to most other approaches, due to its sequential processing 
(\ie treating each combination of rotation and scaling sequentially), 
yet yields near perfect results.
However, running DM with reduced settings is very competitive to the other approaches. 
On MPI-Sintel and Kitti, for instance,
DM with a quarter resolution has a run-time comparable to the fastest method, 
SIFT-NN, with a reasonable memory usage, while still outperforming \red{nearly} all methods
in terms of the accuracy@10 measure.

\newcommand{\imgs}[1]{\includegraphics[width=0.4\linewidth]{fig2_sintel_#1}}
\newcommand{\imgk}[1]{\includegraphics[bb=100px 0px 1126px 262px,clip,width=0.6\linewidth]{fig2_kitti_#1}}
\newcommand{\tbtxt}[1]{{\LARGE #1}}
\newcommand{\tbtxtrot}[1]{\begin{turn}{90}\tbtxt{#1}\end{turn}}

\begin{figure*}
\resizebox{\linewidth}{!}{
\begin{tabular}{ccc||cc||ccc}
  & 
  \tbtxt{Correspondence field} & \tbtxt{Image/Error map} & 
  \tbtxt{Correspondence field} & \tbtxt{Image/Error map} & 
  \tbtxt{Correspondence field} & \tbtxt{Image/Error map} & \tabularnewline
\tbtxtrot{~~~~~~GT}  & 
  \imgs{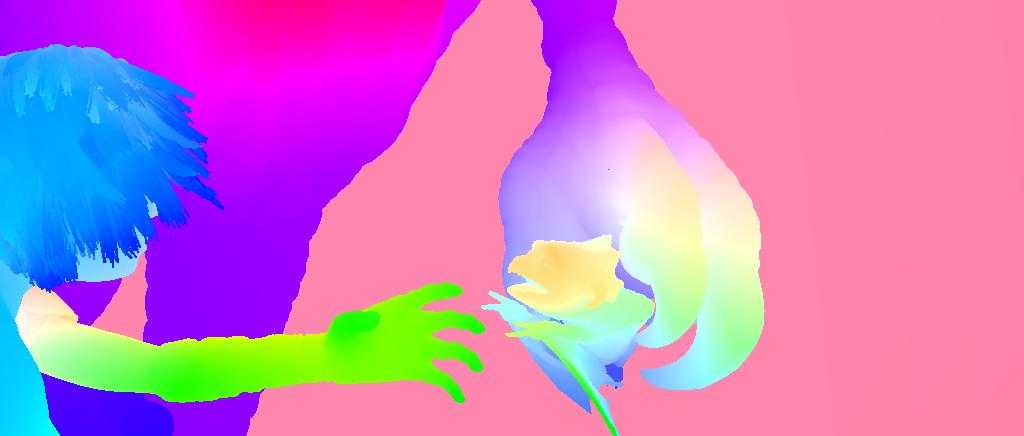} &
  \imgs{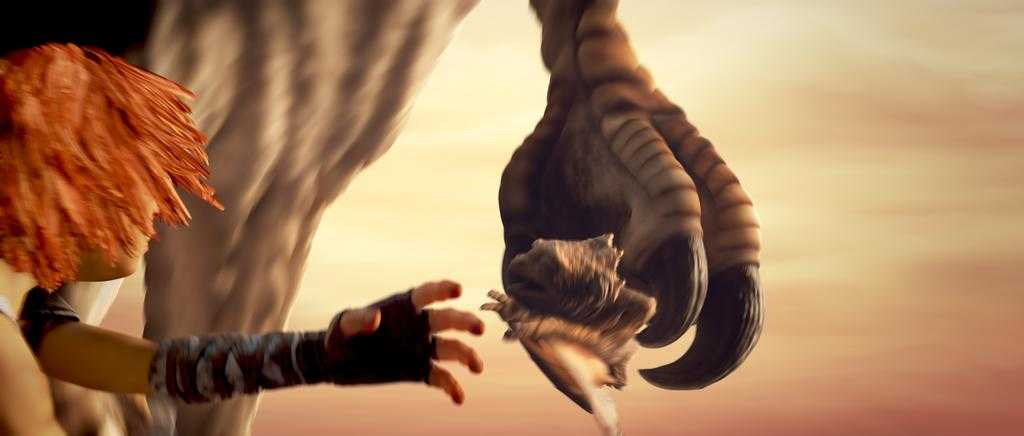}
  & 
  \imgs{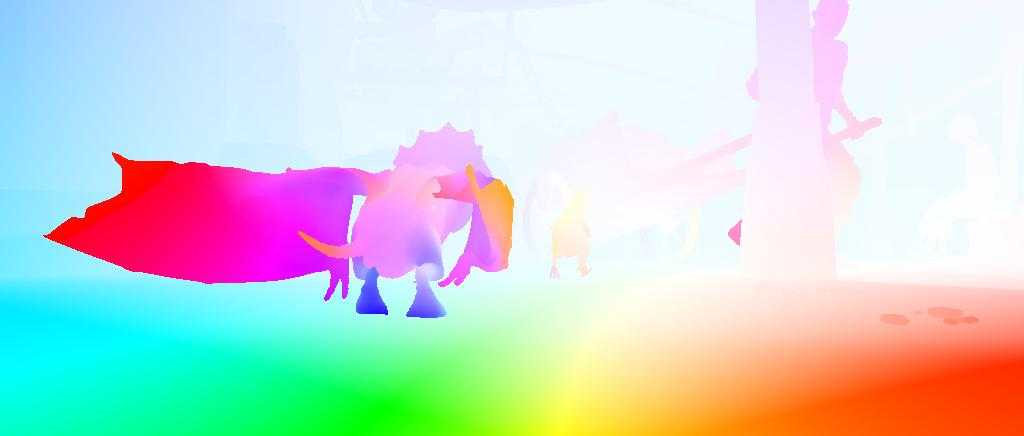} &
  \imgs{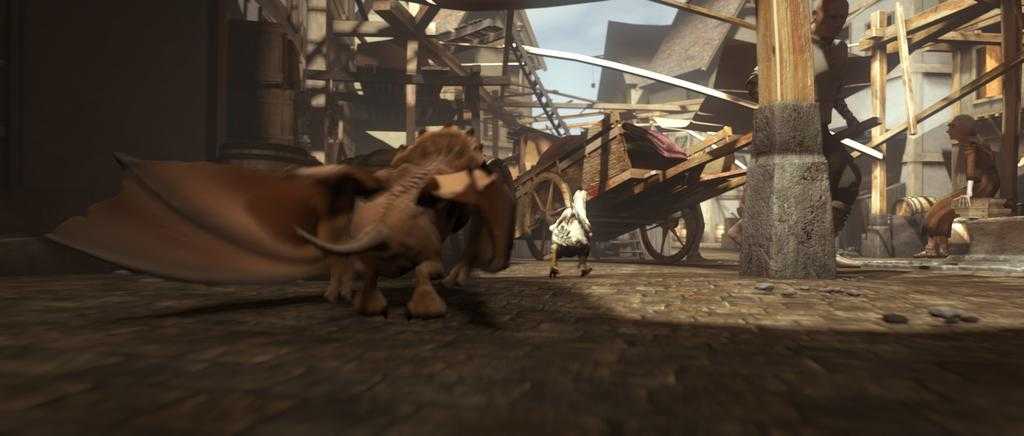}
  & 
  \imgs{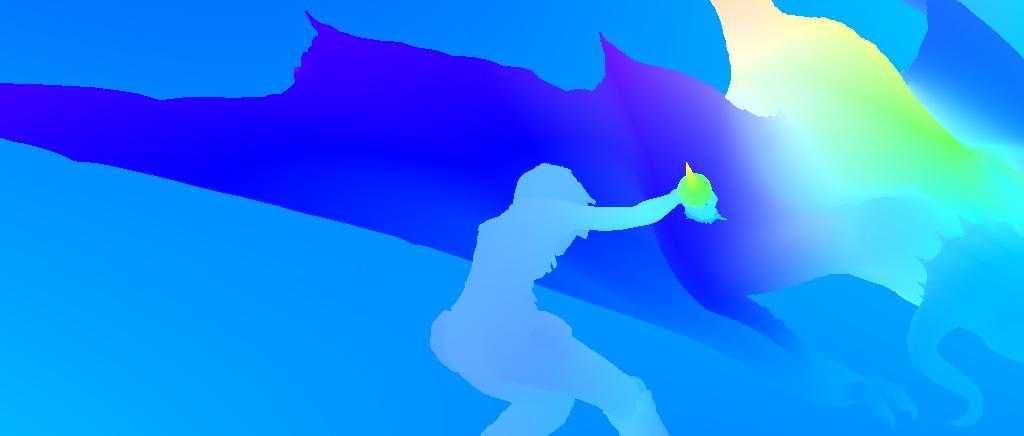} &
  \imgs{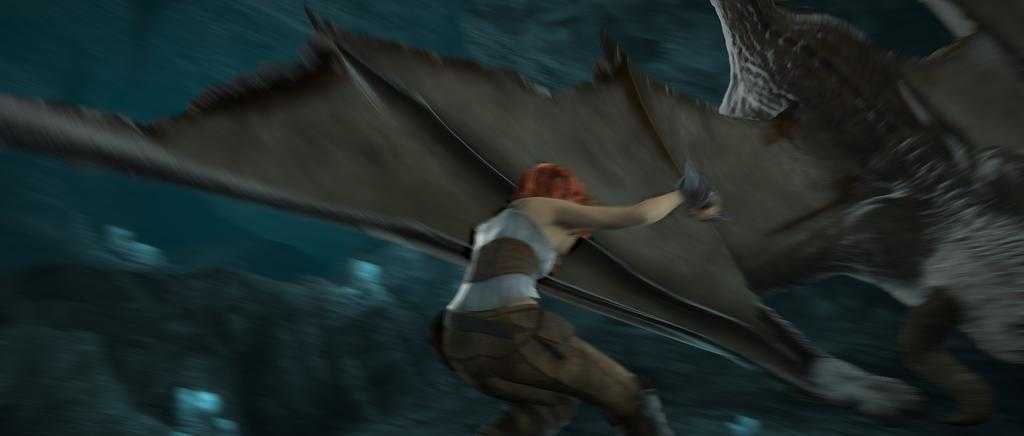} &
  \tabularnewline
\tbtxtrot{~SIFT-NN} &
  \imgs{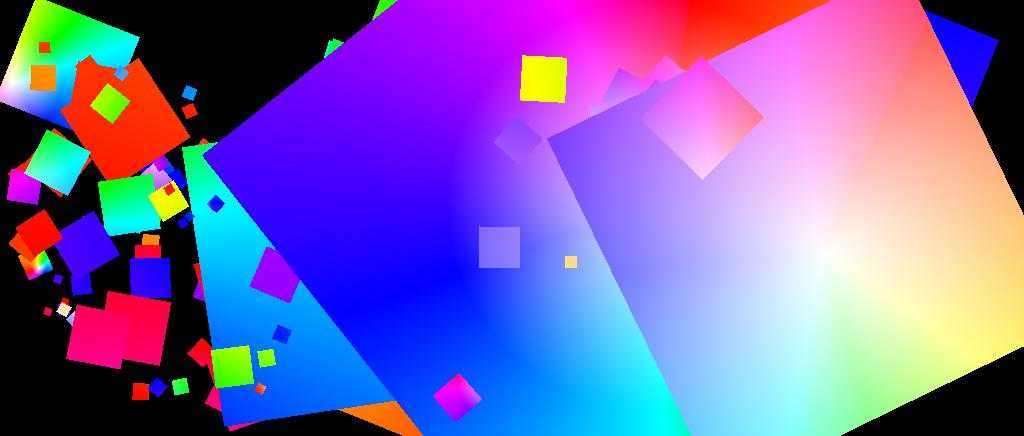} &
  \imgs{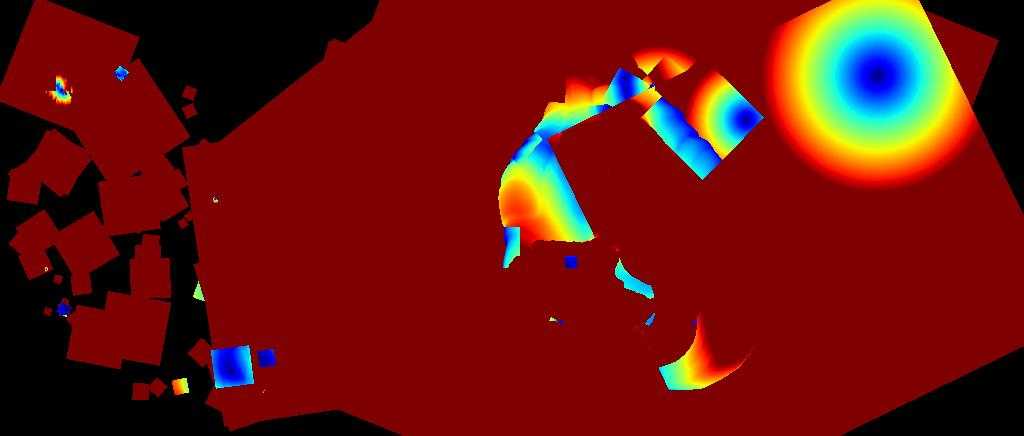}
  & 
  \imgs{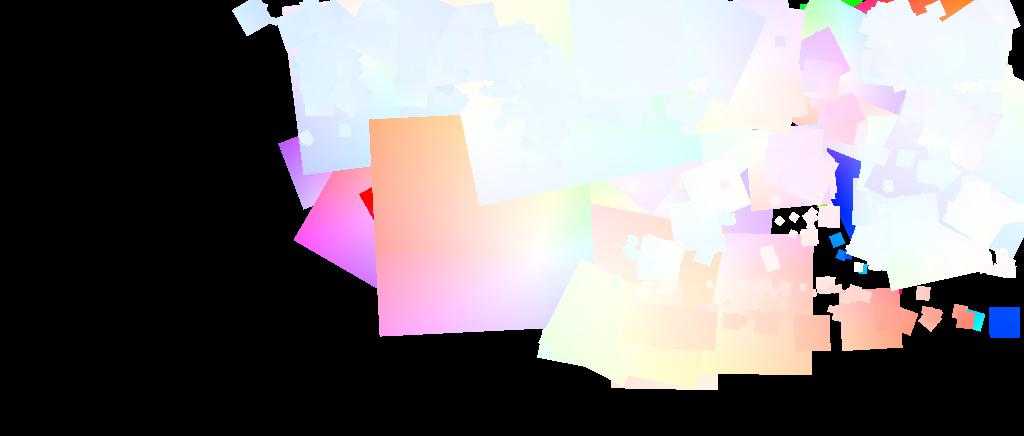} &
  \imgs{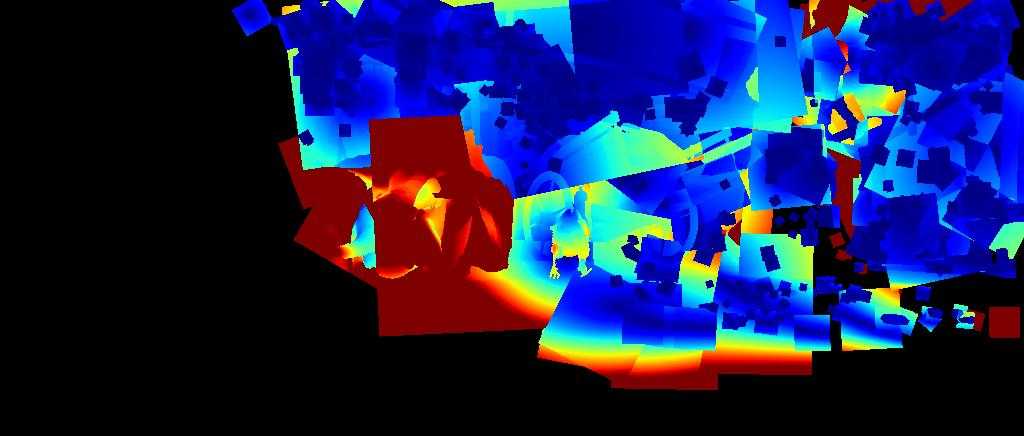}
  & 
  \imgs{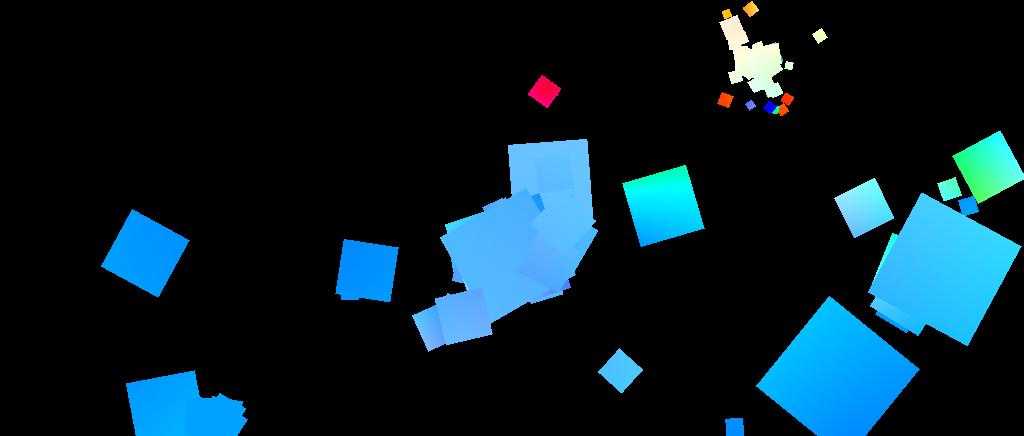} &
  \imgs{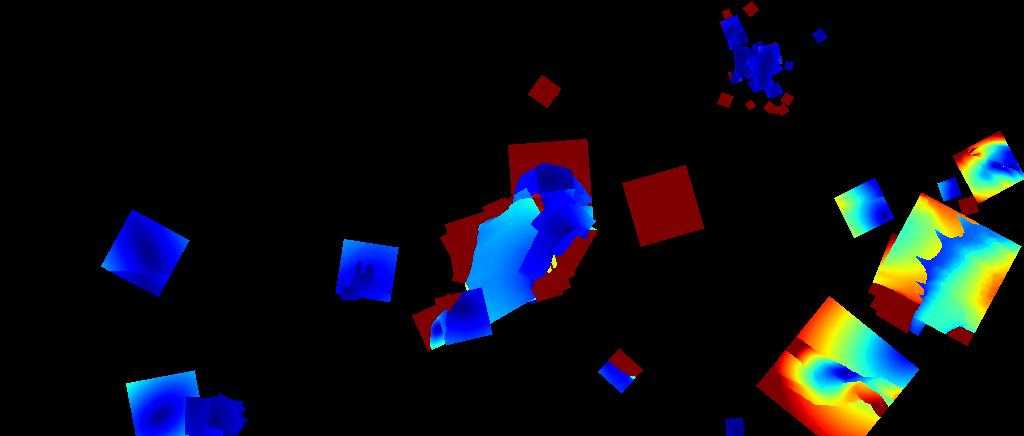} &
  \tabularnewline
\tbtxtrot{~HOG-NN} &
  \imgs{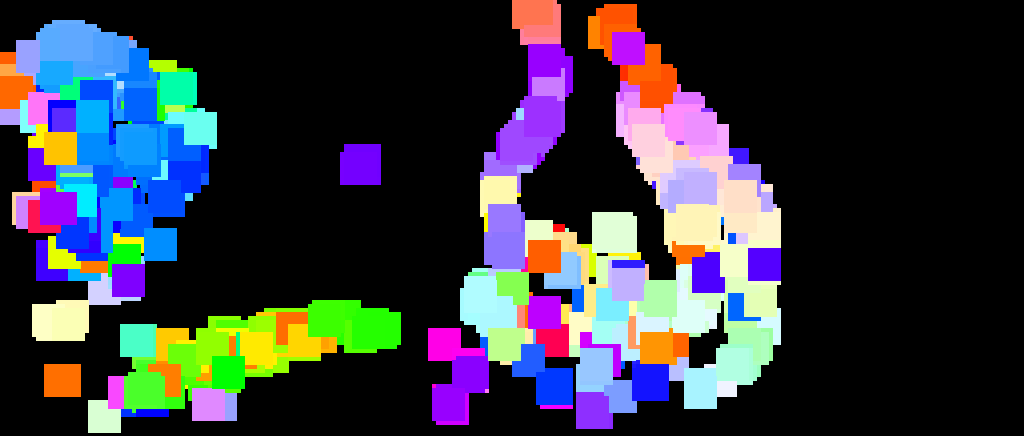} &
  \imgs{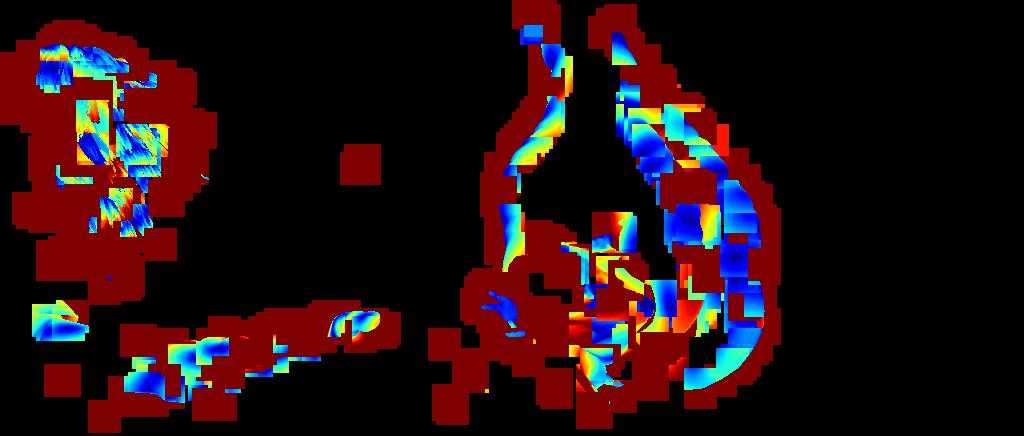}
  & 
  \imgs{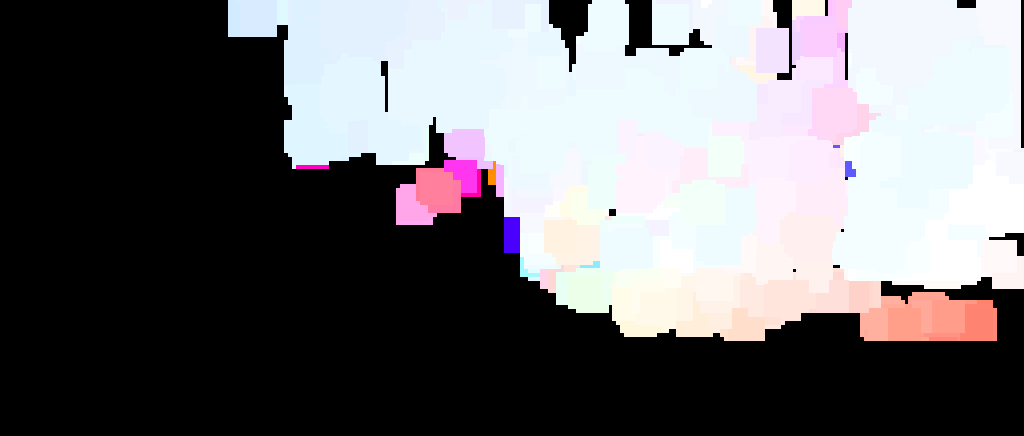} &
  \imgs{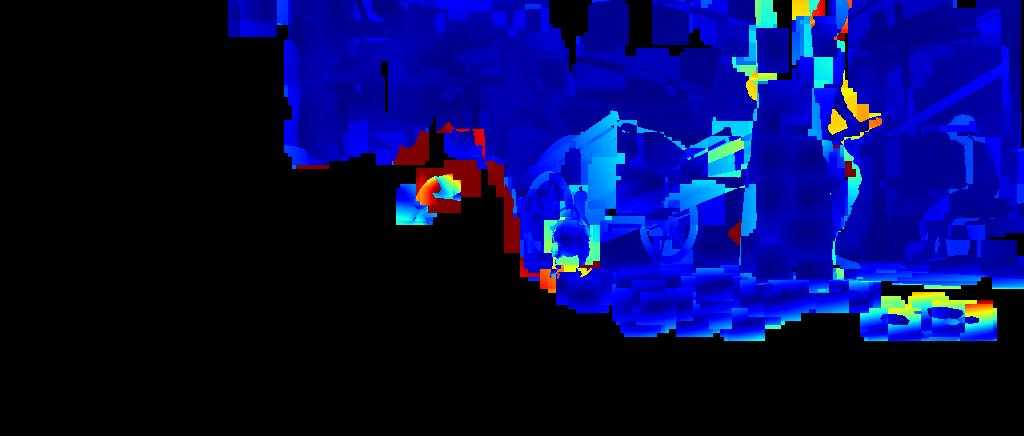}
  & 
  \imgs{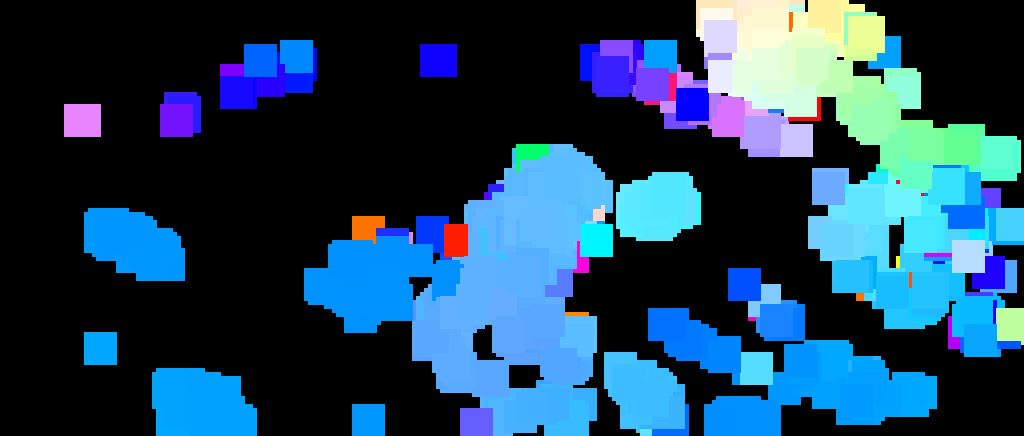} &
  \imgs{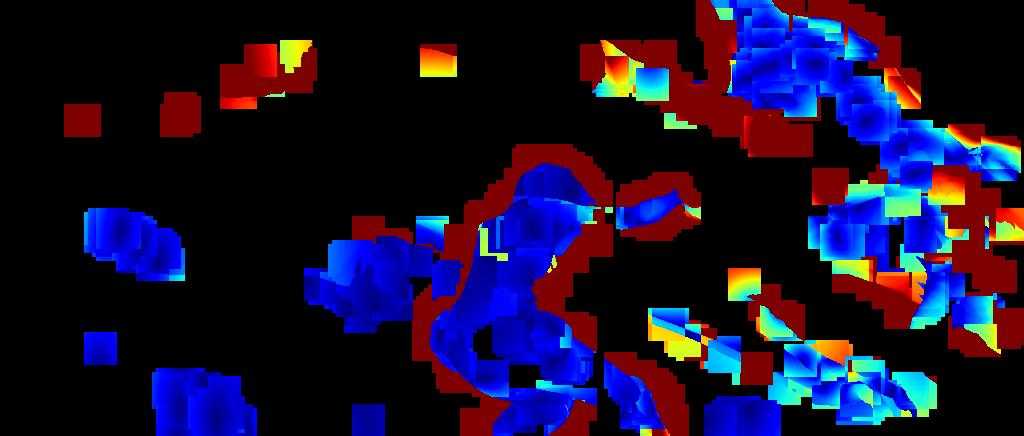} &
  \tabularnewline
\tbtxtrot{~~~~KPM} &
  \imgs{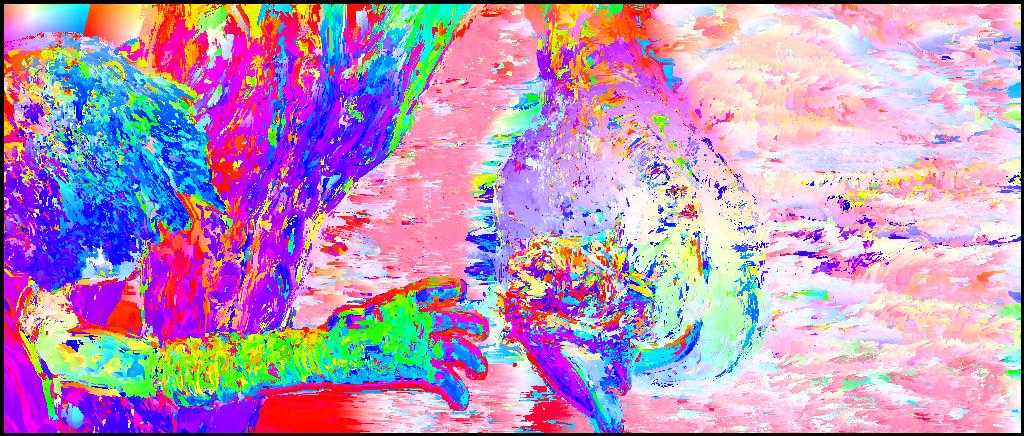} &
  \imgs{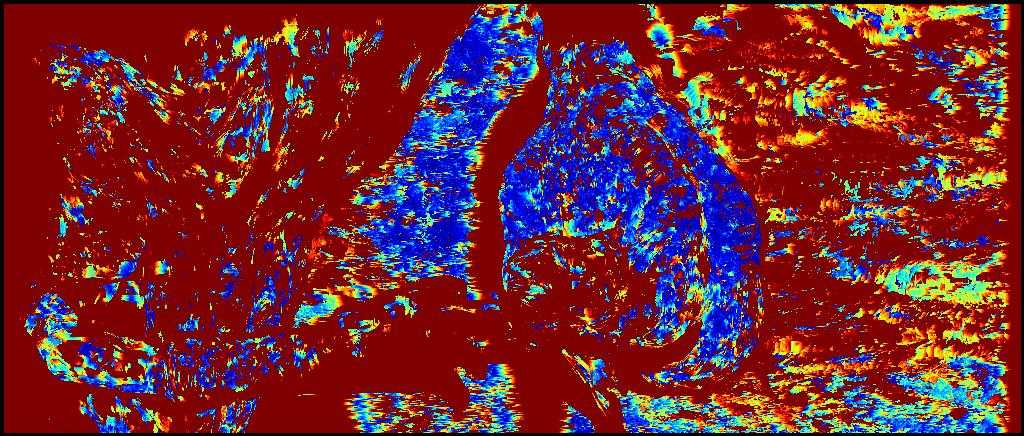}
  & 
  \imgs{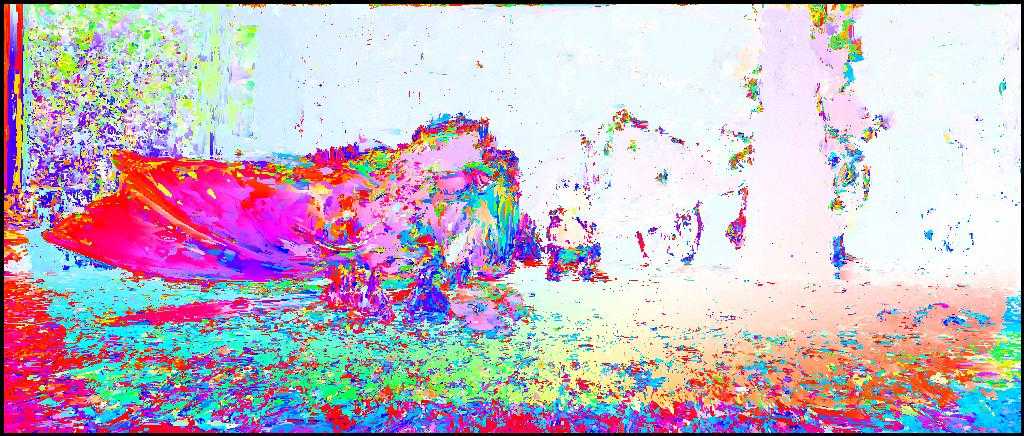} &
  \imgs{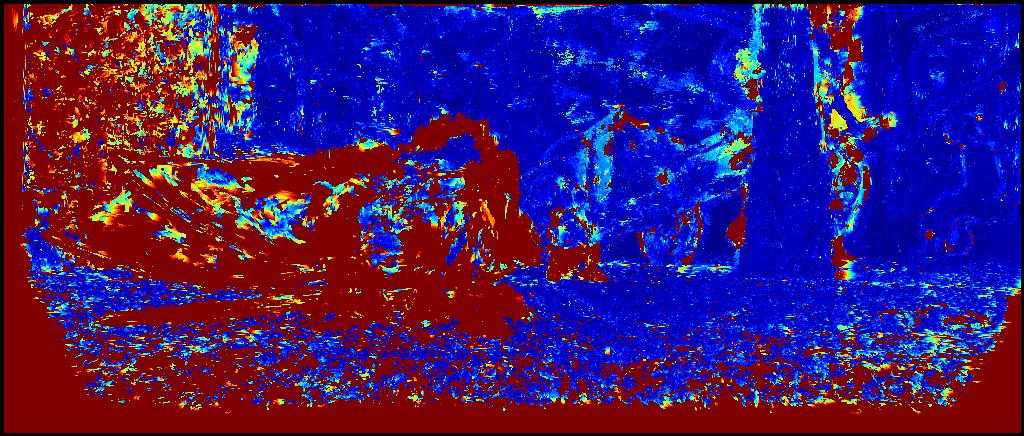}
  & 
  \imgs{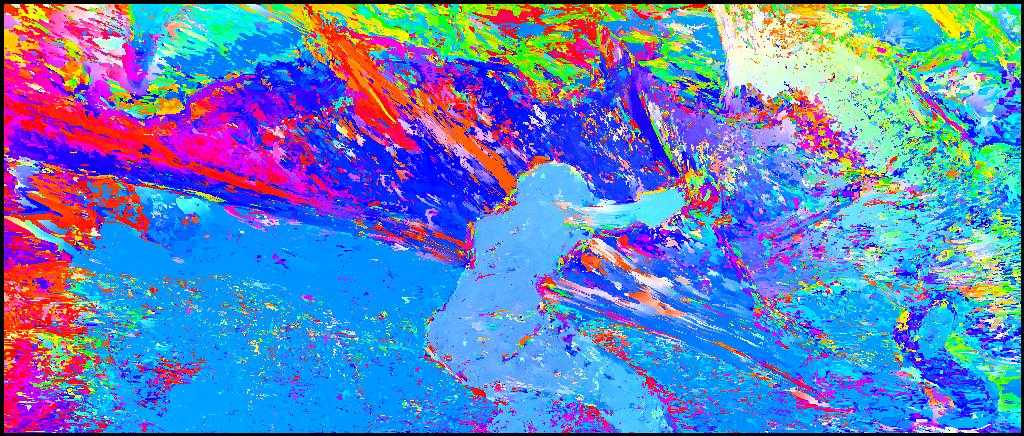} &
  \imgs{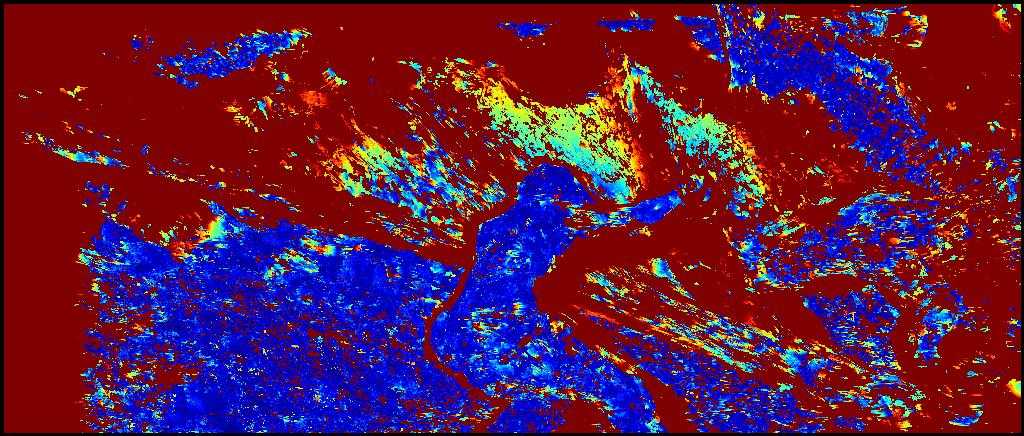} &
  \multirow{7}{*}{\includegraphics[height=75mm]{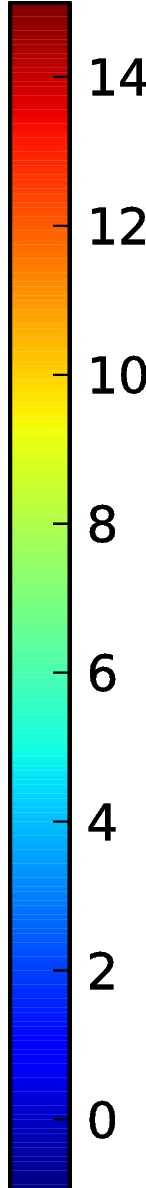}}
  \tabularnewline
\tbtxtrot{~~~~GPM} &
  \imgs{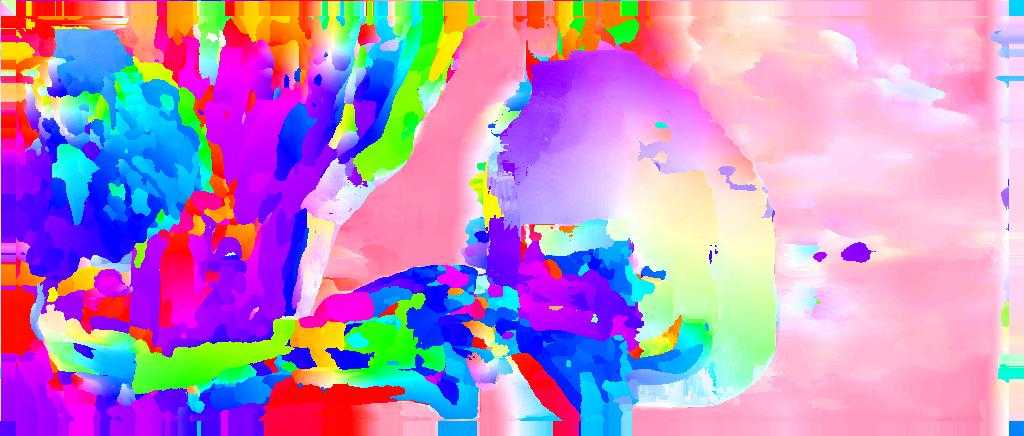} &
  \imgs{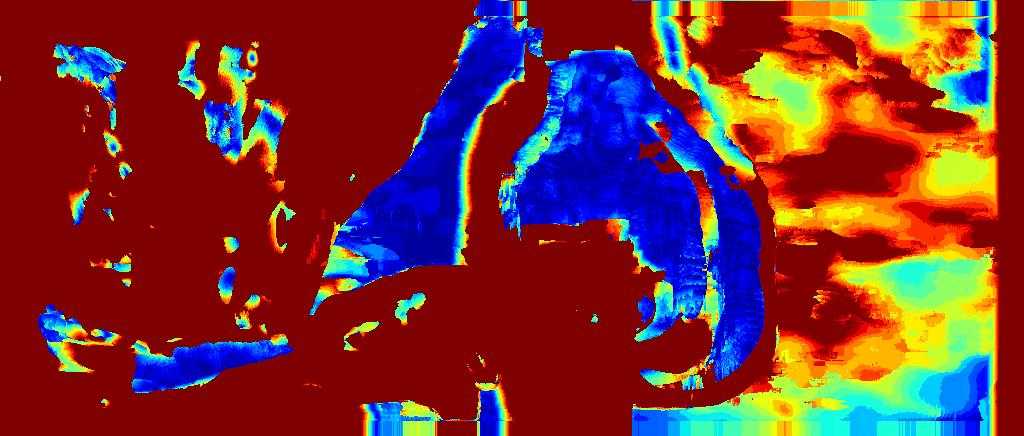}
  & 
  \imgs{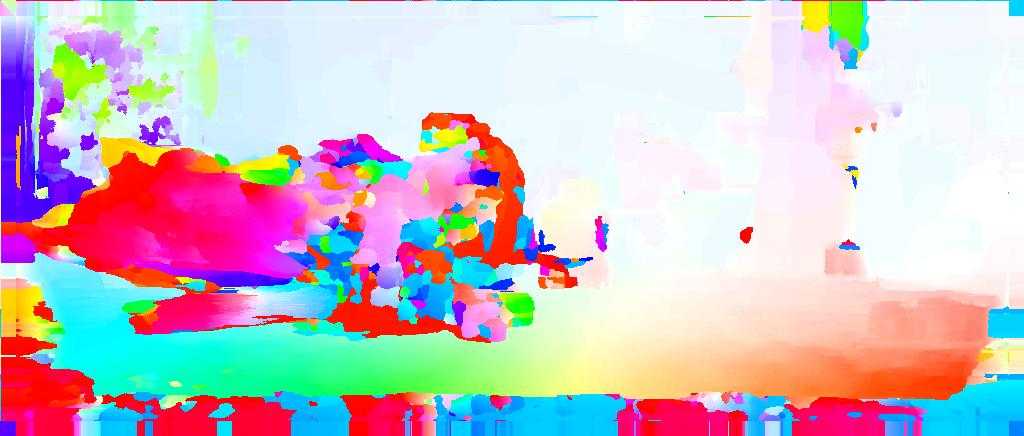} &
  \imgs{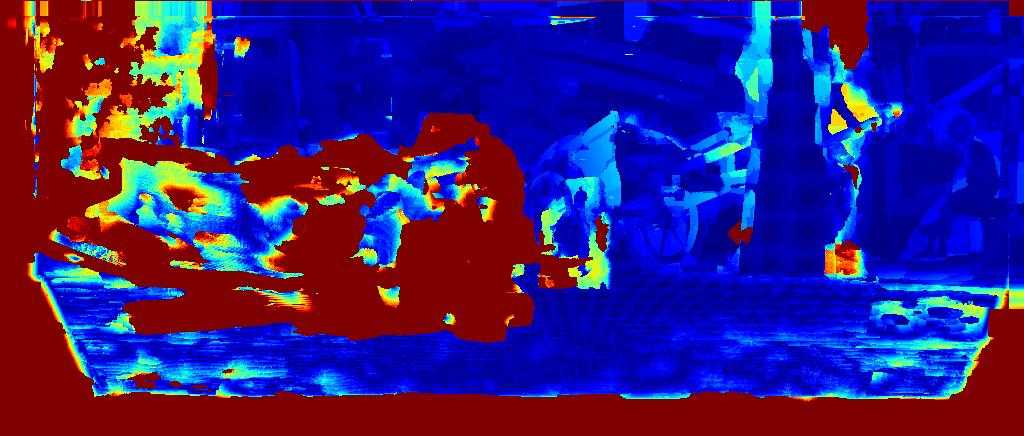}
  & 
  \imgs{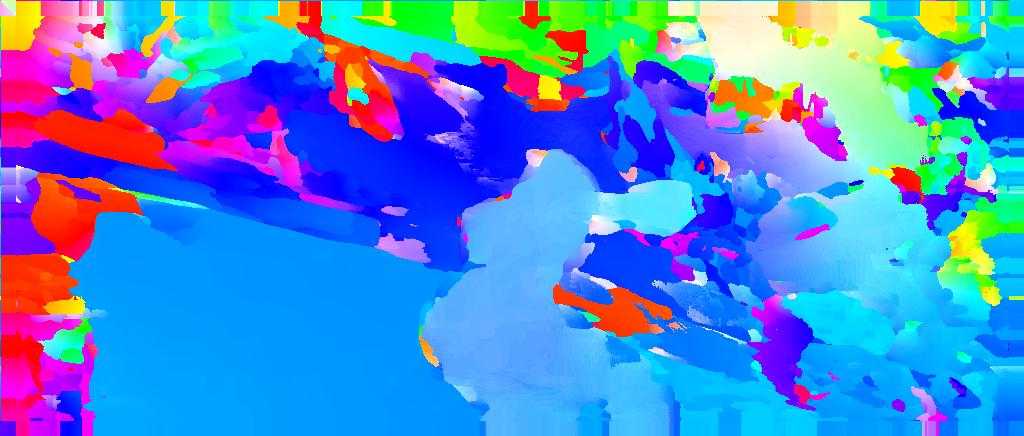} &
  \imgs{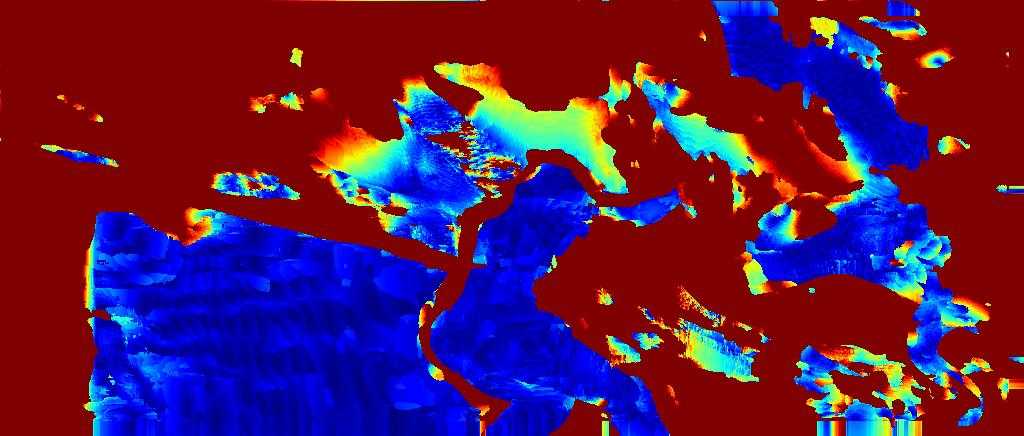} &
  \tabularnewline
\tbtxtrot{SIFT-flow} &
  \imgs{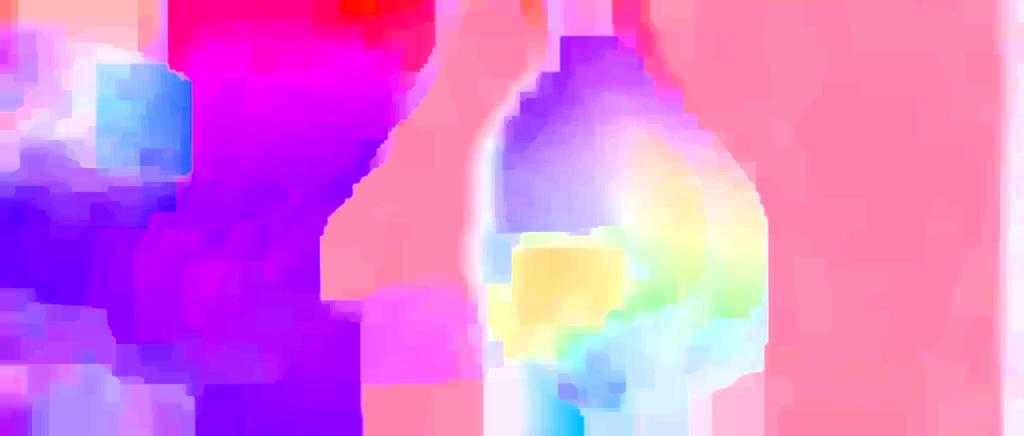} &
  \imgs{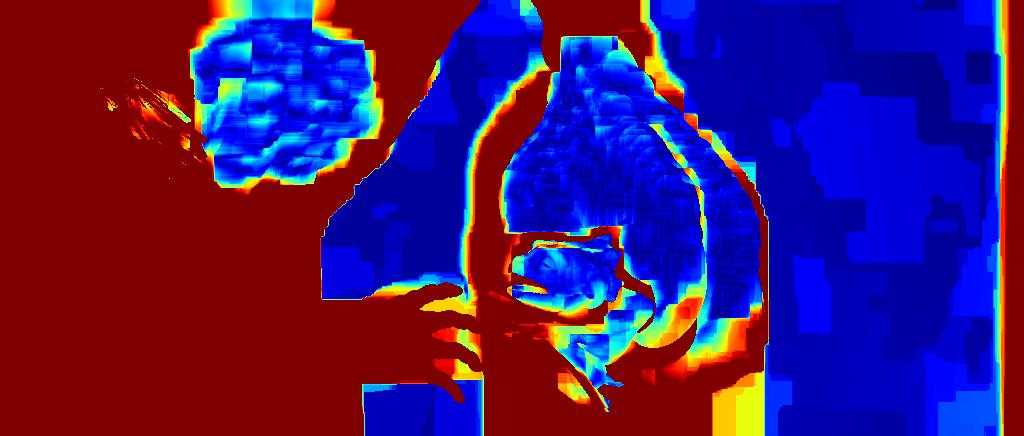}
  & 
  \imgs{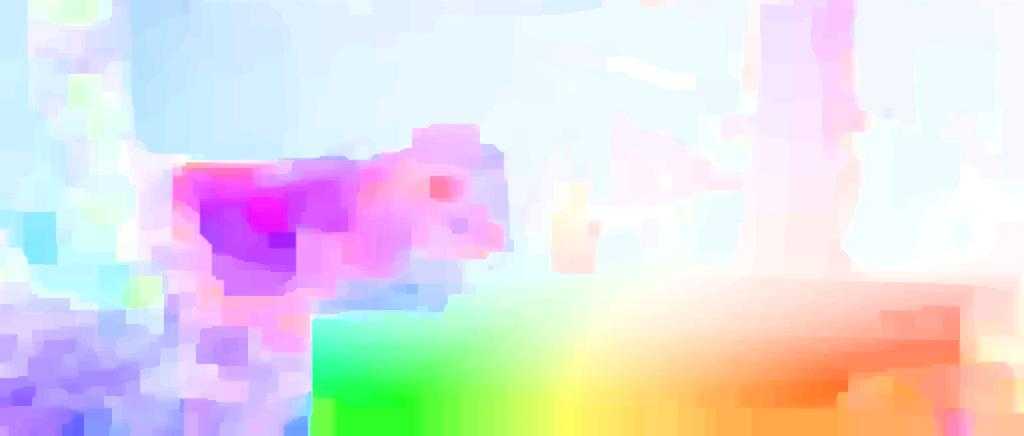} &
  \imgs{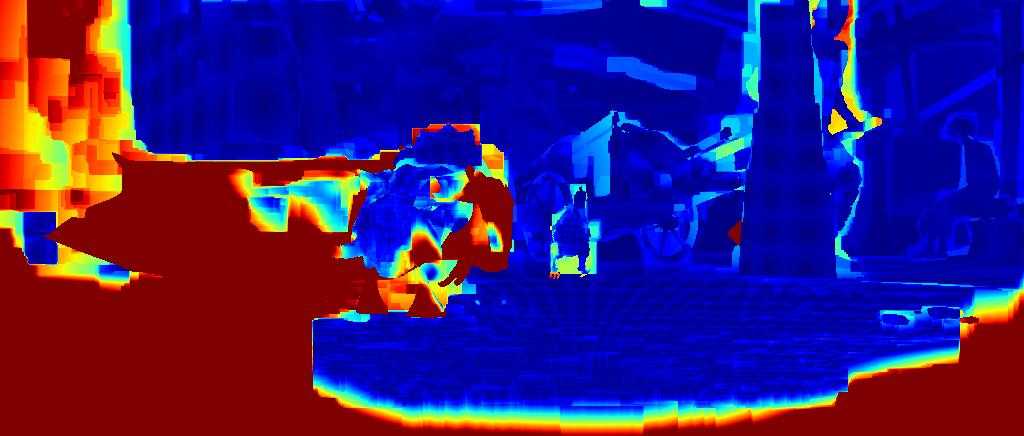}
  & 
  \imgs{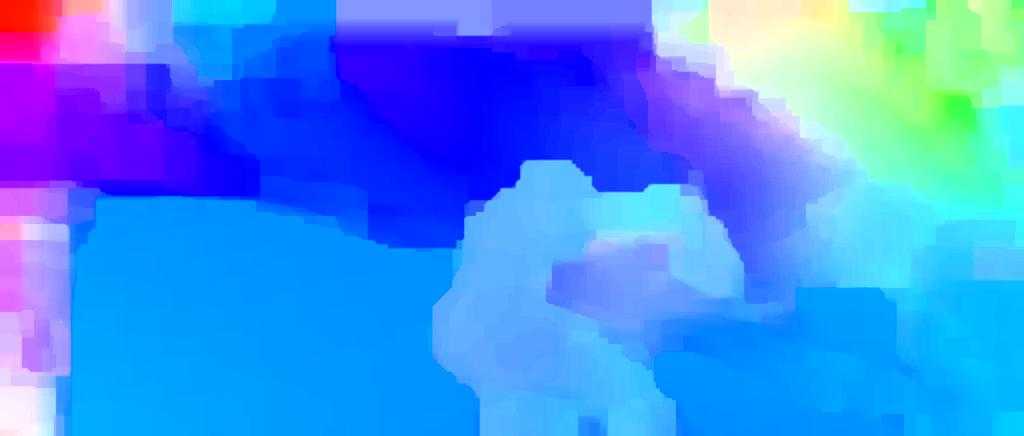} &
  \imgs{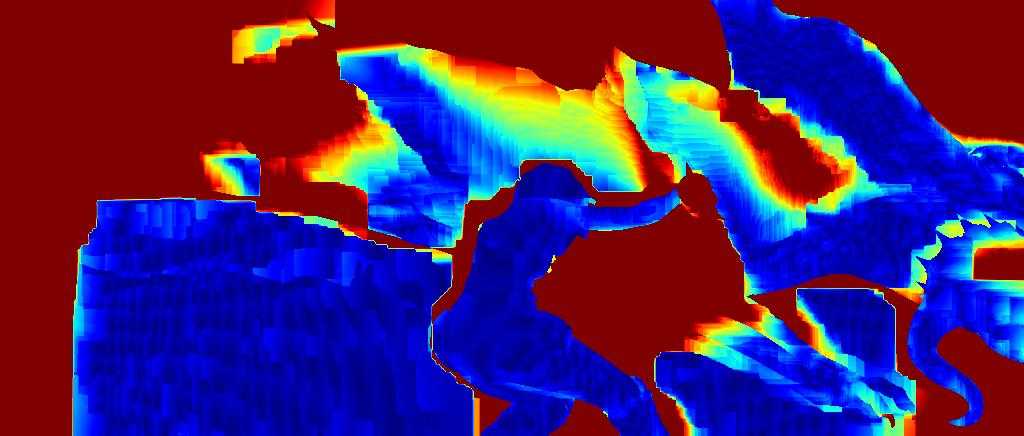} &
  \tabularnewline
\tbtxtrot{~~~~~SLS} &
  \imgs{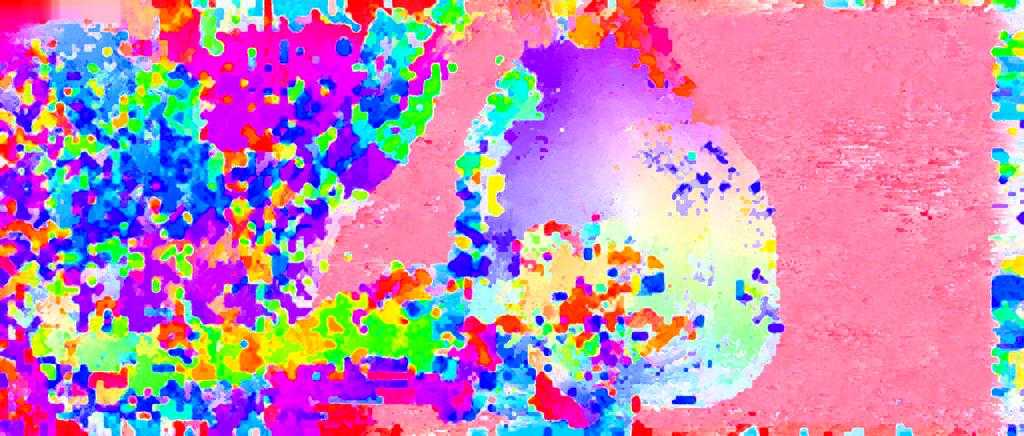} &
  \imgs{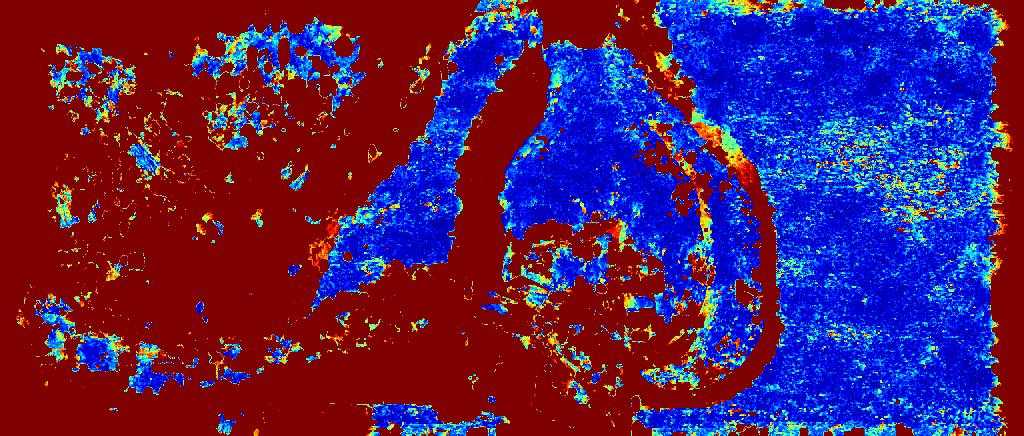}
  & 
  \imgs{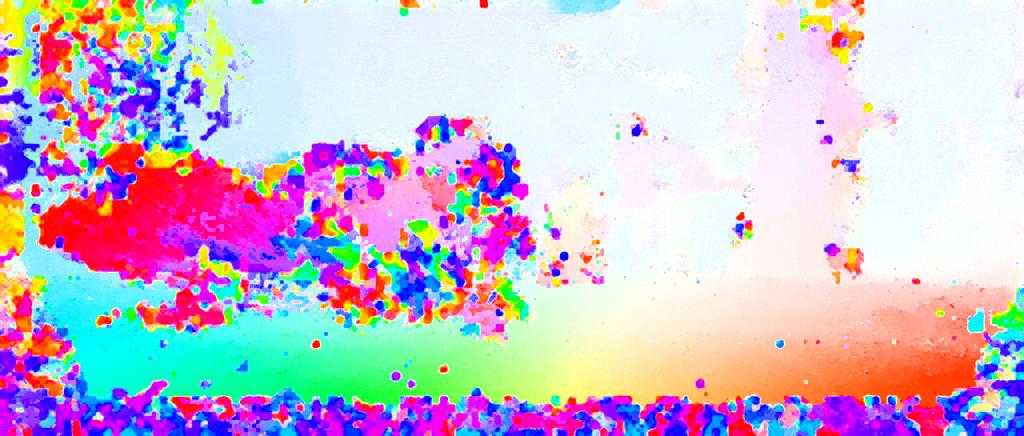} &
  \imgs{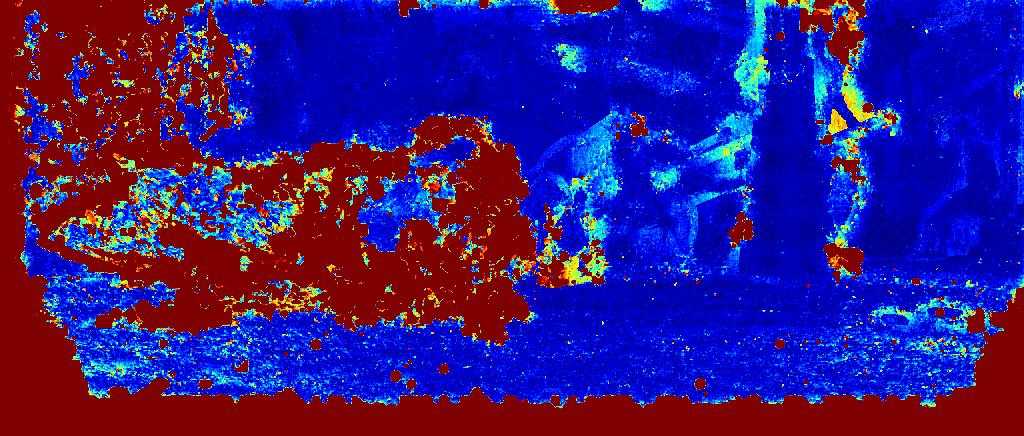}
  & 
  \imgs{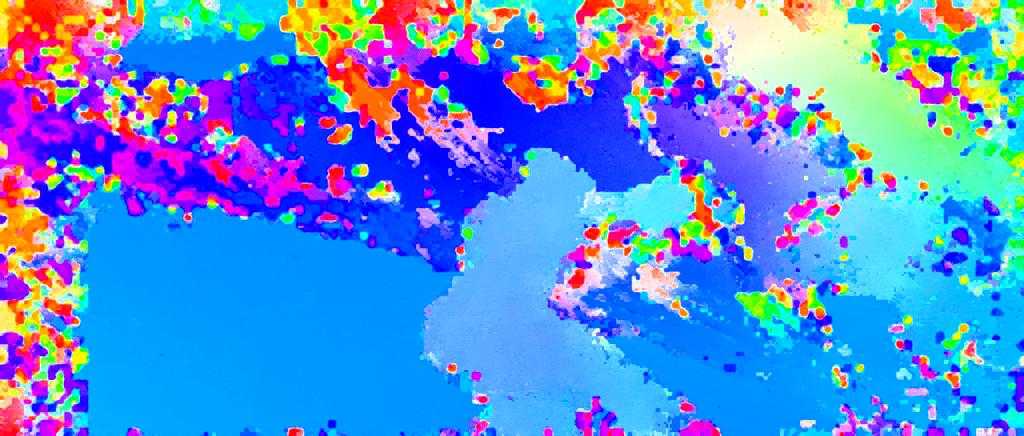} &
  \imgs{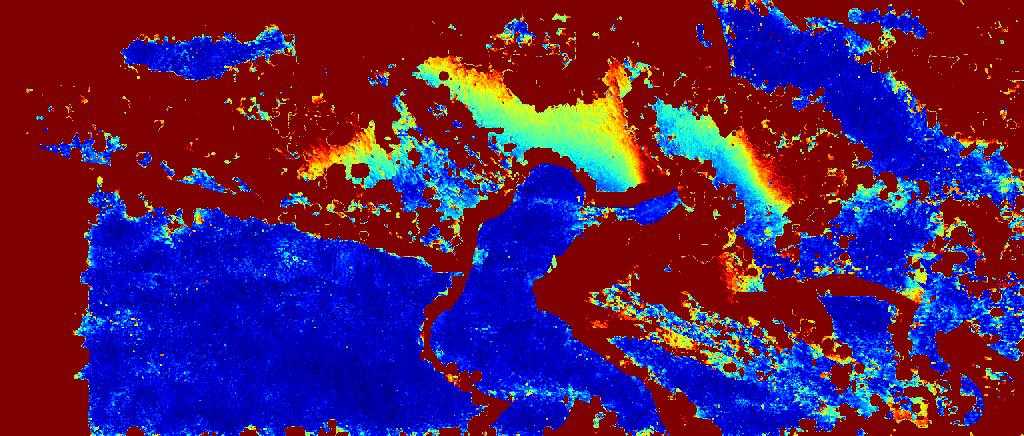} &
  \tabularnewline
\tbtxtrot{~~~DaisyFF} &
  \imgs{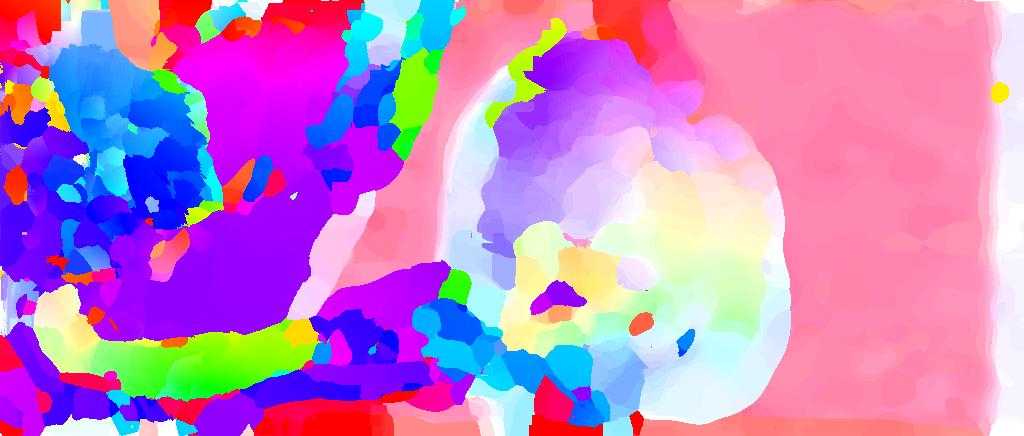} &
  \imgs{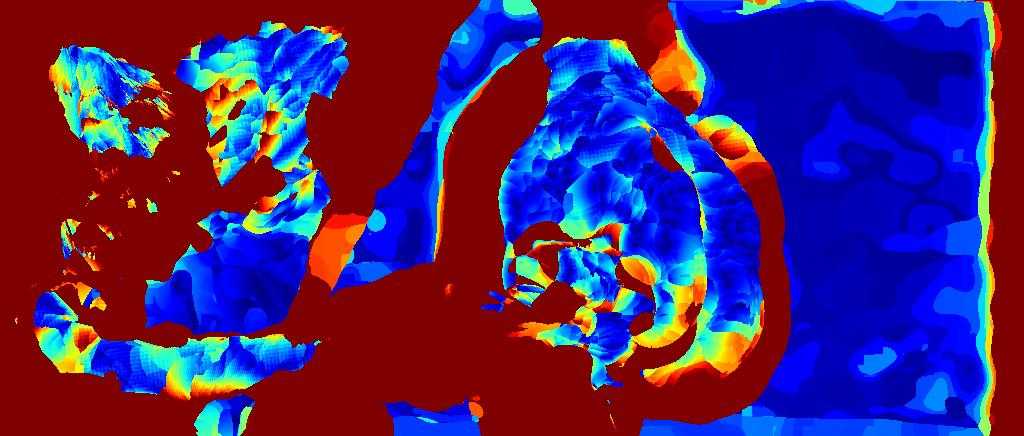}
  & 
  \imgs{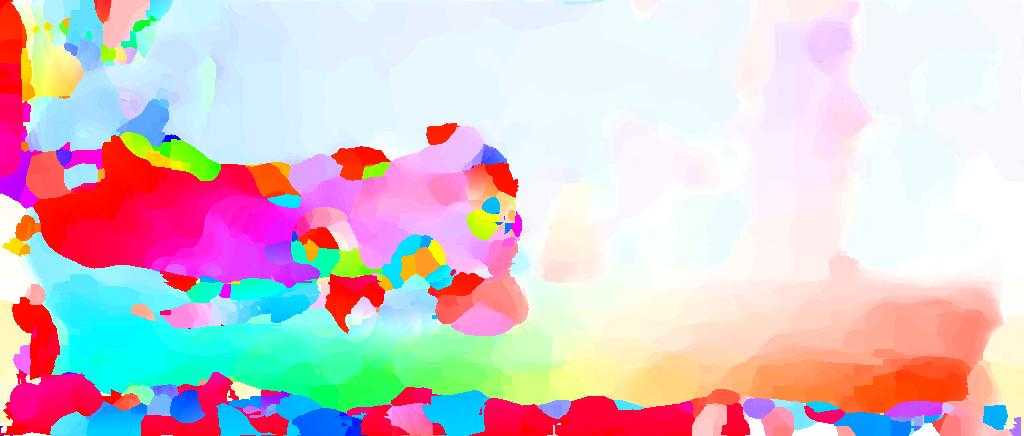} &
  \imgs{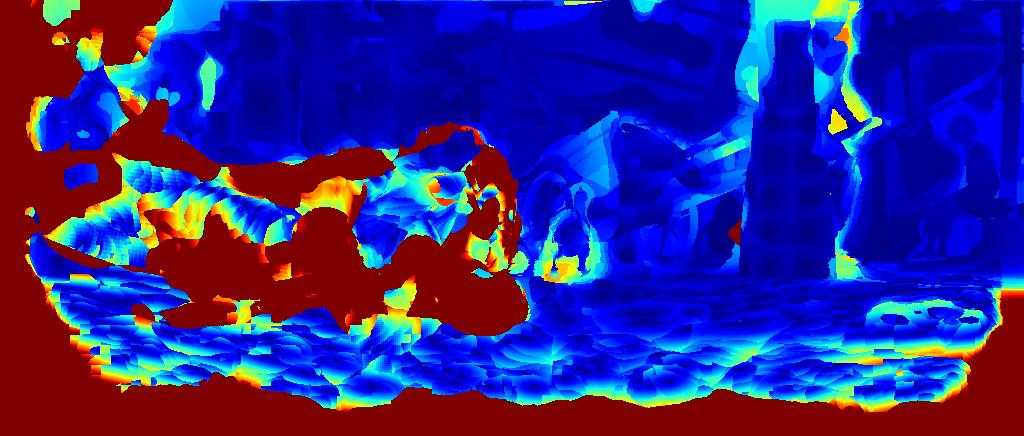}
  & 
  \imgs{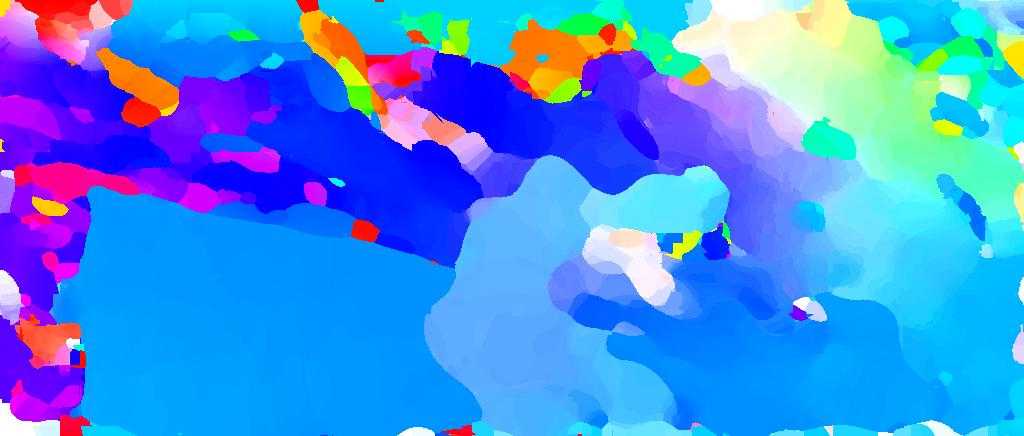} &
  \imgs{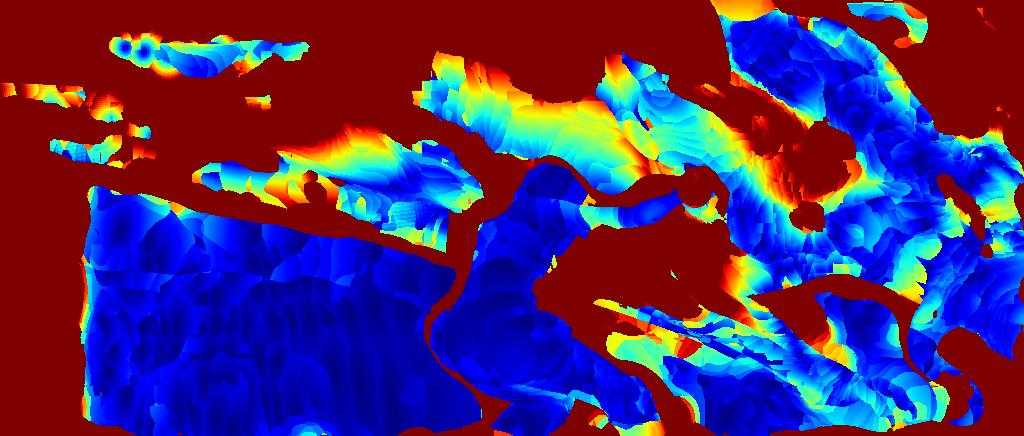} &
  \tabularnewline
\tbtxtrot{~~~~DSP} &
  \imgs{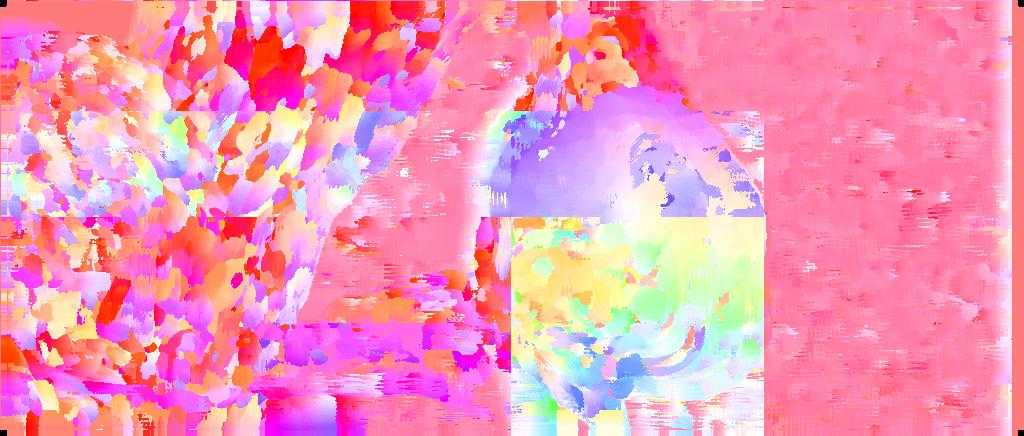} &
  \imgs{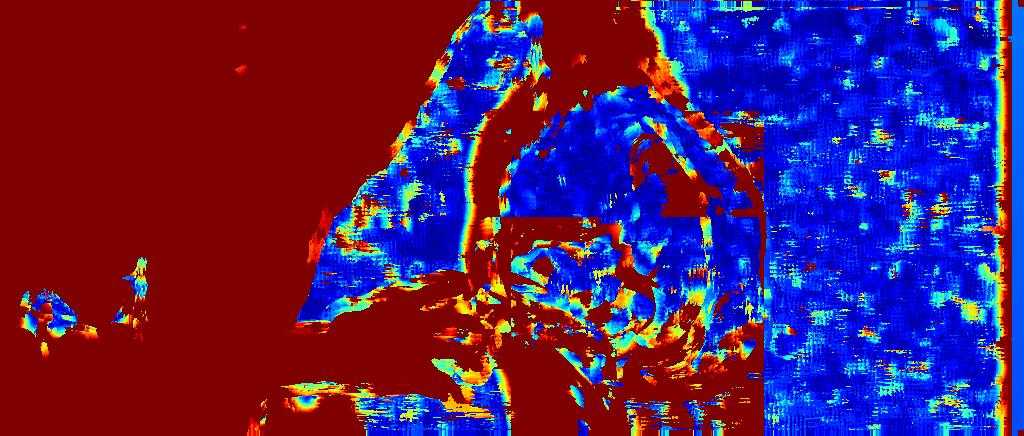}
  & 
  \imgs{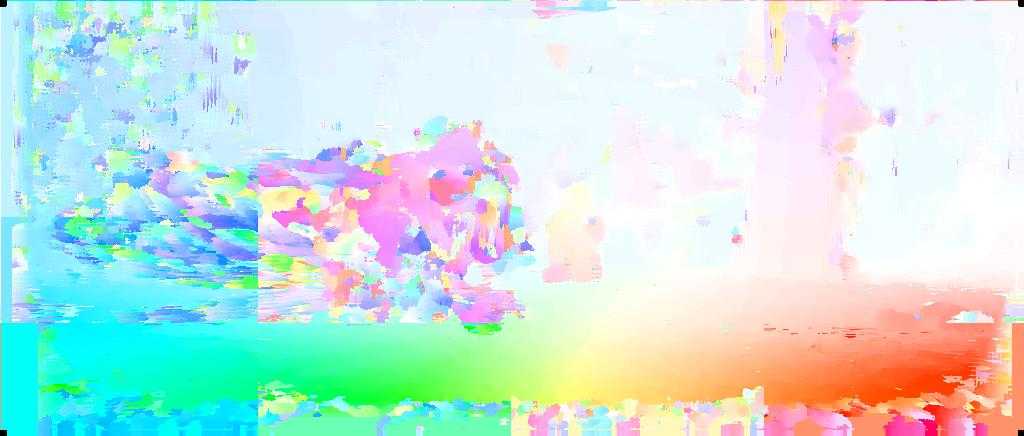} &
  \imgs{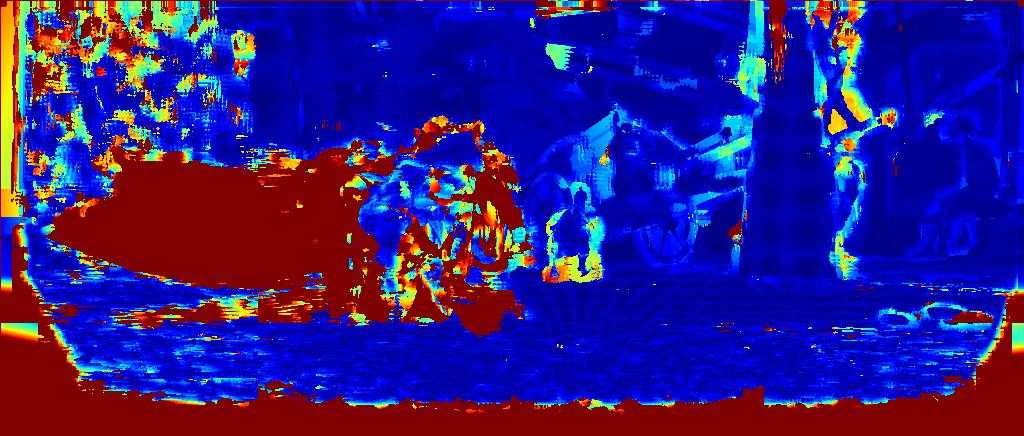}
  & 
  \imgs{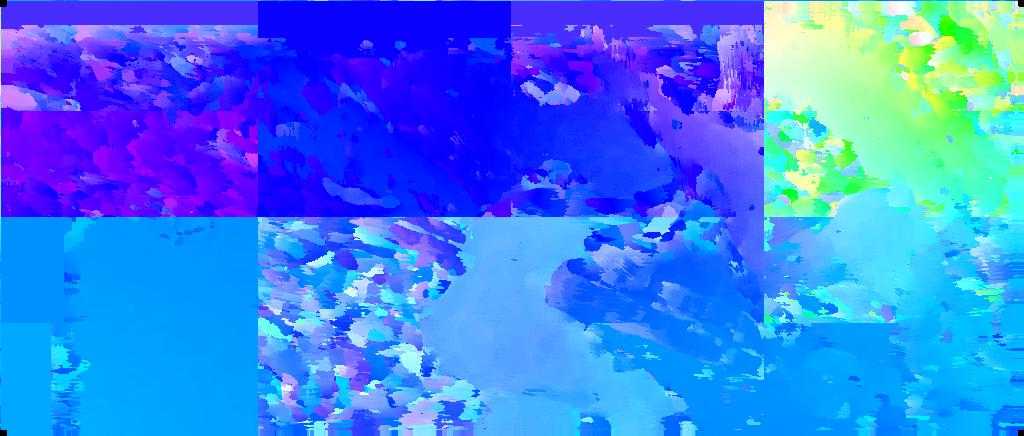} &
  \imgs{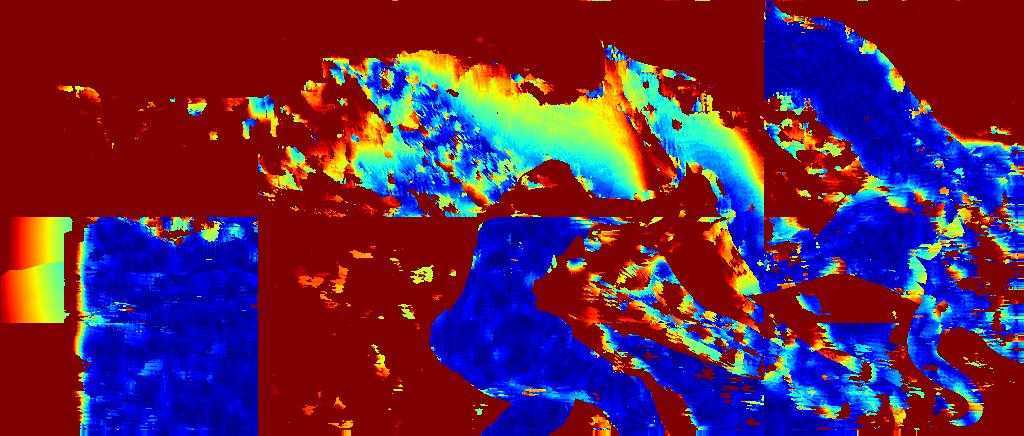} &
  \tabularnewline
\tbtxtrot{~~~~~\textbf{DM}} &
  \imgs{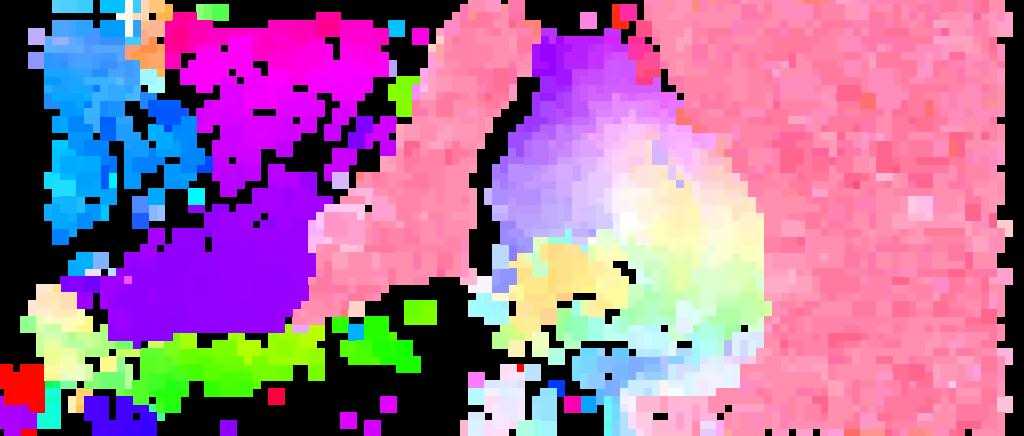} &
  \imgs{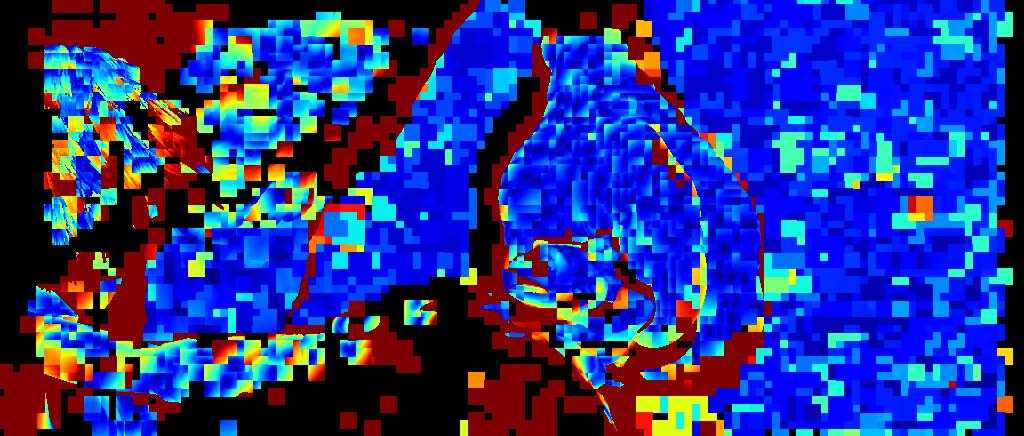}
  & 
  \imgs{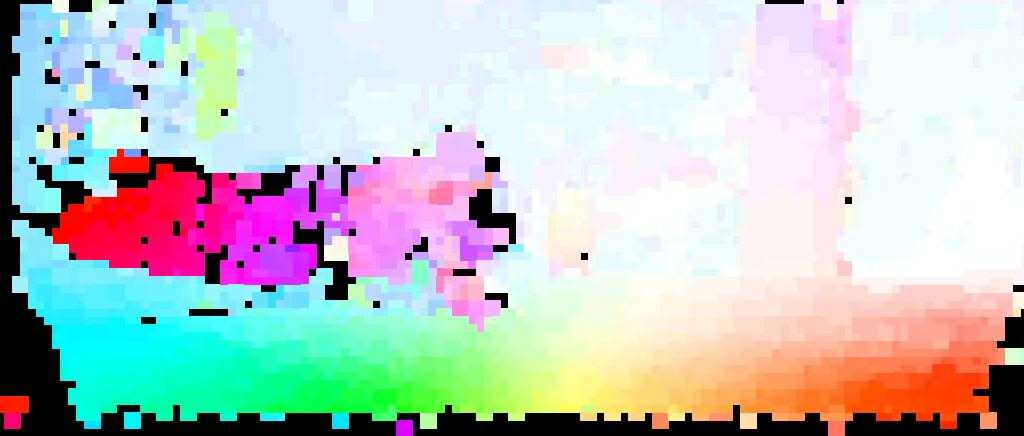} &
  \imgs{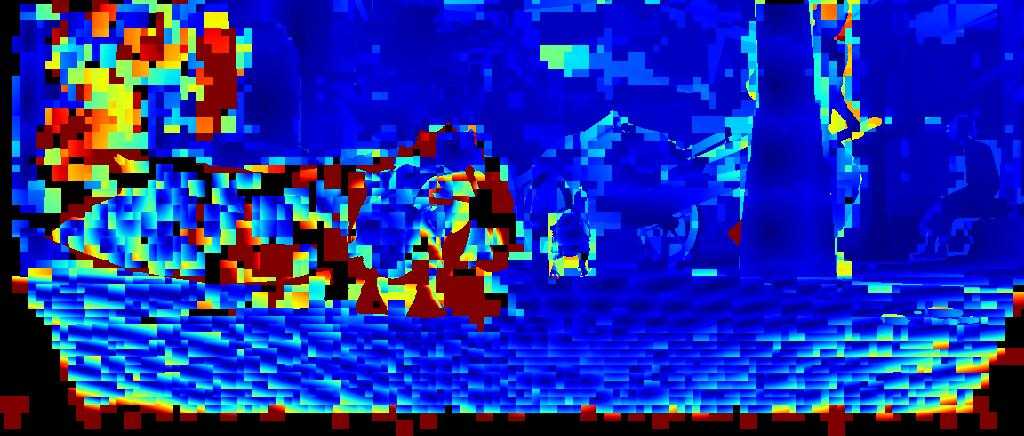}
  & 
  \imgs{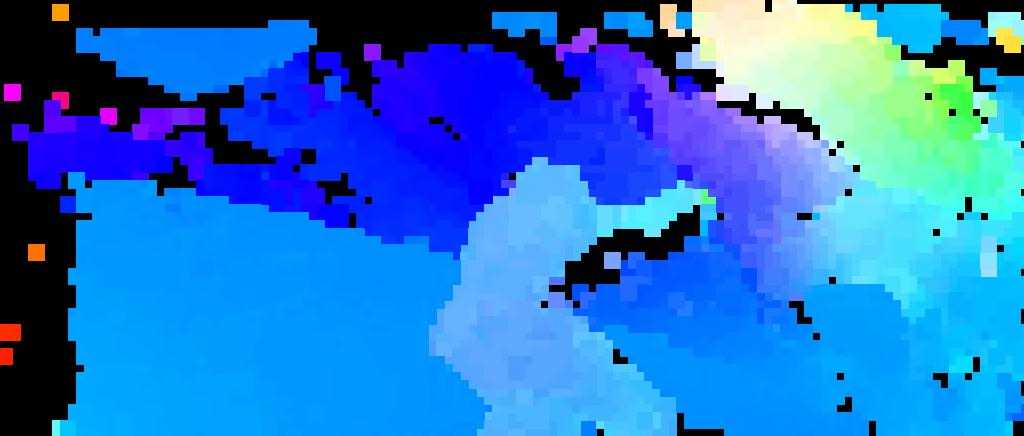} &
  \imgs{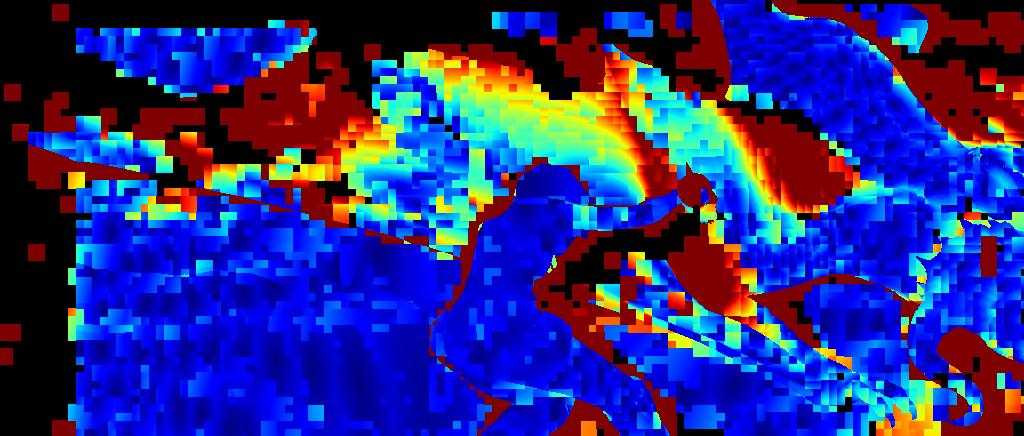} &
  \tabularnewline
\end{tabular}
}
\caption{Comparison of different matching methods on three challenging pairs with non-rigid deformations from MPI-Sintel. 
Each pair of columns shows motion maps (left column) and 
the corresponding error maps (right column). The top row presents the ground-truth (GT) as well as one image. 
For non-dense methods, pixel displacements have been inferred from matching patches. 
Areas without correspondences are in black. 
\label{fig:ex-sintel}
}
\end{figure*}

\begin{figure*}
\resizebox{\linewidth}{!}{
\begin{tabular}{ccc||ccc}
  & 
  \tbtxt{Correspondence field} & \tbtxt{Image/Error map} & 
  \tbtxt{Correspondence field} & \tbtxt{Image/Error map} & \tabularnewline
\tbtxtrot{~~~~~~GT}  & 
  \imgk{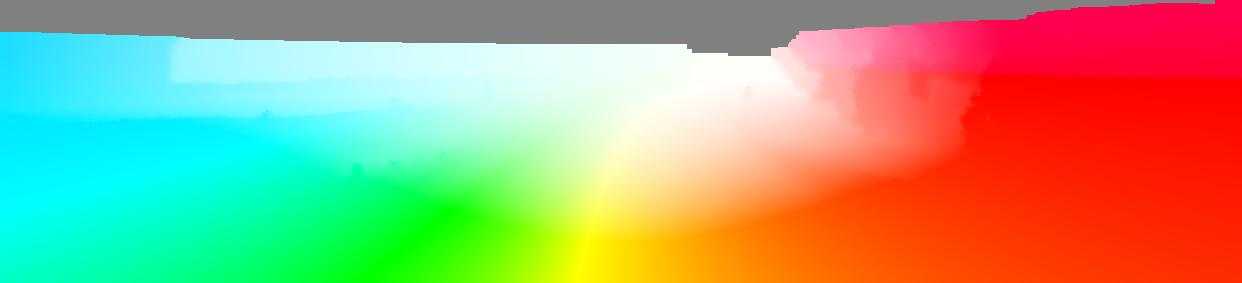} &
  \imgk{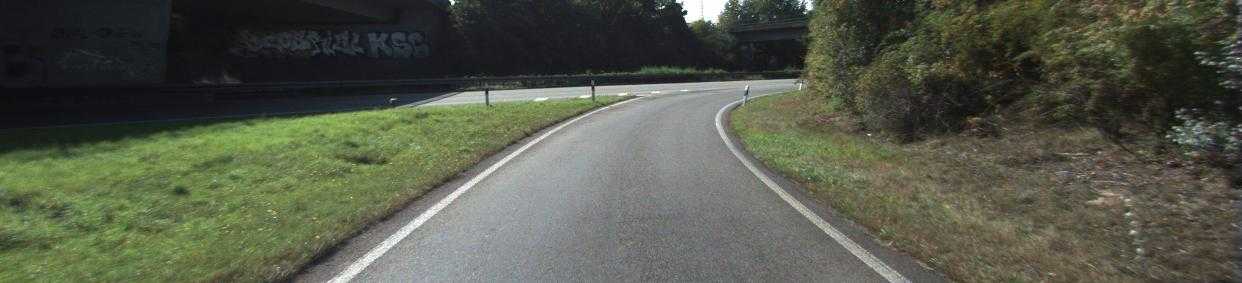}
  & 
  \imgk{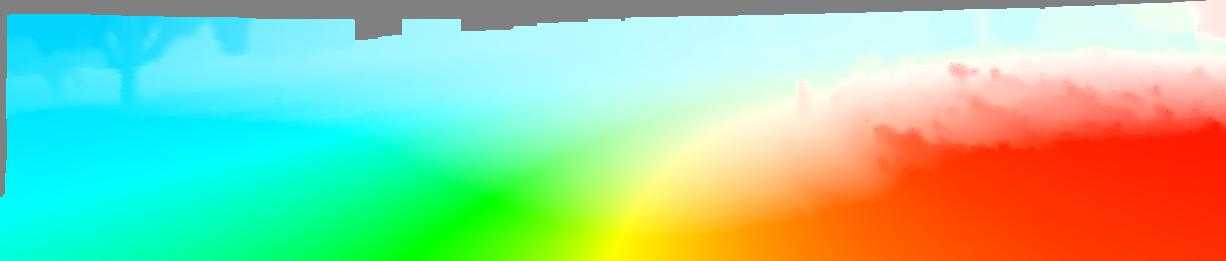} &
  \imgk{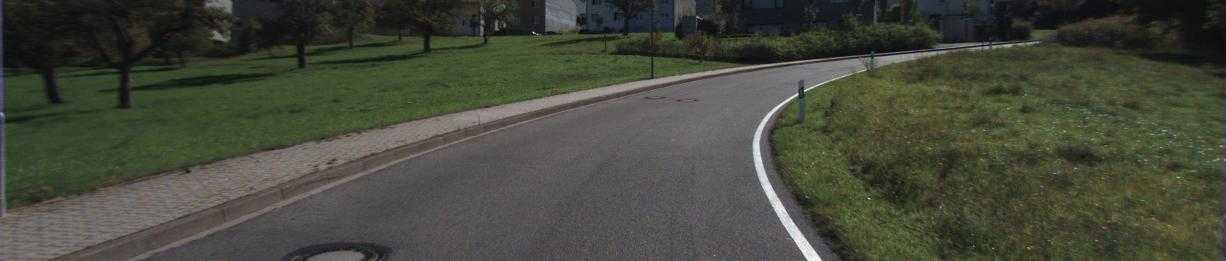} & 
  \tabularnewline
\tbtxtrot{~SIFT-NN} &
  \imgk{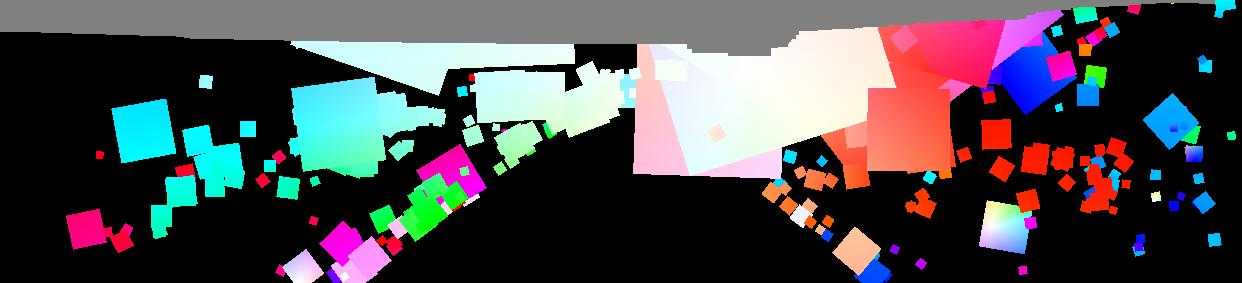} &
  \imgk{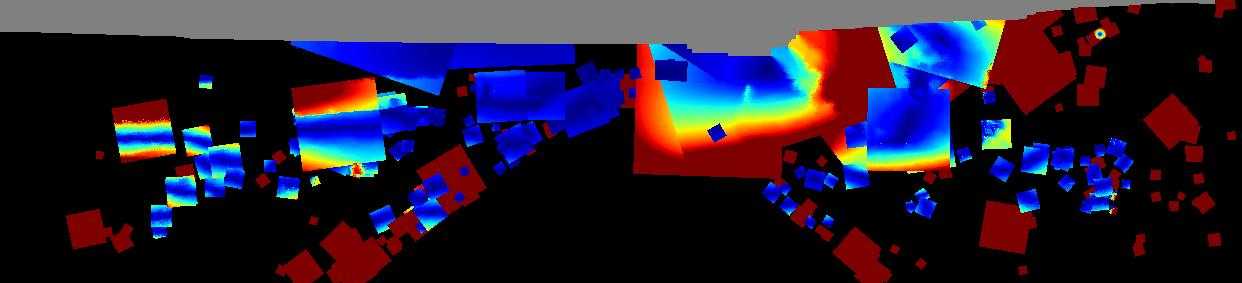}
  & 
  \imgk{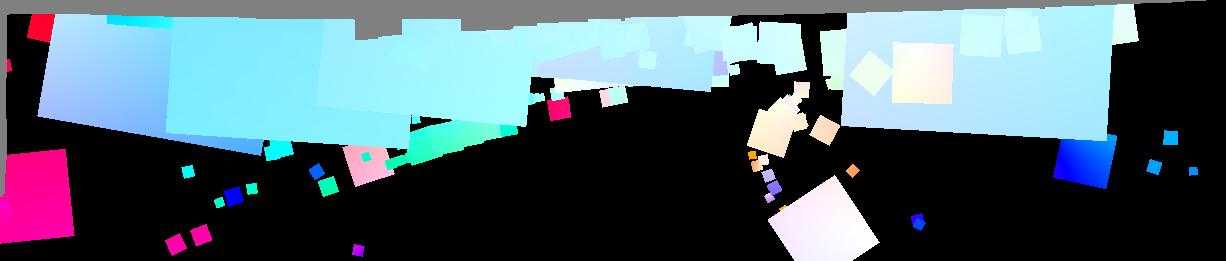} &
  \imgk{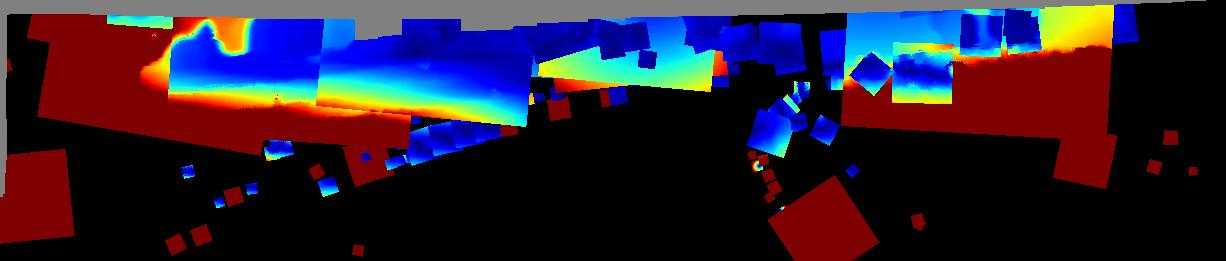} &
  \tabularnewline
\tbtxtrot{HOG-NN} &
  \imgk{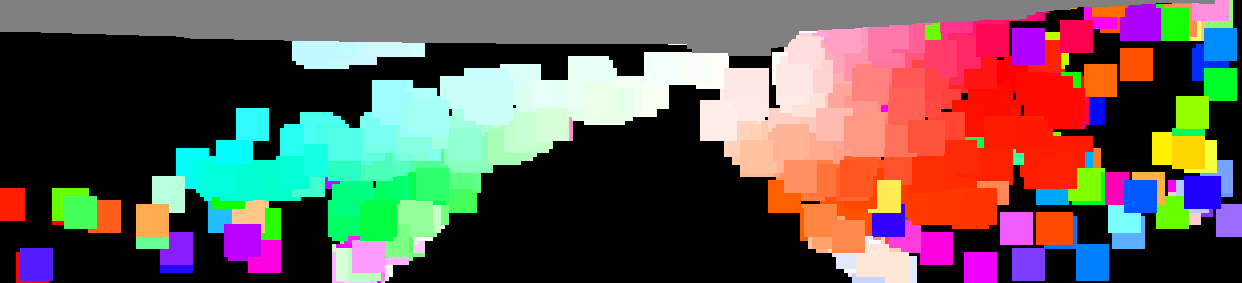} &
  \imgk{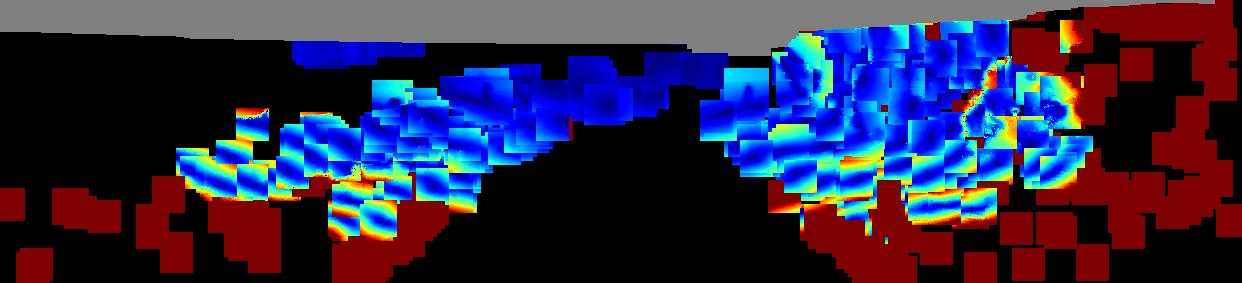}
  & 
  \imgk{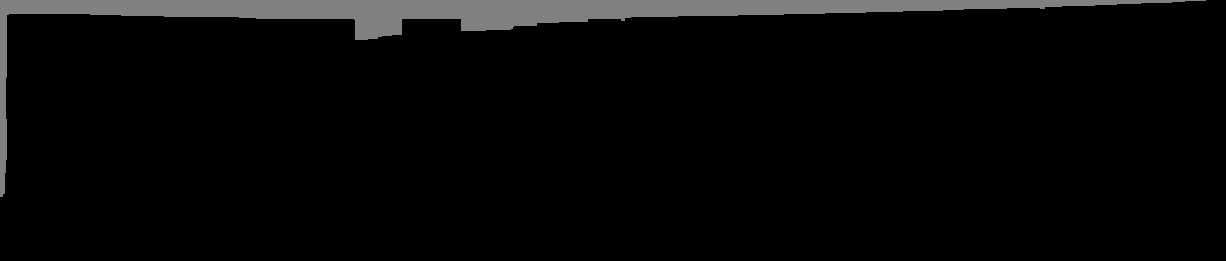} &
  \imgk{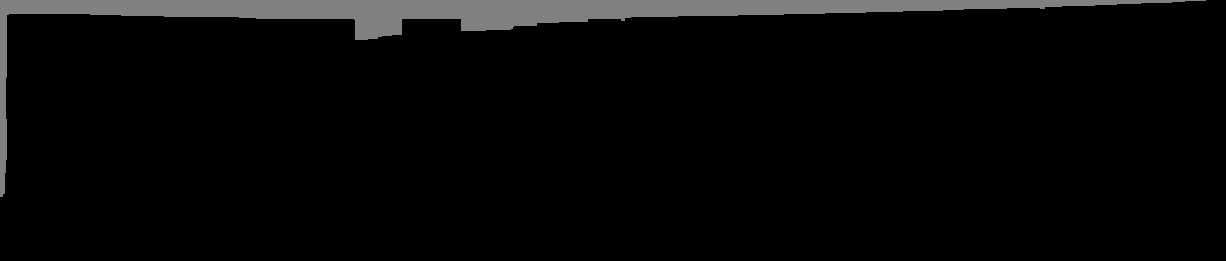} &
  \tabularnewline
\tbtxtrot{~~~KPM} &
  \imgk{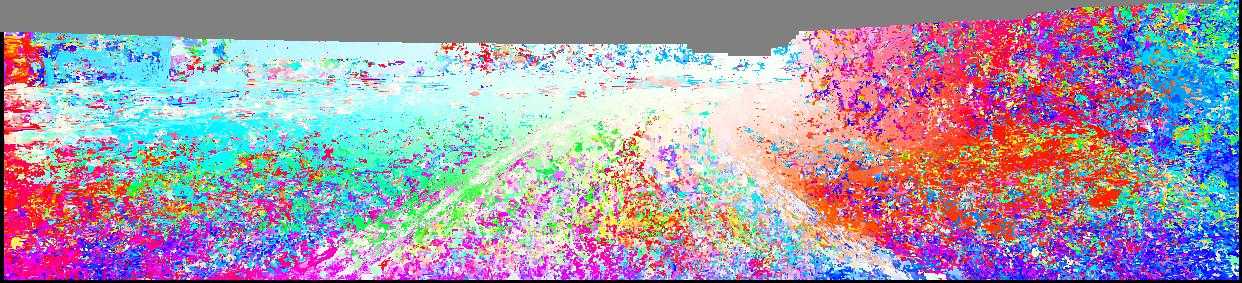} &
  \imgk{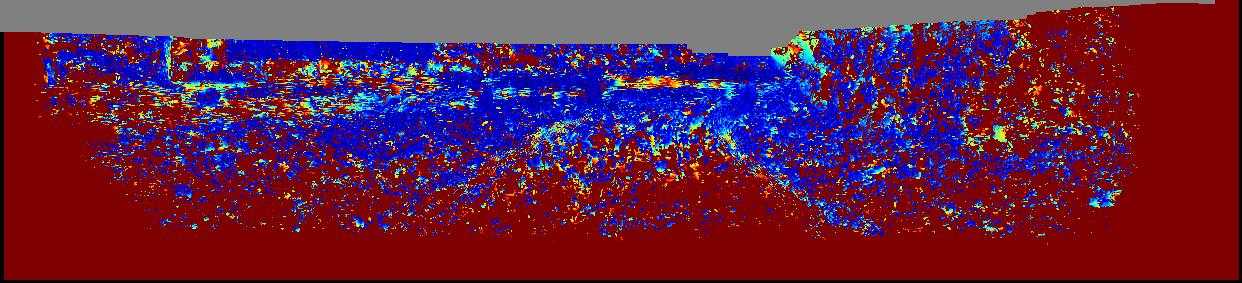}
  & 
  \imgk{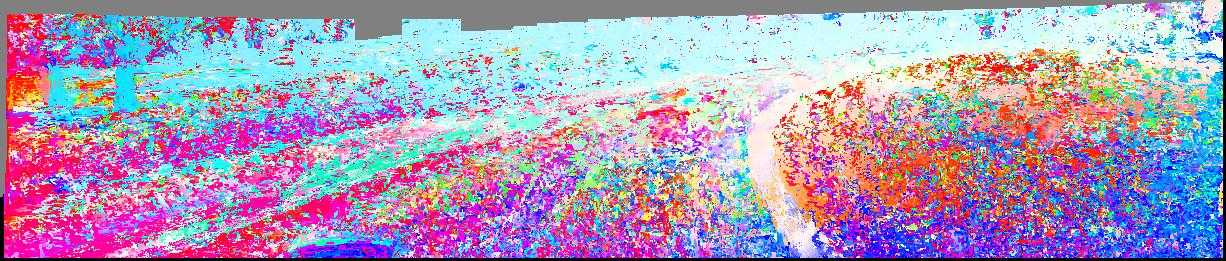} &
  \imgk{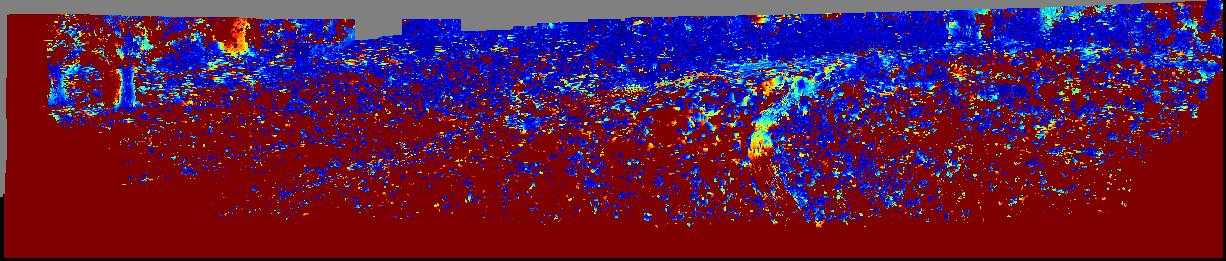} &
  \multirow{7}{*}{\includegraphics[height=75mm]{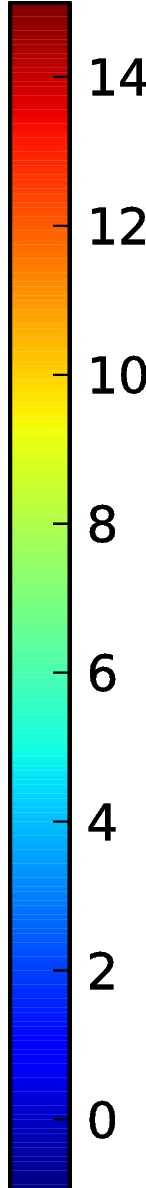}}
  \tabularnewline
\tbtxtrot{~~~GPM} &
  \imgk{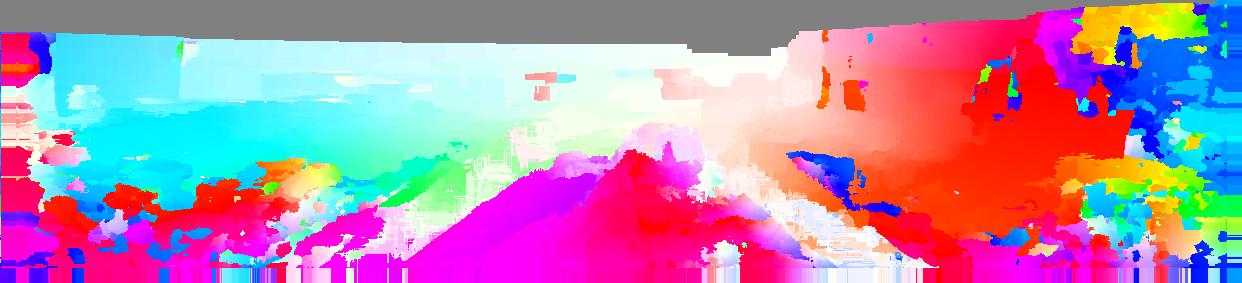} &
  \imgk{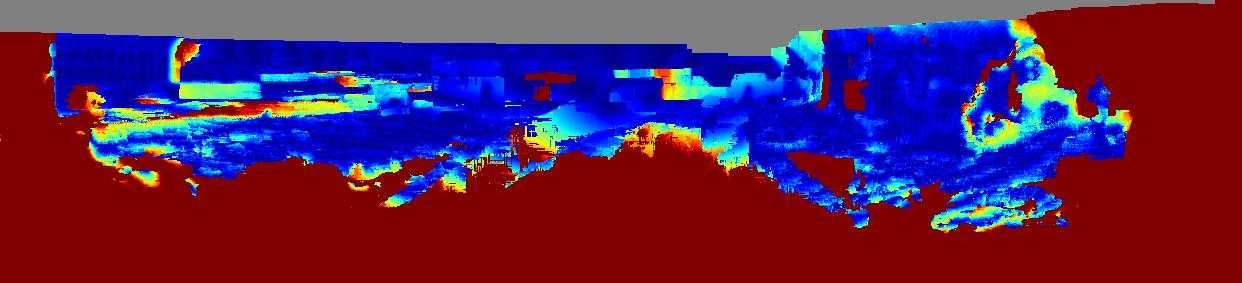}
  & 
  \imgk{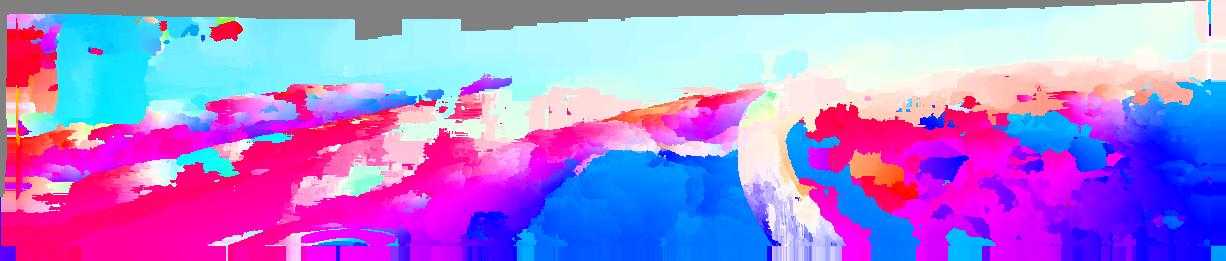} &
  \imgk{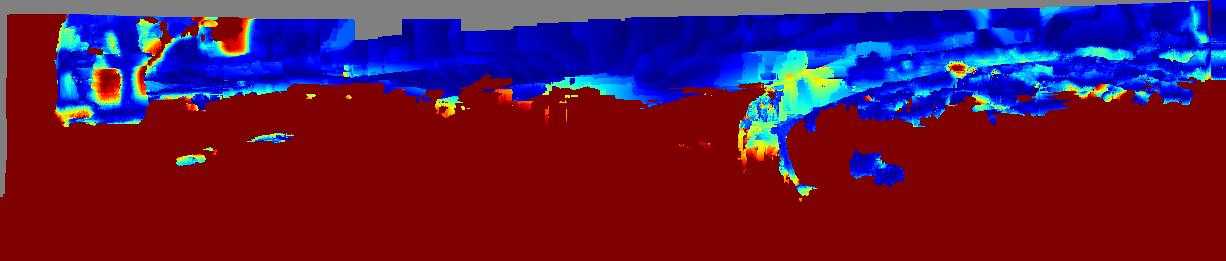} &
  \tabularnewline
\tbtxtrot{SIFT-flow} &
  \imgk{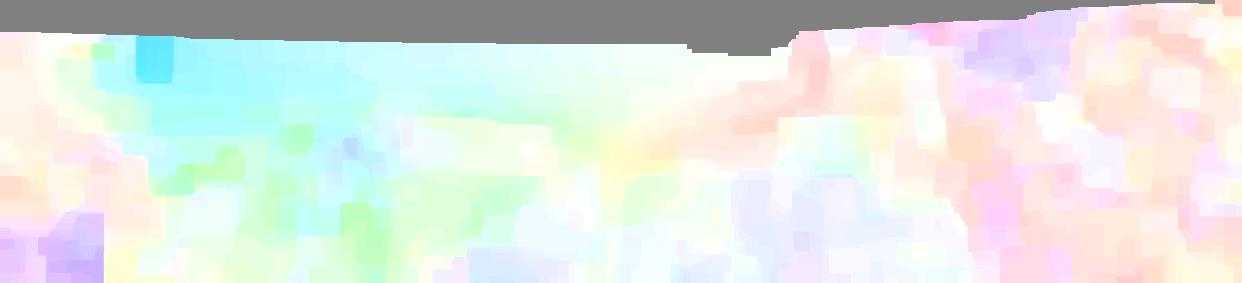} &
  \imgk{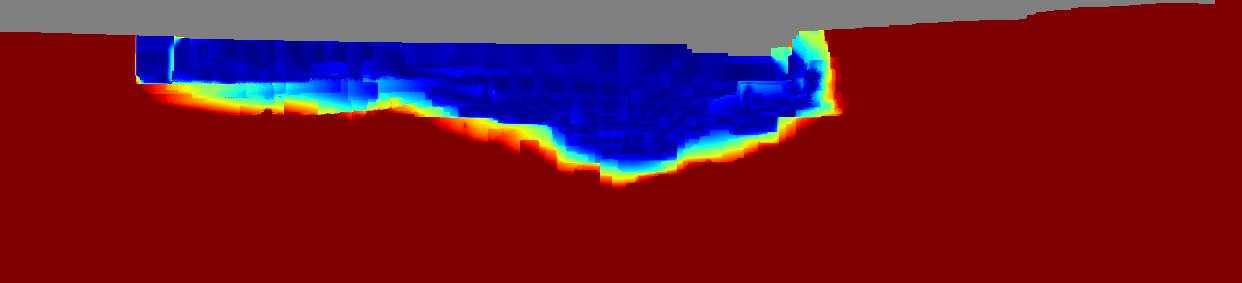}
  & 
  \imgk{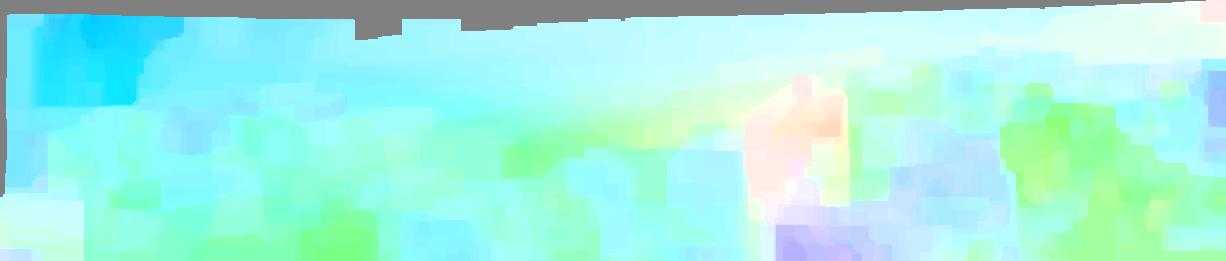} &
  \imgk{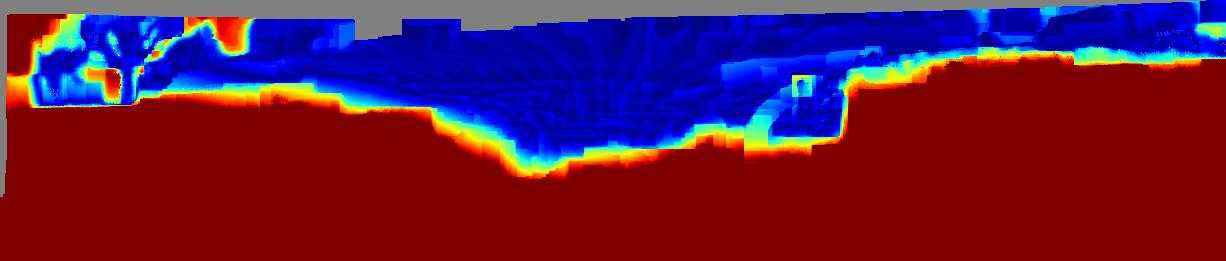} &
  \tabularnewline
\tbtxtrot{~~~~SLS} &
  \imgk{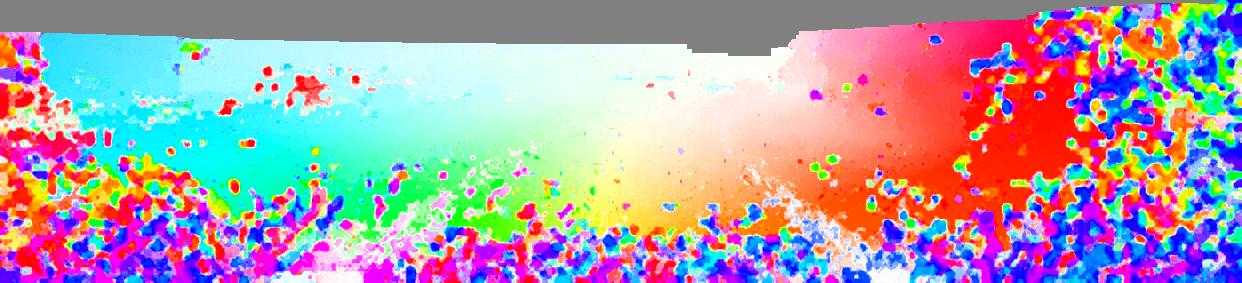} &
  \imgk{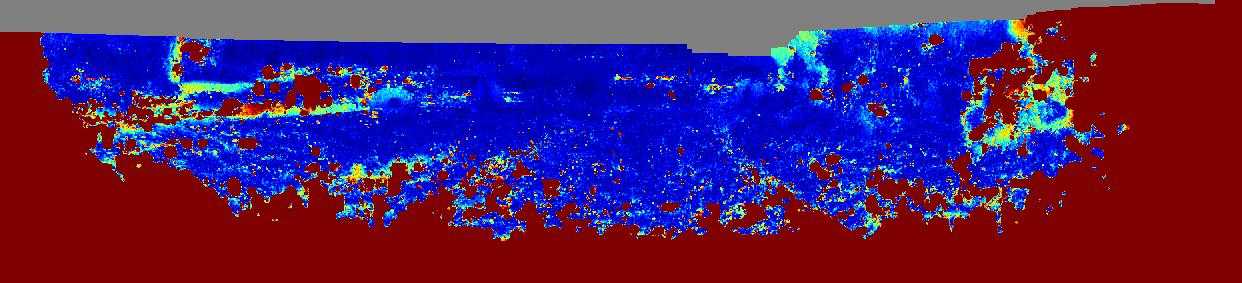}
  & 
  \imgk{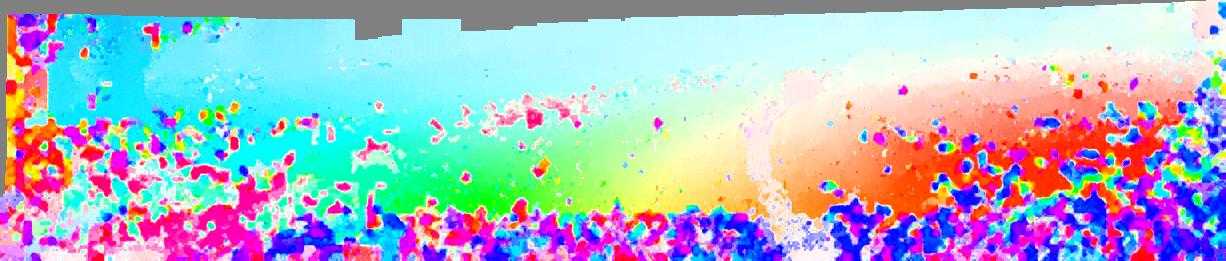} &
  \imgk{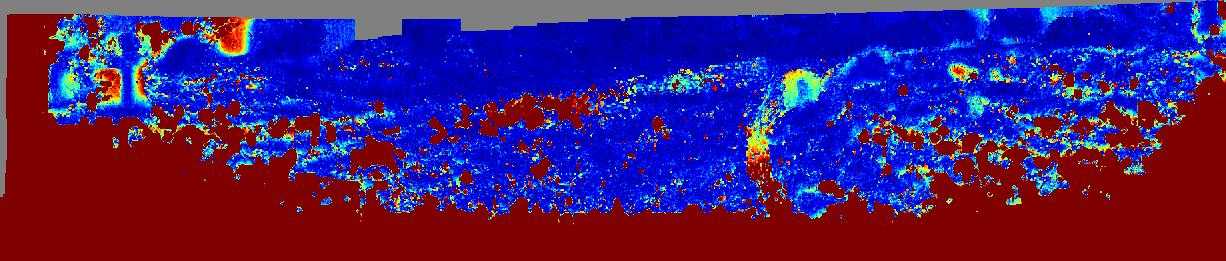} &
  \tabularnewline
\tbtxtrot{~DaisyFF} &
  \imgk{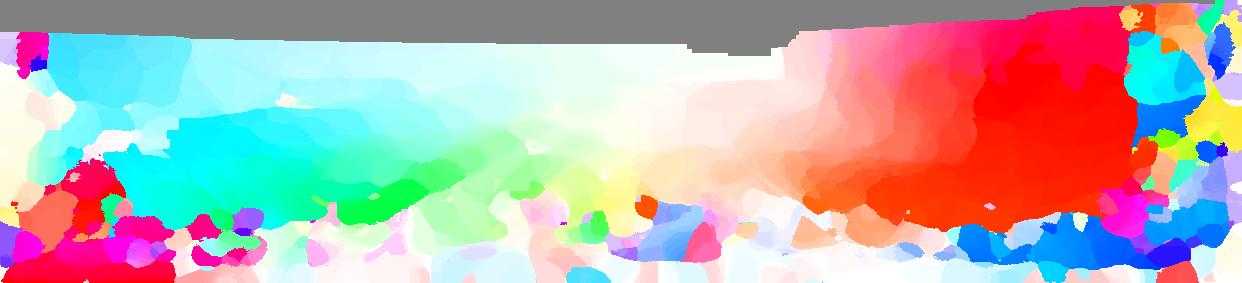} &
  \imgk{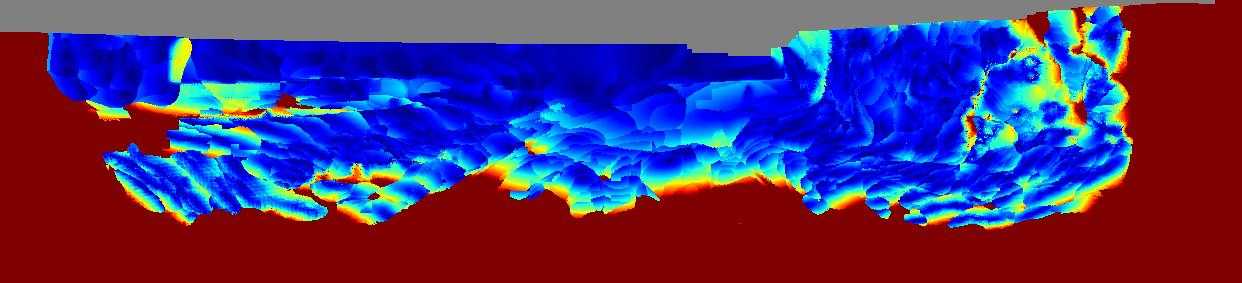}
  & 
  \imgk{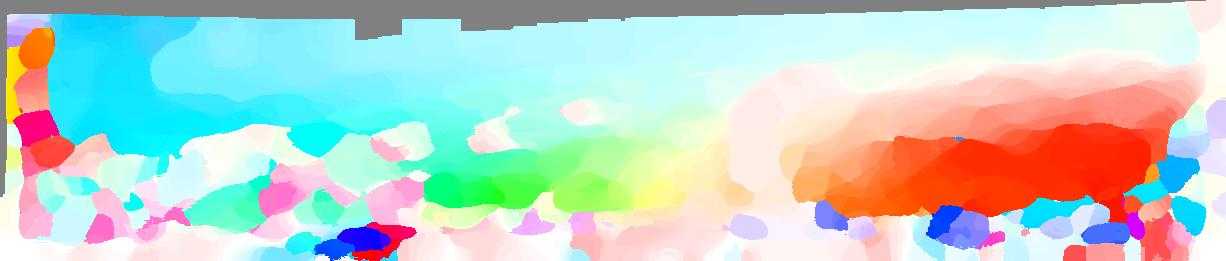} &
  \imgk{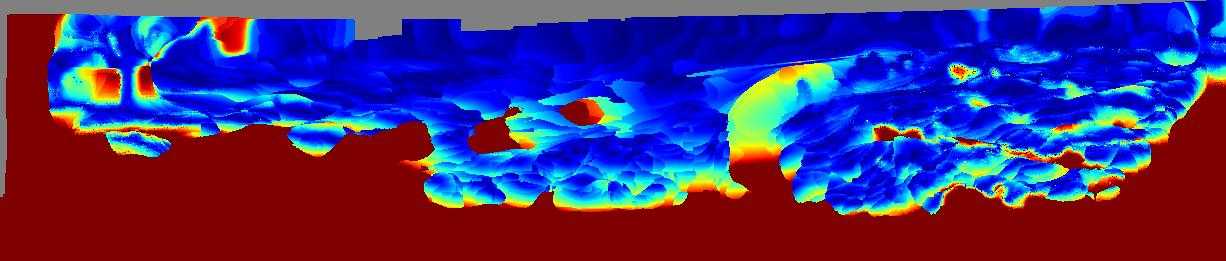} &
  \tabularnewline
\tbtxtrot{~~~DSP} &
  \imgk{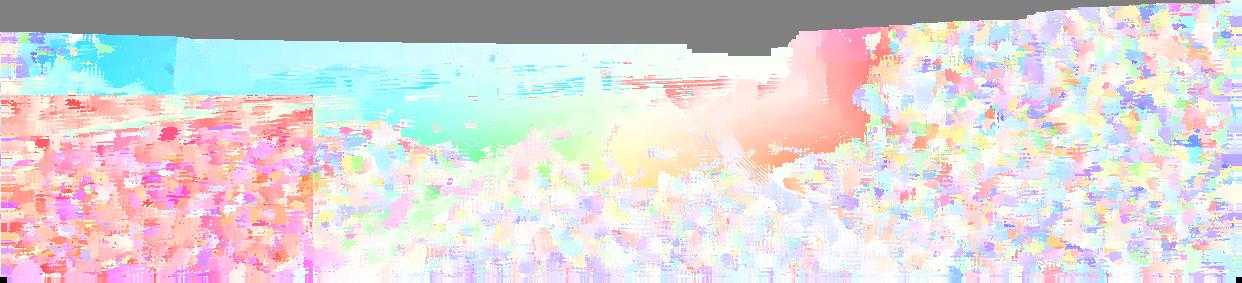} &
  \imgk{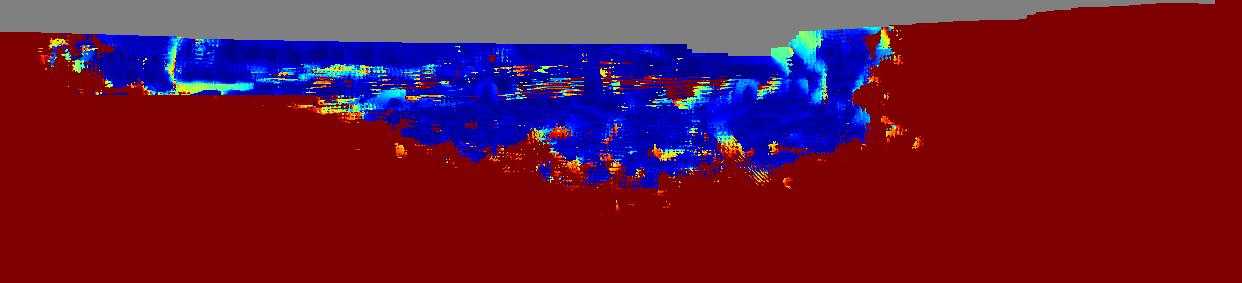}
  & 
  \imgk{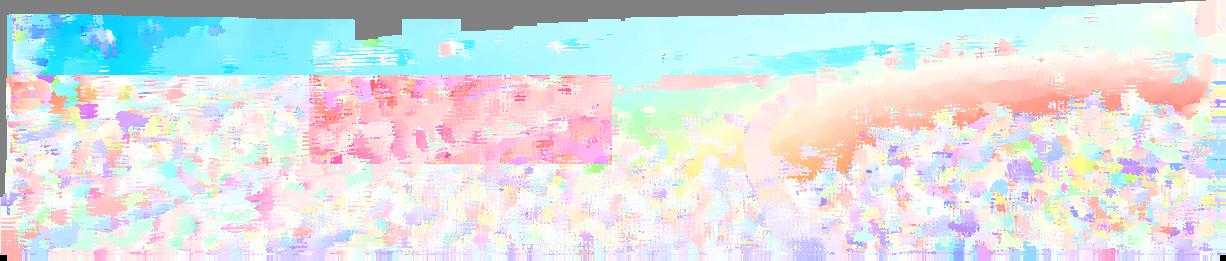} &
  \imgk{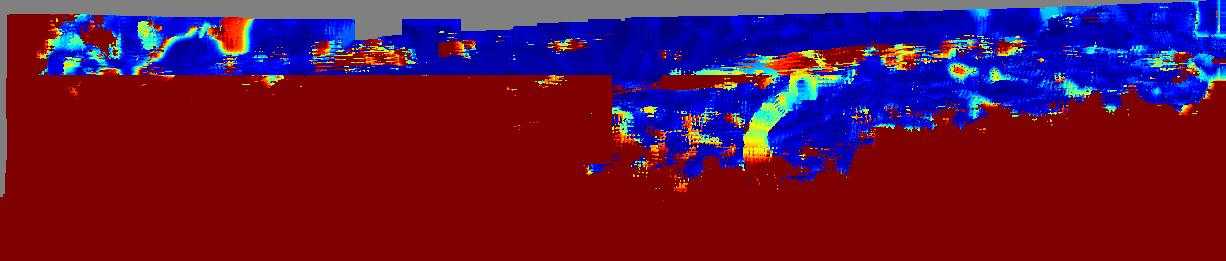} &
  \tabularnewline
\tbtxtrot{~~~~~\textbf{DM}} &
  \imgk{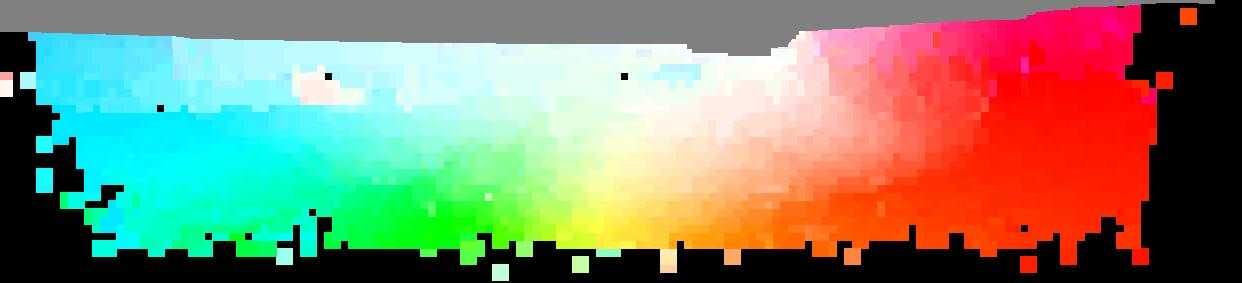} &
  \imgk{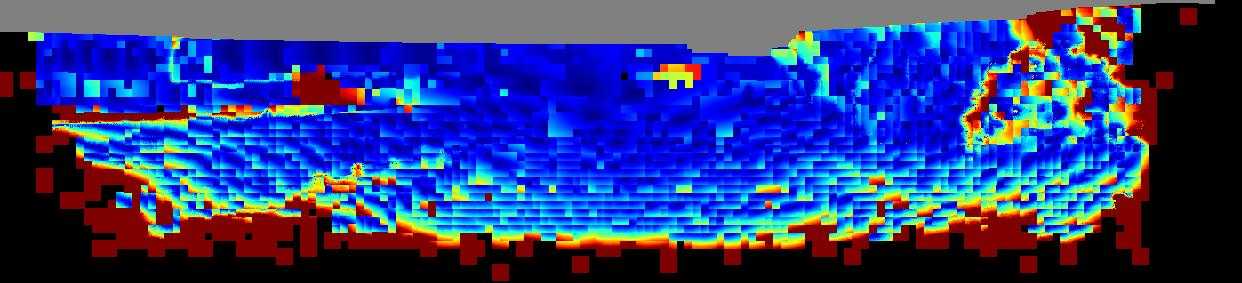}
  & 
  \imgk{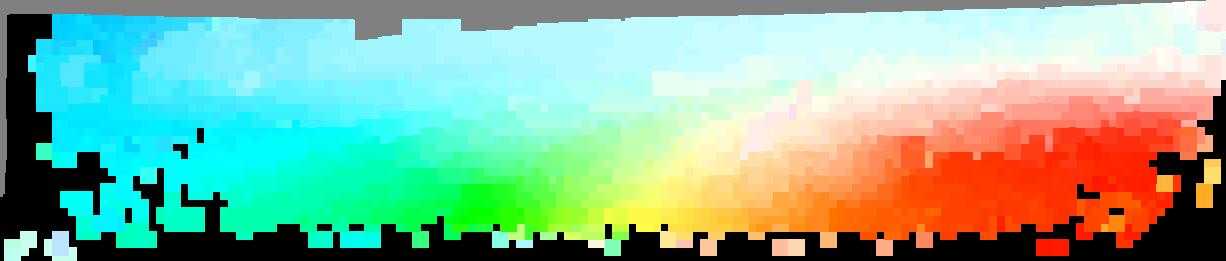} &
  \imgk{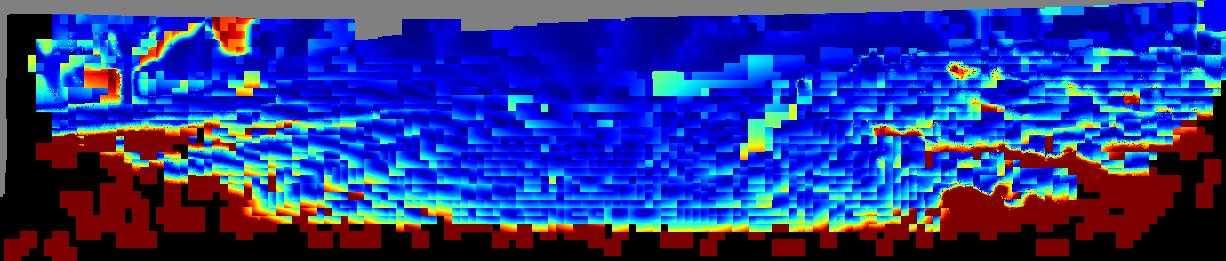} &
  \tabularnewline
\end{tabular}
}
\caption{
Comparison of different matching methods on three challenging pairs from Kitti.
Each pair of columns shows motion maps (left column) and 
the corresponding error maps (right column). The top row presents the ground-truth (GT) as well as one image. 
For non-dense methods, pixel displacements have been inferred from matching patches. 
Areas without correspondences are in black. 
To improve visualization, the sparse Kitti ground-truth is made dense using bilateral filtering.
\label{fig:ex-kitti}
}
\end{figure*}

\subsection{Optical Flow Experiments}
\label{sub:flowxp}

We now present experimental results for the optical flow estimation. 
Optical flow is predicted using the variational framework presented in Section~\ref{sec:flow}
that takes as input a set of matches.
In the following, we evaluate the impact of DeepMatching against 
other matching methods, and compare to the state of the art. 

\subsubsection{Optimization of the parameters}
\label{subsub:flowparams}

We optimize the parameters of DeepFlow on a subset of the MPI-Sintel training set (20\%), called ``small'' set,
\red{and report results on the remaining image pairs (80\%, called ``validation set'') and on the training sets of Kitti and Middlebury.
Ground-truth optical flows for the three test sets are not publicly available, in order to prevent parameter tuning on the test set. }

We first optimize the different flow parameters ($\beta$, $\gamma$, $\delta$, $\sigma$ and $b$) 
by employing a gradient descent strategy with multiple initializations
followed by a local grid search.  
For the data term, we find an optimum at $\delta=0$, which is equivalent to removing the
color constancy assumption. This can be explained by the fact that
the ``final'' version contains atmospheric effects, reflections,
blurs, etc. The remaining parameters are optimal at 
$\beta = 300$, $\gamma = 0.8$, $\sigma = 0.5$, $b = 0.6$.
These parameters are used in the remaining of the experiments 
\red{for DeepFlow, \ie using matches obtained with DeepMatching, except when reporting results on Kitti and 
Middlebury test sets in Section~\ref{sub:dfsota}. In this case the parameters
are optimized on their respective training set}.

\subsubsection{Impact of the matches on the flow}
\label{sub:impactflow}

We examine the impact of different matching methods on the flow, \ie,
different matches are used in DeepFlow, see Section~\ref{sec:flow}. 
For all matching approaches evaluated in the previous section, we use
their output as matching term in \eq\refp{eqn:variational}. 
Because these approaches may output matches with statistics different from DM, 
we separately optimize the flow parameters for each matching approach on the small 
training set of MPI-Sintel\footnote{\red{Note that this systematically
    improves the endpoint error
compared to using the raw dense correspondence fields as flow. }}.

\begin{table}
\centering
\begin{tabular}{|l|c|c||ccc|}
\hline
Method & R & D  & MPI-Sintel & Kitti & Middlebury \\
\hline 
\multicolumn{3}{|l||}{No Match}  & 5.863  & 8.791 & 0.274 \\
\hline
\multicolumn{3}{|l||}{SIFT-NN}  & 5.733  & 7.753  & 0.280  \\
\multicolumn{3}{|l||}{HOG-NN}   & 5.458  & 8.071  & \textbf{0.273}  \\
\multicolumn{3}{|l||}{KPM}      & 5.560  & 15.289  & 0.275  \\
\multicolumn{3}{|l||}{GPM}      & 5.561  & 17.491 & 0.286  \\
\multicolumn{3}{|l||}{\red{SIFT-flow}} & 5.243 & 12.778 & 0.283  \\
\multicolumn{3}{|l||}{\red{SLS}}       & 5.307 & 10.366 & 0.288  \\
\multicolumn{3}{|l||}{\red{DaisyFF}}   & 5.145 & 10.334 & 0.289  \\
\multicolumn{3}{|l||}{\red{DSP}}       & 5.493 & 15.728 & 0.283  \\
\hline
DM & 1/2 & 1024        & 4.350  & 7.899  & 0.320  \\
DM & 1/2 & $\infty$    & \textbf{4.098}  & \textbf{4.407}  & 0.328  \\
\hline
\end{tabular}
\caption{Comparison of average endpoint error on different datasets when changing the input matches in the flow computation.}
\label{tab:matchfloweval} 
\end{table}

Table~\ref{tab:matchfloweval} shows the endpoint error, averaged over all pixels.
Clearly, a sufficiently dense and accurate matching like DM allows to considerably improve the flow estimation
on datasets with large displacements (MPI-Sintel, Kitti). 
In contrast, none of \red{the} methods presented have a tangible 
effect on the Middlebury dataset, where the displacements are small. 

\begin{figure*}
\centering
\begin{tabular}{cccc}
\begin{turn}{90}{~~~~~~Images}\end{turn}  & \includegraphics[width=0.3\linewidth]{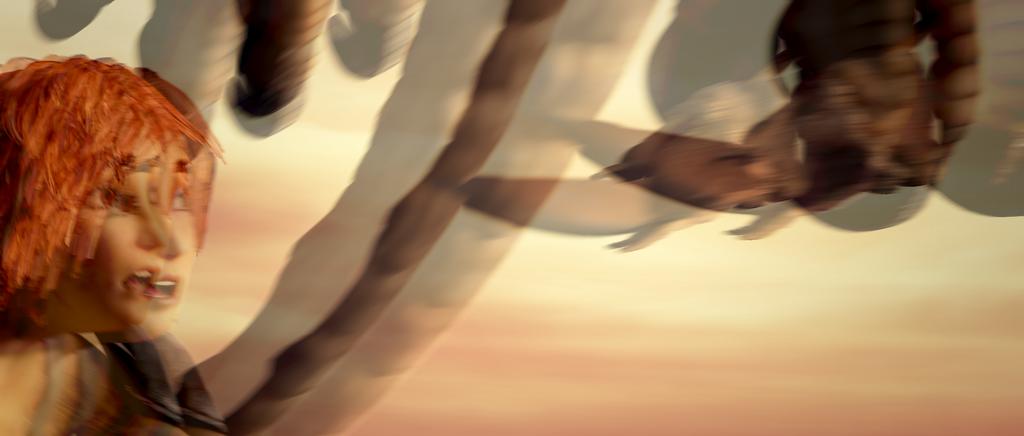}  & \includegraphics[width=0.3\linewidth]{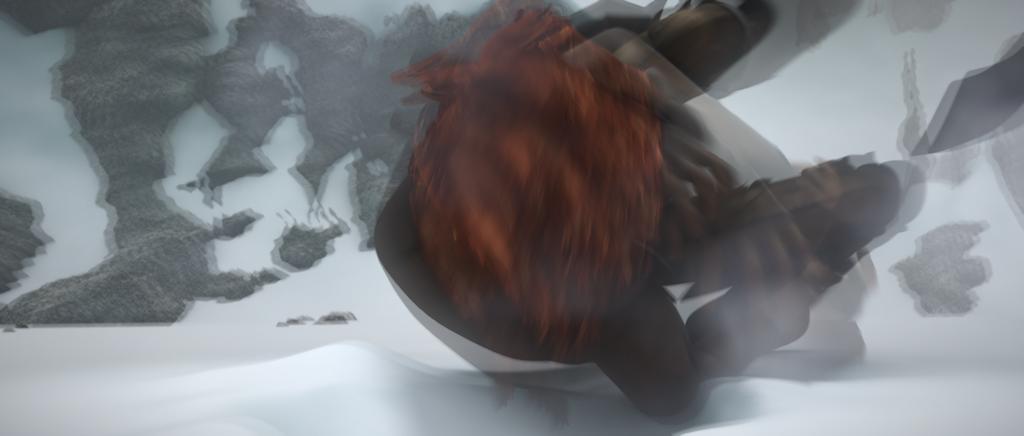}  & \includegraphics[width=0.3\linewidth]{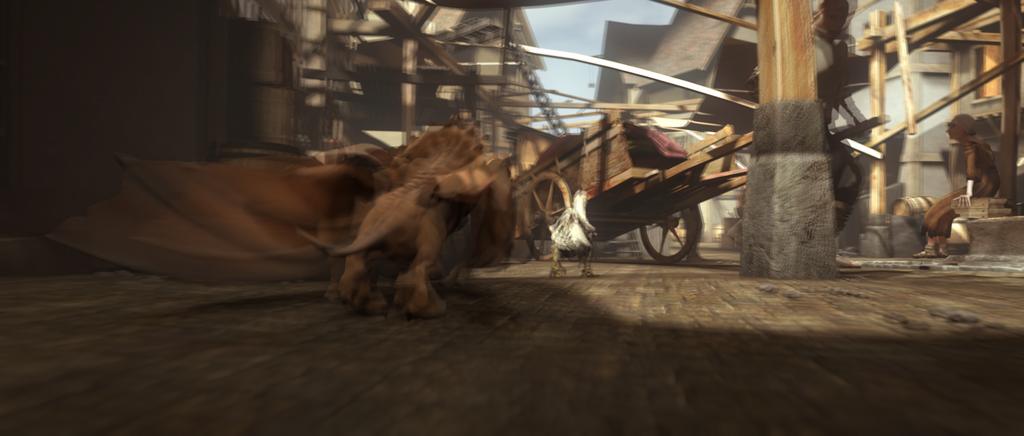} \\
\begin{turn}{90}{~~Ground-Truth}\end{turn}  & \includegraphics[width=0.3\linewidth]{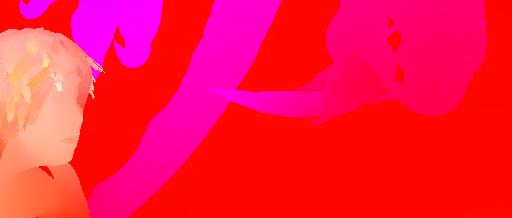}  & \includegraphics[width=0.3\linewidth]{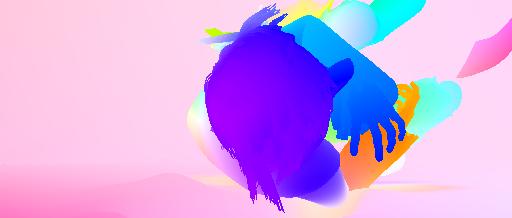}  & \includegraphics[width=0.3\linewidth]{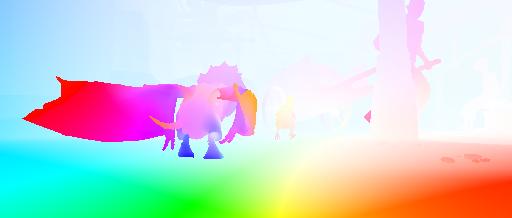} \\
\begin{turn}{90}{\textbf{DeepMatching}}\end{turn}  & \includegraphics[width=0.3\linewidth]{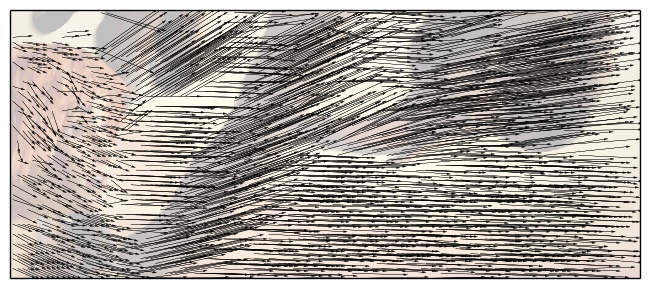}  & \includegraphics[width=0.3\linewidth]{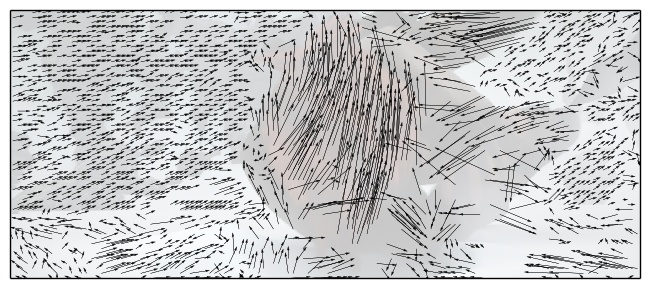}  & \includegraphics[width=0.3\linewidth]{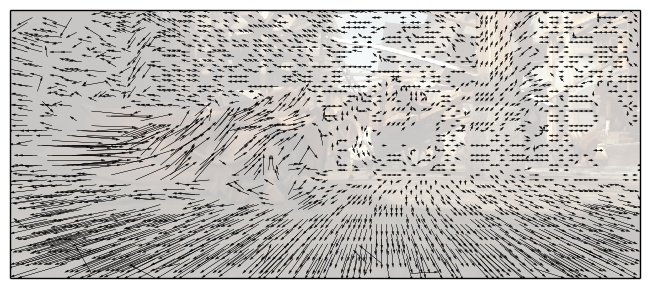} \\
\begin{turn}{90}{~~\textbf{DeepFlow}}\end{turn}  & \includegraphics[width=0.3\linewidth]{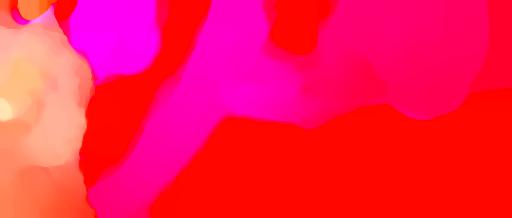}  & \includegraphics[width=0.3\linewidth]{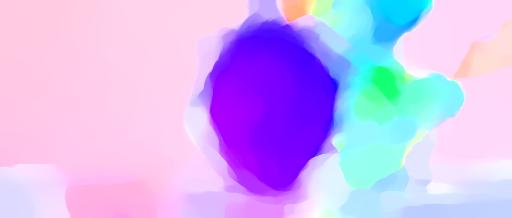}  & \includegraphics[width=0.3\linewidth]{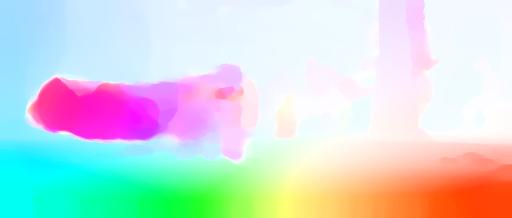} \\
\begin{turn}{90}{~~~MDP-Flow2}\end{turn}  & \includegraphics[width=0.3\linewidth]{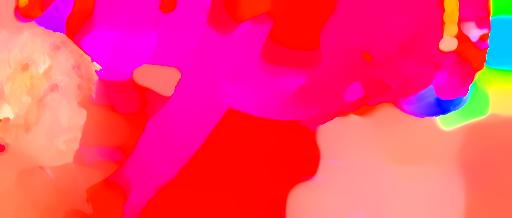}  & \includegraphics[width=0.3\linewidth]{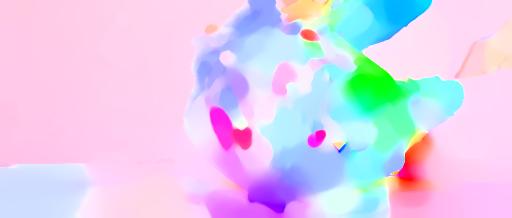}  & \includegraphics[width=0.3\linewidth]{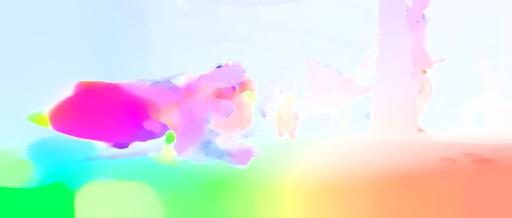} \\
\begin{turn}{90}{~~~~~~~~LDOF}\end{turn}  & \includegraphics[width=0.3\linewidth]{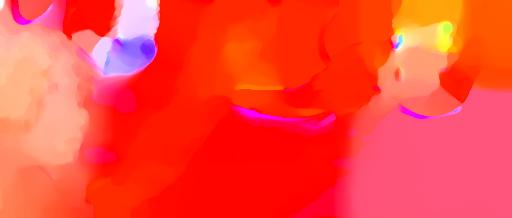}  & \includegraphics[width=0.3\linewidth]{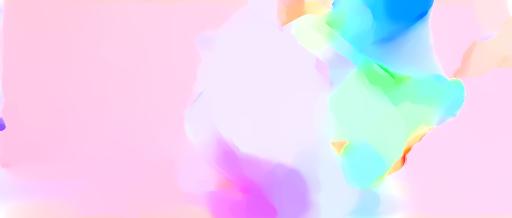}  & \includegraphics[width=0.3\linewidth]{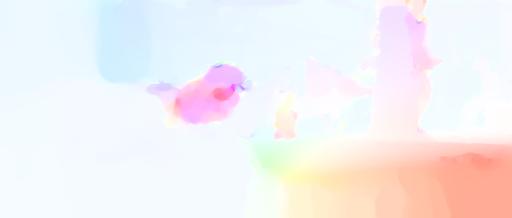}
\end{tabular}
\caption{Each column shows from top to bottom: two consecutive images, the ground-truth optical flow, the DeepMatching, 
         our flow prediction (\emph{DeepFlow}), and two state-of-the-art methods, LDOF~\citep{Bro11a} and MDP-Flow2~\citep{mdpof}.}
 \label{fig:flows}
\end{figure*}

The relatively small gains achieved by SIFT-NN and HOG-NN on MPI-Sintel and Kitti are due to the fact that
a lot of regions with large displacements are not covered by any matches, such as the 
sky or the blurred character in the first and second column of Figure~\ref{fig:flows}. Hence,
SIFT-NN and HOG-NN have only a limited impact on the variational
approach. 
On the other hand, the gains are also small (or even negative) \red{for the dense methods
despite the fact that they output significantly more correspondences}.
We observe for these methods that the weight $\beta$ of the matching term tends to be small after optimizing the parameters, 
thus indicating that the matches are found unreliable and noisy during training.
The cause is clearly visible in Figure~\ref{fig:ex-kitti}, where large portions
containing repetitive textures (\eg road, trees) are incorrectly matched.
The poor quality of these matches even leads to a significant drop 
in performance on the Kitti dataset. 

In contrast, DeepMatching generates accurate matches well covering the image that enable
to boost the optical flow accuracy in case of large displacements. 
Namely, we observe a relative improvement of 30\% on MPI-Sintel and of
50\% on Kitti. 
\red{It is interesting to observe that DM is able to effectively prune 
false matches arising in occluded areas (black areas in Figures~\ref{fig:ex-sintel} and \ref{fig:ex-kitti}). 
This is due to the reciprocal verification filtering incorporated in DM (\eq\refp{eqn:reciprocal}). }
When using the approximation with 1024 prototypes, however, a significant drop is observed 
on the Kitti dataset, while the performance remains good on MPI-Sintel.
This indicates that approximating DeepMatching can result in a significant loss
of robustness when matching repetitive textures, that are more frequent in Kitti than in MPI-Sintel.

\subsubsection{Comparison to the state of the art}
\label{sub:dfsota}

\red{In this section, we compare DeepFlow to the state of the art 
on the test sets of MPI-Sintel, Kitti and Middlebury datasets. 
For theses datasets, the results are submitted to a dedicated server 
which performs the evaluation. Prior to submitting our results for Kitti and Middlebury test sets, 
we have optimized the parameters
on the respective training set.}

\paragraph{Results on MPI-Sintel.}

Table~\ref{tab:sintel} compares our method to state-of-the-art algorithms
on the MPI-Sintel test set.
A comparison with the preliminary version of DeepFlow~\citep{DeepFlow}, 
referred to as DeepFlow*, is also provided. In this early version, we
used a constant smoothness weight instead of a local one here (see
Section~\ref{sec:flow}) and used DM* as input matches. 
We can see that DeepFlow is among the best performing
methods on MPI-Sintel, particularly for large displacements. 
This is due to the use of a reliable matching term in the variational approach,
and this property is shared by all top performing approaches, \eg \citep{epicflow, Leordeanu2013}.
Furthermore, it is interesting to note that among the top performers on
MPI-Sintel, 3 methods out of 6 actually employ DeepMatching. In particular, 
the top-3 method EpicFlow~\citep{epicflow} relies on the
output of DeepMatching to produce a piece-wise affine flow,
and SparseFlowFused \citep{sparseflowfused} combines matches obtained with DeepMatching and another algorithm.

We refer to the webpage of the MPI-Sintel dataset for complete results including the ``clean'' version.

\begin{table}
\resizebox{\linewidth}{!}{
\centering 
\begin{tabular}{|l||c|c|ccc||r|}
\hline
 Method & EPE  & EPE-occ & s0-10  & s10-40  & s40+ & Time \\
\hline 
FlowFields~\citep{flowfields} & 5.810 & 31.799 & 1.157 &	3.739	& 33.890 & 23s \\
DiscreteFlow~\citep{discreteflow} & 5.810	& 31.799 & 1.157 & 3.739 & 33.890 & 180s \\
EpicFlow~\citep{epicflow}  & 6.285  & 32.564 & 1.135  & 3.727 & 38.021 & 16.4s \\
TF+OFM~\citep{kennedyoptical}    & 6.727 & 33.929 & 1.512 & 3.765 & 39.761 & $\sim$400s \\
\textbf{DeepFlow}  & 6.928  & 38.166 & 1.182  & 3.859  & 42.854  & 25s\\
SparseFlowFused~\cite{sparseflowfused} & 7.189	& 3.286	& 	1.275	& 3.963	& 44.319 & 20 \\
DeepFlow*~\citep{DeepFlow}  & 7.212  & 38.781 & 1.284  & 4.107  & 44.118  & 19s\\
S2D-Matching~\citep{Leordeanu2013}  & 7.872 & 40.093 & 1.172  & 4.695  & 48.782 & $\sim$2000s \\
LocalLayering~\citep{LocalLayers} & 8.043 & 40.879 & 1.186 & 4.990 & 49.426 & \\
Classic+NL-P~\citep{Sun2014}  & 8.291  & 40.925 & 1.208  & 5.090  & 51.162  & $\sim$800s \\
MDP-Flow2~\citep{mdpof}  & 8.445 & 43.430 & 1.420  & 5.449  & 50.507 & 709s\\
NLTGV-SC~\citep{ranftl2014non} & 8.746 & 42.242 & 1.587 & 4.780 & 53.860 & \\
LDOF~\citep{Bro11a}  & 9.116  & 42.344 & 1.485  & 4.839  & 57.296 & 30s\\
\hline
\end{tabular}}
\caption{Results on MPI-Sintel test set (final version). 
EPE-occ is the EPE on occluded areas. s0-10 is the EPE
for pixels with motions between 0 and 10 px and similarly for s10-40 and s40+. 
DeepFlow* denotes the preliminary version of DeepFlow published in \cite{DeepFlow}.
}
\label{tab:sintel} 
\end{table}

\paragraph{Timings.} 
As mentioned before, DeepMatching at half the resolution takes 15 seconds to compute on CPU and 0.2 second on GPU.
The variational part requires 10 additional seconds on CPU. 
Note that by implementing it on GPU, we could obtain a significant speed-up as well.
DeepFlow consequently takes 25 seconds in total on 
a single CPU core @ 3.6 GHz or 10.2s with GPU+CPU. This is in the same order of magnitude as
the fastest among the best competitors, EpicFlow~\citep{epicflow}.

\paragraph{Results on Kitti.}

Table~\ref{tab:kitti} summarizes the main results on the Kitti benchmark
(see official website for complete results), when optimizing the parameters
on the Kitti training set. EPE-Noc is the EPE computed only in
non-occluded areas. 
``Out 3'' corresponds to the proportion of incorrect pixel correspondences for an error
threshold of 3 pixels, \ie it corresponds to $1-\mathrm{accuracy@3}$, and likewise for ``Out-Noc 3'' 
for non-occluded areas. 
In terms of EPE-noc, DeepFlow is on par with the best approaches, 
but performs somewhat worse in the occluded areas. This is 
due to a specificity of the Kitti dataset, in which motion is mostly homographic 
(especially on the image borders, where most surfaces like roads and walls are planar). 
In such cases, flow is better predicted using an affine motion prior, 
which locally well approximates homographies (a constant motion prior is used in DeepFlow).
As a matter of facts, all top performing methods in terms of total EPE output
piece-wise affine optical flow, either due to affine regularizers 
(BTF-ILLUM \citep{demetz2014learning}, NLTGB-SC \citep{ranftl2014non}, TGV2ADCSIFT \citep{Braux-Zin_2013_ICCV})
or due to local affine estimators (EpicFlow \citep{epicflow}).

Note that the learned parameters
on Kitti and MPI-Sintel are close. In particular, running the experiments
with the same parameters as MPI-Sintel decreases EPE-Noc by only $0.1$
pixels on the training set. This shows that our method does
not suffer from overfitting.

\begin{table}
\resizebox{\linewidth}{!}{\centering 
\begin{tabular}{|l||c|c||c|c||r@{ }l|}
\hline
Method  & EPE-noc  & EPE  & Out-Noc 3  & Out 3  & Time & \\
\hline
DiscreteFlow~\citep{discreteflow} & 1.3 & 3.6 & 5.77\% & 16.63\% & 180s & \\
FlowFields~\citep{flowfields} & 1.4 & 3.5 & 6.23\% & 14.01\% & 23s & \\
\textbf{DeepFlow}     & \textbf{1.4} & 5.3 & 6.61\% & 17.35\% & 22s &\\
BTF-ILLUM~\citep{demetz2014learning} & 1.5 & 2.8 & 6.52\% & 11.03\% & 80s &\\
EpicFlow~\citep{epicflow}  & 1.5  & 3.8  & 7.88\%  & 17.08\% & 16s &\\
TGV2ADCSIFT~\citep{Braux-Zin_2013_ICCV}  & 1.5  & 4.5  & 6.20\%  & 15.15\% & 12s$^\bullet$ &\\
DeepFlow*~\citep{DeepFlow}  & 1.5  & 5.8  & 7.22\%  & 17.79\% & 17s &\\
NLTGV-SC~\citep{ranftl2014non}  & 1.6 & 3.8 & 5.93\% & 11.96\% & 16s$^\bullet$ &\\
Data-Flow~\citep{Vogel2013}  & 1.9  & 5.5  & 7.11\%  & 14.57\% & 180s & \\
TF+OFM~\citep{kennedyoptical} & 2.0 & 5.0 & 10.22\% & 18.46\% & 350s &\\
\hline
\end{tabular}}
\caption{Results on Kitti test set. EPE-noc is the EPE over
non-occluded areas. Out-Noc 3 (resp.\ Out 3) refers to the percentage
of pixels where flow estimation has an error above 3 pixels in non-occluded
areas (resp.\ all pixels). 
DeepFlow* denotes the preliminary version of DeepFlow published in \cite{DeepFlow}.
$^\bullet$ denotes the usage of a GPU.
}
\label{tab:kitti} 
\end{table}

\paragraph{Results on Middlebury.}

We optimize the parameters on the Middlebury training set by minimizing
the average angular error with the same strategy as for MPI-Sintel.
We find weights quasi-zero for the matching term due to the absence
of large displacements. DeepFlow obtained an average endpoint error
of $0.4$ on the test which is competitive with the state of the art.

\section{Conclusion}

We have introduced  a dense matching algorithm, termed DeepMatching. 
The proposed algorithm gracefully handles complex non-rigid 
object deformations and repetitive textured regions. 
DeepMatching yields state-of-the-art performance for image matching,
 on the Mikolajczyk~\citep{Mikolajczyk2005}, the MPI-Sintel~\citep{sintel} 
 and the Kitti~\citep{kitti} datasets.
Integrated in a variational energy minimization approach \citep{Bro11a}, 
the resulting approach for optical flow estimation, termed DeepFlow,
shows competitive performance on optical flow benchmarks.

Future work includes incorporating a weighting of the patches in \eq\refp{eqn:quadrant} instead of 
weighting all patches equally to take into account that
different parts of a large patch may belong to different objects.
This could improve the performance of DeepMatching for thin objects, such as human limbs.

\begin{acknowledgements}
This work was supported by the European
integrated project AXES, the MSR/INRIA joint project, the LabEx PERSYVAL-Lab
(ANR-11-LABX-0025), and the ERC advanced grant ALLEGRO.
\end{acknowledgements}

\bibliographystyle{spbasic}

\end{document}